\documentclass{article} 
\usepackage{iclr2024_conference,times}

\usepackage[utf8]{inputenc} 
\usepackage[T1]{fontenc}    
\usepackage{microtype}
\usepackage{times}
\usepackage{graphicx}
\usepackage{amsmath}
\usepackage{amsfonts}
\usepackage{acronym}
\usepackage{enumitem}
\usepackage[pagebackref=true,breaklinks=true,colorlinks]{hyperref}
\usepackage{cleveref}
\definecolor{cite_color}{RGB}{80, 135, 208} 
\definecolor{link_color}{RGB}{0, 48, 143}
\hypersetup{colorlinks,linkcolor={link_color},citecolor={cite_color},urlcolor={red}}  
\usepackage{balance}
\usepackage{xspace}
\usepackage{setspace}
\usepackage{subcaption}
\usepackage[skip=3pt,font=small]{caption}
\usepackage{booktabs,tabularx,colortbl,multirow,array,makecell}
\usepackage{algorithm} 
\usepackage{algpseudocode}

\usepackage{amsmath,amsfonts,bm}

\def\eqref#1{equation~\ref{#1}}

\def\1{\bm{1}}

\DeclareMathAlphabet{\mathsfit}{\encodingdefault}{\sfdefault}{m}{sl}
\SetMathAlphabet{\mathsfit}{bold}{\encodingdefault}{\sfdefault}{bx}{n}

\newcommand{\R}{\mathbb{R}}

\DeclareMathOperator*{\argmax}{arg\,max}

\newtheorem{definition}{Definition}

\newtheorem{theorem}{Theorem}[section]

\newtheorem{lemma}[theorem]{Lemma}
\usepackage{booktabs}
\usepackage{arydshln}
\newcommand\Tstrut{\rule{0pt}{2.4ex}}         
\newcommand\Bstrut{\rule[-0.9ex]{0pt}{0pt}}   
\usepackage{wrapfig}

\usepackage[toc,page,header]{appendix}
\usepackage{minitoc}
\usepackage{diagbox} 

\usepackage{xcolor}

\usepackage{hyperref}
\usepackage{url}

\newcommand{\bing}[1]{{\color{red}{\small\bf\sf [bing: #1]}}}

\title{Class Incremental Learning via Likelihood \\ Ratio Based Task Prediction}

\author{
Haowei Lin$^{1}$, Yijia Shao$^{2}$, Weinan Qian$^{3}$, Ningxin Pan$^3$, Yiduo Guo$^3$, and Bing Liu$^{4}$\thanks{Correspondance author. Bing Liu <\texttt{liub@uic.edu}>}\\ 
$^1$Institute for Artificial Intelligence, Peking University\quad$^2$Stanford University\\$^3$Wangxuan Institute of Computer Technology, Peking University\\
$^4$Department of Computer Science, University of Illinois at Chicago\\
$^1$\texttt{linhaowei@pku.edu.cn}\quad $^2$\texttt{shaoyj@stanford.edu}\\$^3$\texttt{\{ypqwn, 2100017816, yiduo\}@stu.pku.edu.cn}\quad$^4$\texttt{liub@uic.edu}
}

\iclrfinalcopy 
\begin{document}
\doparttoc 
\faketableofcontents 

\maketitle

\begin{abstract}
\textit{Class incremental learning} (CIL) is a challenging setting of continual learning, which learns a series of tasks sequentially. Each task consists of a set of unique classes. The key feature of CIL is that no task identifier (or task-id) is provided at test time. Predicting the task-id for each test sample is a challenging problem. An emerging theory-guided approach (called TIL+OOD) is to train a task-specific model for each task in a shared network for all tasks based on a \textit{task-incremental learning} (TIL) method to deal with \textit{catastrophic forgetting}. The model for each task is an \textit{out-of-distribution} (OOD) detector rather than a conventional classifier. The OOD detector can perform both \textit{within-task} (\textit{in-distribution} (IND)) class prediction and OOD detection. The OOD detection capability is the key to task-id prediction during inference. However, this paper argues that using a traditional OOD detector for task-id prediction is sub-optimal because additional information (e.g., the replay data and the learned tasks) available in CIL can be exploited to design a better and principled method for task-id prediction. We call the new method \textbf{TPL} (\textit{\textbf{T}ask-id \textbf{P}rediction based on \textbf{L}ikelihood Ratio}). TPL markedly outperforms strong CIL baselines and has {\textbf{negligible catastrophic forgetting}}.\footnote{~The code of TPL is publicly available at \url{https://github.com/linhaowei1/TPL}.}
  
\end{abstract}

\section{Introduction}
\label{sec:introduction}

Continual learning learns a sequence of tasks, $1, 2, \cdots,  T$, 
incrementally~\citep{ke2022continualsurvey,de2021continual}. Each task $t$ 
consists of a set of classes to be learned. 
This paper focuses on the challenging CL setting of \emph{class-incremental learning} (CIL)~\citep{rebuffi2017icarl}. The key challenge of CIL lies in the absence of task-identifier (task-id) in testing. There is another CL setting termed \textit{task-incremental learning} (TIL), which learns a separate model or classifier for each task. In testing, the task-id is provided for each test sample so that it is classified by the task specific model.  

A main assumption of continual learning is that once a task is learned, its training data is no longer accessible. This causes  
\textit{catastrophic forgetting} (CF), which refers to performance degradation of previous tasks due to parameter updates in learning each new task~\citep{McCloskey1989}. An additional challenge specifically for CIL is \textit{inter-task class separation} (ICS) \citep{kimtheoretical}. That is, when learning a new task, it is hard to establish decision boundaries between the classes of the new task and the classes of the previous tasks without the training data of the previous tasks. 
Although in replay-based methods \citep{rebuffi2017icarl,kemker2017fearnet,lopez2017gradient}, a small number of training samples can be saved from each task (called the \textit{replay data}) to help deal with CF and ICS to some extent by jointly training the new task data and the replay data from previous tasks, the effect on CF and ICS is limited as the number of replay samples is very small. 


An emerging theoretically justified approach to solving CIL is to combine a TIL technique with an out-of-distribution (OOD) detection method, called the TIL+OOD approach~\citep{kimtheoretical}. The TIL method learns a model for each task in a shared network. The model for each task is not a traditional classifier but an OOD detector. Note that almost all OOD detection methods can perform two tasks (1) \textit{in-distribution} (IND) classification and (2) \textit{out-of-distribution} (OOD) detection~\citep{vaze2022open}. At test time, for each test sample, the system first computes a \textit{task-id prediction} (TP) probability and a \textit{within-task prediction} (WP) probability  \citep{kimtheoretical} (same as IND classification) for each task. The two probabilities are then combined to make the final classification decision, which produces state-of-the-art results~\citep{kimtheoretical,kim2023learnability}. In this approach, WP is usually very accurate because it uses the task-specific model. \textbf{TP is the key challenge}.

There is a related existing approach that first predicts task-id and then predicts the class of the test sample using the task-specific model~\citep{rajasegaran2020itaml,abati2020conditional,von2019continual}. However, what is new is that \cite{kimtheoretical} theoretically proved that TP is correlated with OOD detection of each task. Thus, the OOD detection capability of each task model can be used for task-id prediction of each test sample. The previous methods did not realize this and thus performed poorly~\citep{kimtheoretical}. 
In~\citet{kimtheoretical}, the authors used the TIL method \texttt{HAT}~\citep{serra2018overcoming} and OOD detection method \texttt{CSI}~\citep{tack2020csi}. HAT is a parameter isolation method for TIL, which learns a model for each task in a shared network and each task model is protected with learned masks to overcome CF. Each task model is an OOD detector based on \texttt{CSI}.\footnote{~In~\citep{kim2023learnability}, it was also shown that based on this approach, CIL is learnable.} 

{\color{black}Our paper argues that using traditional OOD detectors is not optimal for task-id prediction as they are not designed for CIL and thus do not exploit the information available in CIL for better task-id prediction. By leveraging the information in CIL, we can do much better. A new method for task-id prediction is proposed, which we call \textbf{TPL} (\textit{\textbf{T}ask-id \textbf{P}rediction based on \textbf{L}ikelihood Ratio}). 
It consists of two parts: (1) a new method to train each task model and (2) a novel and principled method for task-id prediction, i.e., to estimate the probability of a test sample $\boldsymbol{x}$ belonging to a task $t$, i.e., $\mathbf{P}(t|\boldsymbol{x})$). We formulate the estimation of $\mathbf{P}(t|\boldsymbol{x})$ as a binary selection problem between two events ``$\boldsymbol{x}$ belongs to $t$'' and ``$\boldsymbol{x}$ belongs to $t^c$''. $t^c$ is $t$'s complement with regard to the universal set $U_{\textit{CIL}}$, which consists of all tasks that have been learned, i.e., $U_{\textit{CIL}} = \{1,2,\cdots, T\}$ and $t^c = U_{\textit{CIL}} - \{t\}$. 


The idea of TPL is analogous to using OOD detection for task-id prediction in the previous work. However, there is a \textbf{crucial difference}. In traditional OOD detection, given a set $U_{\textit{IND}}$ of in-distribution classes, we want to estimate the probability that a test sample does not belong to any classes in $U_{\textit{IND}}$. This means the universal set $U_{\textit{OOD}}$ for OOD detection includes all possible classes in the world (except those in $U_{\textit{IND}}$), which is at least very large if not infinite in size and we have no data from $U_{\textit{OOD}}$. Then, there is no way we can estimate the distribution of $U_{\textit{OOD}}$. However, we can estimate the distribution of $U_{\textit{CIL}}$ based on the {saved replay data}\footnote{~In our case, the saved replay data are used to estimate the distribution of $U_{\textit{CIL}}$ rather than to replay them in training a new task like replay-based methods. Also, our work is not about online continual learning.}} from each task in CIL. This allows us to use the \textit{likelihood ratio} of $\mathcal P_{t}$ and $\mathcal P_{t^c}$ to provide a principled solution towards the binary selection problem and consequently to produce the task-id prediction probability $\mathbf{P}(t|\boldsymbol{x})$ as analyzed in Sec.~\ref{sec:likelihood_ratio_principle}, where $\mathcal P_{t}$ is the distribution of the data in task $t$ and $\mathcal P_{t^c}$ is the distribution of the data in $t^c$ (all other tasks than $t$), i.e., $t$'s complement ($t^c=U_{\textit{CIL}} - \{t\}$).


The proposed system (also called TPL) uses the learned masks in the TIL method \texttt{HAT} for overcoming CF but the model for each task within \texttt{HAT}  is not a traditional classifier but a model that facilitates task-id prediction (Sec.~\ref{sec.overview}). At test time, given a test sample, the proposed likelihood ratio method is integrated with a logit-based score using an energy function to compute the task-id prediction probability and within-task prediction probability for the test sample to finally predict its class. Our experiments \textbf{\textit{with and without}} using a pre-trained model show that TPL markedly outperforms strong baselines. {\color{black}With a pre-trained model, TPL has \textbf{almost no forgetting}} or \textbf{performance deterioration}. {We also found that the current formula for computing the \textbf{\textit{forgetting rate}} is \textbf{n\textit{ot appropriate}} for CIL.}

\section{Related Work}
\label{sec:related work}
\vspace{-1mm}
\textbf{OOD Detection.} OOD detection has been studied extensively. 
\citet{hendrycks2016baseline} use the maximum softmax probability (MSP) as the OOD score. 
Some researchers also exploit the {logit space}~\citep{liang2017enhancing, liu2020energy, sun2021react}, 
and the {feature space} to compute the distance from the test sample to the training data/IND distribution, e.g., Mahalanobis distance~\citep{lee2018simple} and KNN~\citep{sun2022out}. Some use real/generated OOD data~\citep{wang2022cmg,liu2020energy,lee2018training}. 
Our task-id prediction \textbf{does not use} any existing OOD method. 


\textbf{Continual Learning (CL).} 
Existing CL methods are of four main types. (1) \emph{Regularization-based} methods address forgetting (CF) by using regularizers in the loss function~\citep{kirkpatrick2017overcoming,zhu2021prototype} or orthogonal projection~\citep{zeng2019continual} to preserve previous important parameters. The regularizers in \texttt{DER}~\citep{yan2021dynamically} and \texttt{BEEF}~\citep{wang2022beef} are similar to OOD detection but they expand the network for each task and perform markedly poorer than our method. (2) \emph{Replay-based} methods save a few samples from each task and replay them in training new tasks~\citep{kemker2017fearnet,lopez2017gradient,li2022continual}. {However, replaying causes data imbalance~\citep{guo2023dealing,xiang2023tkil,ahn2021ss}.} (3) \emph{Parameter isolation} methods train a sub-network for each task. 
\texttt{HAT}~\citep{serra2018overcoming} and \texttt{SupSup}~\citep{wortsman2020supermasks} are two representative methods. This approach is mainly used in task-incremental learning (TIL) and can eliminate CF. (4) \textit{TIL+OOD} based methods have been discussed in Sec.~\ref{sec:introduction}. 

Recently, using pre-trained models has become a standard practice for CL in both NLP~\citep{ke2021achieving,ke2021adapting,ke2023continual,shao2023class}. and computer vision (CV)~\citep{kim2022multi,wang2022learning}. {See the surveys~\citep{ke2022continualsurvey,wang2023comprehensive,de2021continual,hadsell2020embracing}}.

Our work is closely related to CIL methods that employ a TIL technique and a task-id predictor. \texttt{iTAML}~\citep{rajasegaran2020itaml} assumes that each test batch is from a single task and uses the whole batch to detect the task-id. This assumption is unrealistic.   
\texttt{CCG}~\citep{abati2020conditional} uses a separate network to predict the task-id. \texttt{Expert} \texttt{Gate}~\citep{aljundi2017expert} builds a distinct auto-encoder for each task. \texttt{HyperNet}~\citep{von2019continual} and \texttt{PR-Ent}~\citep{henning2021posterior} use entropy to predict the task-id. However, these systems perform poorly as they did not realize that \emph{OOD detection is the key to task-id prediction}~\citep{kimtheoretical}, which proposed the TIL+OOD approach. \citet{kimtheoretical} gave two methods \texttt{HAT+CSI} 
and  \texttt{SupSup+CSI}~\citep{kimtheoretical}. 
These two methods do not use a pre-trained model or replay data. The same approach was also taken in  \texttt{MORE}~\citep{kim2022multi} and \texttt{ROW}~\citep{kim2023learnability} but they employ a pre-trained model and replay data in CIL. These methods have established a state-of-the-art performance. We have discussed how our proposed method \texttt{TPL} is different from them in the introduction section. 

\section{Overview of the Proposed Method}
\label{sec.overview}
\vspace{-1mm}
\textbf{Preliminary}. \textit{Class incremental learning} (CIL) learns a sequence of tasks $1,...,T$. Each task $t$ has an input space $\mathcal X^{(t)}$, a label space $\mathcal Y^{(t)}$, and a training set $\mathcal{D}^{(t)} = \{(\boldsymbol{x}_j^{(t)}, y_j^{(t)})\}_{j=1}^{n^{(t)}}$ drawn $i.i.d.$ from $\mathcal P_{\mathcal X^{(t)}\mathcal Y^{(t)}}$. The class labels of the tasks are disjoint, i.e., $\mathcal Y^{(i)} \cap \mathcal Y^{(k)} = \emptyset,\forall i \neq k$. The goal of CIL is to learn a function $f:\cup_{t=1}^{T} \mathcal X^{(t)} \rightarrow \cup_{t=1}^{T} \mathcal Y^{(t)}$ to predict the class label of each test sample $\boldsymbol{x}$.

\citet{kimtheoretical} proposed a theory for solving CIL. It decomposes the CIL probability of a test sample $\boldsymbol{x}$ of the $j$-th class $y_j^{(t)}$ in task $t$ into two probabilities (as the classes in all tasks are disjoint),
\begin{align}
    \mathbf{P}(y_j^{(t)} | \boldsymbol{x}) = \mathbf{P}(y_j^{(t)} | \boldsymbol{x}, t) \mathbf{P}(t | \boldsymbol{x}). \label{eq:cil_decomposition}
\end{align}
The two probabilities on the right-hand-side (R.H.S) define the CIL probability on the left-hand-side (L.H.S). 
The first probability on the R.H.S. is the \textbf{\textit{within-task prediction} (WP)} probability and the second probability on the R.H.S. is the \textbf{\textit{task-id prediction} (TP)} probability. Existing TIL+OOD methods basically use a traditional OOD detection method to build each task model. The OOD detection model for each task is exploited for estimating both TP and WP probabilities (see Sec.~\ref{sec:introduction}). 

\vspace{+1mm}
\textbf{Overview of the Proposed TPL.}
This paper focuses on proposing a novel method for estimating task-id prediction probability, i.e., the probability of a test sample $\boldsymbol{x}$ belonging to (or drawing from the distribution of) a task $t$, i.e.,  $\mathbf{P}(t | \boldsymbol{x})$ in Sec.~\ref{eq:cil_decomposition}. The WP probability $\mathbf{P}(y_j^{(t)} | \boldsymbol{x}, t)$ can be obtained directly from the model of each task. 

The mask-based method in \texttt{HAT} 
is used by our method to prevent CF. Briefly, in learning each task, it learns a model for the task and also a set of masks for those important neurons to be used later to prevent the model from being updated by future tasks. In learning a new task, the masks of previous models stop the gradient flow to those masked neurons in back-propagation, which eliminates CF. In the forward pass, all the neurons can be used, so the network is shared by all tasks. {We note that our method can also leverage some other TIL methods other than \texttt{HAT} to prevent CF (see~\Cref{HAT}).} 

\begin{figure} 
\centering
\includegraphics[width=1.0\textwidth]{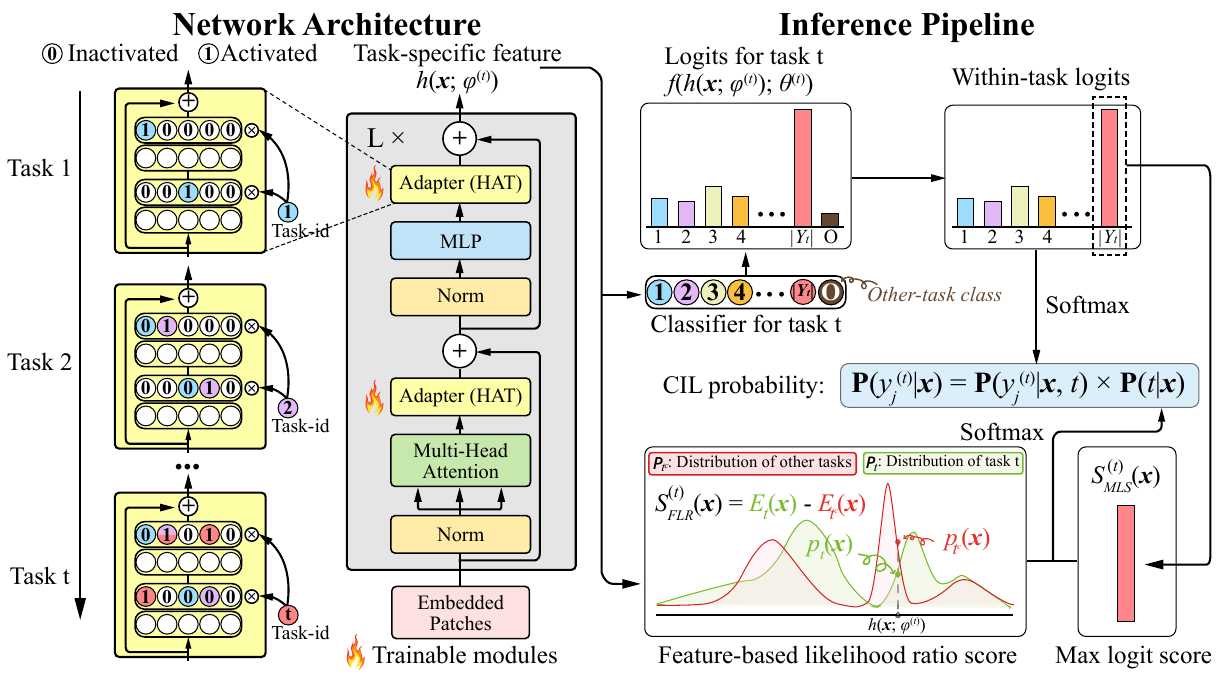}
\caption{Illustration of the proposed \texttt{TPL}. {We use a pre-trained transformer network (in the grey box) (see Sec.~\ref{sec:exp setup} for the case without using a pre-trained network). The pre-trained network is fixed and only the  adapters~\citep{houlsby2019parameter} inserted into the transformer are trainable to adapt to specific tasks.   \textbf{It is important to note that} the adapter (in yellow) used by HAT learns all tasks within the same adapter. The yellow boxes on the left show the progressive changes to the adapter as more tasks are learned. 
}}
\vspace{-3mm}
\label{fig:main}
\end{figure}

{The proposed method \texttt{TPL} is illustrated in~\Cref{fig:main}.} It has two techniques for accurate estimation of $\mathbf{P}(t | \boldsymbol{x})$, one in training and one in testing (inference). 


(1) \textbf{Training}: In the original \texttt{HAT}, each model is a traditional supervised classifier trained with cross-entropy. However, for our purpose of predicting task-id, this is insufficient because it has no consideration of the other classes learned from other tasks. In \texttt{TPL}, each model for a task $t$ is trained using the classes $\mathcal Y^{(t)}$ of task $t$ and an extra class (called O, for others) representing the replay buffer data $\textit{Buf}_{<t}$ of all the previous tasks. This enables each model to consider not only the new task data but also previous tasks' data, which facilitates more accurate computation of $\mathbf{P}(t | \boldsymbol{x})$. 


{For each task $t$, its model consists of a feature extractor $h(\boldsymbol{x}; \phi^{(t)})$ (partially shared with other tasks based on \texttt{HAT}), and a task-specific classifier $f(\boldsymbol{z}; \theta^{(t)})$. When learning task $t$, the model receives the training data $\mathcal D^{(t)}$ and the replay data $\textit{Buf}_{<t}$ (stored in a memory buffer). Then we minimize the loss:
\begin{align}
\label{eq.training_loss}
    \mathcal L(\theta^{(t)},\phi^{(t)}) = \mathbb E_{(\boldsymbol{x}, y)\sim \mathcal D^{(t)} \cup {\textit{Buf}_{<t}}}\left[\mathcal L_{\textit{CE}}(f(h(\boldsymbol{x};\phi^{(t)});\theta^{(t)}), {y})\right] + \mathcal L_{\textit{HAT}},
\end{align}
where $\mathcal L_{\textit{CE}}$ is the cross-entropy loss, $\mathcal L_{\textit{HAT}}$ is the regularization loss used in \texttt{HAT} (see~\Cref{HAT}).
} 

(2) \textbf{Testing (or inference)}: We follow~\cref{eq:cil_decomposition} to compute the CIL probability. The WP probability ($\mathbf{P}(y_j^{(t)} | \boldsymbol{x}, t)$) for each test sample is computed through softmax on only the original classes $\mathcal Y^{(t)}$ of task $t$, 
the first term on the right of \cref{eq:til+ood_practice} (also see the top right part in \Cref{fig:main}). The O class is not used in inference. 
{Note that the probabilities for different tasks can be computed in parallel.}   
\begin{equation}
    \mathbf{P}(y_j^{(t)} | \boldsymbol{x}) = \left[\textit{softmax}\left(f(h(\boldsymbol{x}; \phi^{(t)}); \theta^{(t)})\right)\right]_j \cdot  \mathbf{P}(t|\boldsymbol{x})
\label{eq:til+ood_practice}
\end{equation}
The class $y_j^{(t)}$ with the highest probability will be predicted as the class for test sample $\boldsymbol{x}$.\footnote{We also calibrate the probabilities from different task models, but it has little effect (see~\Cref{calibration}).} 
{\color{black}We discuss the proposed method for computing task-id prediction probability $\mathbf{P}(t|\boldsymbol{x})$ (see the bottom right part in \Cref{fig:main}) in the next section. Training will not be discussed any further.}

\section{Estimating Task-id Prediction Probability}
\label{method}




\subsection{Theoretical Analysis}
\label{sec:likelihood_ratio_principle}

As noted in Sec.~\ref{sec:introduction}, we estimate the TP probability $\textbf{P}(t|\boldsymbol{x})$ by predicting whether a sample $\boldsymbol{x}$ is drawn from the distribution $\mathcal P_{t}$ of task $t$ or drawn from the distribution of $t$'s complement $t^c$, i.e., $\mathcal{P}_{t^c}$. We denote the universal set $U_{\textit{CIL}}$ of all tasks (or task-ids) that have been learned, i.e., $U_{\textit{CIL}}=\{1, 2, \cdots,  T\}$ and $t^c = U_{\textit{CIL}}-\{t\}$.
From a frequentist perspective, our objective can be formulated as a binary hypothesis test:
\begin{equation}
\mathcal H_0: \boldsymbol{x}\sim \mathcal P_{t}\quad v.s. \quad \mathcal H_1:\boldsymbol{x} \sim \mathcal P_{t^c} ,
\label{hypothesis test}
\end{equation}
Using the Neyman-Pearson lemma~\citep{neyman1933ix}, we can derive a theorem that demonstrates the principled role of likelihood ratio in this task (the proofs are given in~\Cref{proof1}):

\begin{theorem}
\label{theorem:lr}
    A test with rejection region $\mathcal R$ defined as follows is a unique \textbf{uniformly most powerful (UMP) test} for the hypothesis test problem defined in~\cref{hypothesis test}:
    \[\mathcal R:=\{\boldsymbol x:  p_{t}(\boldsymbol x) / p_{t^c}(\boldsymbol x) < \lambda_0\}.\]
    where $\lambda_0$ is a threshold that can be chosen to obtain a specified significance level.
\end{theorem}

\begin{theorem}
\label{theorem:max_auc}
    The UMP test for hypothesis test defined in~\cref{hypothesis test} maximizes the Area Under the Curve (AUC) of binary classification between $\mathcal P_t$ and $\mathcal P_{t^c}$.
\end{theorem}

\Cref{theorem:lr,theorem:max_auc} highlight the importance of detecting samples that do not belong to task $t$ based on low $t$ density $p_{t}(\boldsymbol x)$ and high $t^c$ density $p_{t^c}(\boldsymbol x)$. 

Note that in traditional OOD detection, the system has no access to the true OOD distribution $\mathcal P_{t^c}$ but only $\mathcal P_{t}$ (IND distribution). Some existing methods resort to a proxy distribution $\mathcal P_{t^c}^{\textit{proxy}}$, such as a uniform distribution~\citep{nalisnick2018deep} or an auxiliary data distribution~\citep{Lin2023FLatSPO} as the universal set $U$ is the set of all classes in the world and the universal set of all OOD classes for task $t$ denoted by $U^{(t)}_{\textit{OOD}}$ is very large if not infinite. This approach can lead to potential risks. For instance, consider a scenario where $\mathcal P_{t^c} = \mathcal N(0, 0.01)$ and $\mathcal P_{t} =  \mathcal N(0, 1)$. It is apparent that $p_{t}(0) > p_{t}(1)$, {but 0 is more likely to belongs to $\mathcal P_{t^c}$ than 1 as $0.1 = {p_{t}(0)}/{p_{t^c}(0)} < {p_{t}(1)}/{p_{t^c}(1)} =0.1\cdot\mathrm{e}^{49.5}$. {We further show the failure cases in real CIL scenarios in~\Cref{app:vis}.}

\textbf{Good News for CIL.} In CIL, the IND distribution $\mathcal P_{t}$ for task $t$ can be interpreted as the marginal distribution $\mathcal P_{\mathcal X^{(t)}}$, while $\mathcal P_{t^c}$ corresponds to a mixture distribution $\mathcal P_{\mathcal X^{(t^c)}}$ comprising the individual marginal distributions 
$\{\mathcal P_{\mathcal X^{(t^*)}}\}_{t^*\ne t}$ 
(which can be estimated based on the saved replay data), each of which is assigned the equal mixture weight. Consequently, we have the knowledge of $\mathcal P_{t^c}$ in CIL, thereby offering an opportunity to estimate $\mathcal P_{t^c}$ to be used to compute task-id prediction $\mathbf{P}(t | \boldsymbol{x})$ more accurately. This 
leads to our design of \texttt{TPL} in the following subsections.

\subsection{Computing Task-ID Prediction Probability}
\label{sec:OODscore}


We now present the proposed method for computing the task-id prediction probability $\mathbf{P}(t | \boldsymbol{x})$, which has three parts: (1) estimating both $\mathcal P_{t}$ and $\mathcal P_{t^c}$ (as analyzed in Sec.~\ref{sec:likelihood_ratio_principle}) and computing the likelihood ratio, (2) integrating the likelihood ratio based score with a logit-based score for further improvement, and (3) applying a softmax function on the scores for all tasks to obtain the task-id prediction probability for each task. {The three parts correspond to the bottom right part of~\Cref{fig:main}.}


\subsubsection{Estimating $\mathcal P_{t}$ and $\mathcal P_{t^c}$ and Computing Likelihood Ratio}
\label{sec:LR}

Guided by~\Cref{theorem:lr}, we design a task-id prediction score based on the likelihood ratio $p_{t}(\boldsymbol{x})/p_{t^c}(\boldsymbol{x})$. However, due to the challenges in directly estimating the data distribution within the high-dimensional raw image space, we instead consider estimation in the low-dimensional \emph{feature space}. Interestingly, many distance-based OOD detection scores can function as density estimators that estimate the IND density $p(\boldsymbol{x})$ in the \emph{feature space} (see~\Cref{sec:distance_density} for justifications). For instance, \texttt{MD} (\textit{Mahalanobis Distance})~\citep{lee2018simple} estimates distributions using Gaussian mixture models, while \texttt{KNN}~\citep{sun2022out} uses non-parametric estimation. Our method \texttt{TPL} also uses the two scores to estimate distributions (i.e., $\mathcal P_{t}$ and $\mathcal P_{t^c}$ in our case).  

To connect the normalized probability density with unnormalized task-id prediction scores, we leverage energy-based models (EBMs) to parameterize $\mathcal P_{t}$ and $\mathcal P_{t^c}$. Given a test sample $\boldsymbol x$, it has density $p_{t}(\boldsymbol{x})=\exp\{E_{t}(\boldsymbol{x})\}/Z_1$ in $\mathcal P_{t}$, and density $p_{t^c}(\boldsymbol{x})=\exp\{E_{t^c}(\boldsymbol{x})\}/Z_2$ in $\mathcal P_{t^c}$, where $Z_1,Z_2$ are normalization constants that ensure the integral of densities $p_{t}(\boldsymbol{x})$ and $p_{t^c}(\boldsymbol{x})$ equal 1, and $E_{t}(\cdot), E_{t^c}(\cdot)$ are called \emph{energy functions}.\footnote{~In EBMs, the density $p(x)$ is typically defined as $\exp\{-E(x)\}/Z$. Since our task-id prediction score is defined to measure the likelihood that the test sample belongs to a task, the energy function here is defined as positively related to the probability density.} Consequently, we can design a feature-based task-id prediction score using the \textbf{L}ikelihood \textbf{R}atio (LR), {which is also shown at the bottom right of~\Cref{fig:main}}: 
\begin{align}
S^{(t)}_{\textit{LR}}(\boldsymbol{x}) = \log(p_{t}(\boldsymbol{x})/p_{t^c}(\boldsymbol{x})) = E_{t}(\boldsymbol{x})-E_{t^c}(\boldsymbol{x}) + \log(Z_2/Z_1).
\label{eq:S-LR}
\end{align} 
Since $\log(Z_2/Z_1)$ is a constant, it can be omitted in the task-id prediction score definition:
\vspace{-0.01em}
\begin{align}
    S^{(t)}_{\textit{LR}}(\boldsymbol{x}) := E_{t}(\boldsymbol{x})-E_{t^c}(\boldsymbol{x}),
    \label{eq:ind-ood}
\end{align} 

Since the energy functions $E_{t}(\cdot)$ and $E_{t^c}(\cdot)$ need not to be normalized, we estimate them with the above scores. 
We next discuss how to choose specific $E_{t}(\cdot)$ and $E_{t^c}(\cdot)$ for~\cref{eq:ind-ood}. 

For in-task energy $E_{t}(\boldsymbol x)$ of a task, we simply adopt an OOD detection score $S_{\textit{MD}}(\boldsymbol x)$, which is the OOD score for \texttt{MD} and is defined as the inverse of the minimum Mahalanobis distance of feature $h(\boldsymbol x; \phi^{(t)})$ to all class centroids. The details of how $S^{(t)}_{\textit{MD}}(\boldsymbol x)$ is computed are given in~\Cref{sec:MD}.

For out-of-task energy $E_{t^c}(\boldsymbol x)$ of a task, we use \textbf{replay data} from other tasks for estimation. Let $\textit{Buf}_{t^c}$ be the set of buffer/replay data excluding the data of classes in task $t$. We set $E_{t^c}(\boldsymbol x) = -d_{\textit{KNN}}(\boldsymbol x,\textit{Buf}_{t^c})$, where $d_{\textit{KNN}}(\boldsymbol x, \textit{Buf}_{t^c})$ is the $k$-nearest distances of the feature $h(\boldsymbol x;\phi)^{(t)}$ to the set of features of the replay $\textit{Buf}_{t^c}$ data. If $d_{\textit{KNN}}(\boldsymbol x, \textit{Buf}_{t^c})$ is small, it means the distance between $\boldsymbol x$ and replay $\textit{Buf}_{t^c}$ data is small in the feature space. The vanilla $\texttt{KNN}$ score is $S^{(t)}_{\textit{KNN}}(\boldsymbol{x})=-d_{\textit{KNN}}(\boldsymbol{x}, \mathcal D^{(t)})$, which was originally designed to estimate $p_{t}(\boldsymbol{x})$ using the training set $\mathcal D^{(t)}$. Here we adopt it to estimate $p_{t^c}(\boldsymbol{x})$ using the replay data ($\textit{Buf}_{t^c}$). Finally, we obtain,
\begin{align}
    S^{(t)}_{\textit{LR}}(\boldsymbol{x}):= \underbrace{\alpha\cdot S^{(t)}_{\textit{MD}}(\boldsymbol x)}_{E_{t}(\boldsymbol x)} + \underbrace{d_{\textit{KNN}}(\boldsymbol{x}, \textit{Buf}_{t^c})}_{-E_{t^c}(\boldsymbol x)},
\label{eq:LR}
\end{align}
where $\alpha$ is a hyper-parameter to make the two scores comparable. This is a principled task-id prediction score as justified in Sec.~\ref{sec:likelihood_ratio_principle}.

\textbf{Remarks.}  We can also use some other feature-based estimation methods instead of \texttt{MD} and \texttt{KNN} in $S_{\textit{LR}}(\boldsymbol{x})$. The reason why we choose \texttt{MD} to estimate $\mathcal P_{t}$ is that it does not require the task data at test time (but \texttt{KNN} does), and we choose \texttt{KNN} to estimate $\mathcal P_{t^c}$ because the non-parametric estimator \texttt{KNN} is high performing~\citep{yang2022openood} and we use only the saved replay data for this. We will conduct an ablation study using different estimation methods for both $\mathcal P_{t}$ and $\mathcal P_{t^c}$ in Sec.~\ref{sec:ablation}.

\subsubsection{Combining with a Logit-Based Score}
\label{sec:composition}

To further improve the task-id prediction score, we combine the feature-based $S_{\textit{LR}}$ score with a logit-based score, which has been shown quite effective in OOD detection~\citep{wang2022vim}.  



We again develop an energy-based model (EBM) framework for the combination that offers a principled approach to composing different task-id prediction scores.
Specifically, to combine the proposed feature-based $S^{(t)}_{\textit{LR}}(\cdot)$ score with a logit-based score (an energy function) $S^{(t)}_{\textit{logit}}(\cdot)$, 
we can make the composition as:
\begin{align}
    E_{\textit{composition}}(\boldsymbol{x}) = 
        \log( \exp\{\alpha_{1}\cdot S^{(t)}_{\textit{logit}}(\boldsymbol{x})\} +\exp\{\alpha_2\cdot S^{(t)}_{\textit{LR}}(\boldsymbol{x})\}), 
\label{eq:energy_composition}
\end{align}
where $\alpha_1$ and $\alpha_2$ are scaling terms to make different scores comparable. As noted in~\citep{du2020compositional}, the composition emulates an OR gate for energy functions.  

To choose a logit-based method for $S^{(t)}_{\textit{logit}}(\cdot)$ in~\cref{eq:energy_composition}, we opt for the simple yet effective method \texttt{MLS} score $S^{(t)}_{\textit{MLS}}(\boldsymbol x)$, which is defined as the \emph{maximum logit} of $\boldsymbol x$ {(also shown on the right of~\Cref{fig:main})}. 

Our final score $S^{(t)}_{\textit{TPL}}(\boldsymbol{x})$, which integrates feature-based $S^{(t)}_{\textit{LR}}(\cdot)$ and the logit-based $S^{(t)}_{\textit{MLS}}(\cdot)$ scores, uses the composition in Eq.~\ref{eq:energy_composition}:
\vspace{-1mm}
\begin{align}
    S^{(t)}_{\textit{TPL}}(\boldsymbol{x}) = \log\left(\exp\{\beta_1\cdot S^{(t)}_{\textit{MLS}}(\boldsymbol{x})\}
    +\exp\{\beta_2\cdot S^{(t)}_{\textit{MD}}(\boldsymbol{x}) + d_{\textit{KNN}}(\boldsymbol{x}, \textit{Buf}_{t_c})\}\right),
    \label{eq:final_score}
\end{align}
where $\beta_1$ and $\beta_2$ 
are scaling terms, which are given by merging $\alpha$ in~\cref{eq:LR} and $\alpha_1,\alpha_2$ in~\cref{eq:energy_composition}. Since the scale of $d_{\textit{KNN}}(\cdot)$ is near to 1, we simply choose $\beta_1$ and $\beta_2$ to be the inverse of empirical means of $S^{(t)}_{\textit{MLS}}(\boldsymbol x)$ and $S^{(t)}_{\textit{MD}}(\boldsymbol{x})$ estimated by the training data $\mathcal D^{(t)}$ to make different scores comparable:
\begin{align}
    \frac1{\beta_1} = \frac1{|\mathcal D^{(t)}|}\sum_{\boldsymbol{x}\in \mathcal D^{(t)}} S^{(t)}_{\textit{MLS}}(\boldsymbol{x}),\quad \frac1{\beta_2} = \frac1{|\mathcal D^{(t)}|}\sum_{\boldsymbol{x}\in \mathcal D^{(t)}} S^{(t)}_{\textit{MD}}(\boldsymbol{x})
    \label{eq:beta1_beta2}
\end{align}
\textbf{Remarks}. We exploit \textbf{EBMs}, which are known for their \emph{flexibility} but suffering from \emph{intractability}. However, we exploit EBMs' \emph{flexibility} to derive principled task-id prediction score following~\Cref{theorem:lr,eq:energy_composition}, while keeping the \emph{tractability} via approximation using OOD scores (\texttt{MD}, \texttt{KNN}, \texttt{MLS}) in practice. This makes our proposed \texttt{TPL} maintain both theoretical and empirical soundness.

\subsection{Converting Task-id Prediction Scores to Probabilities}
Although theoretically principled as shown in Sec.~\ref{sec:likelihood_ratio_principle}, our final task-id prediction score is still an unnormalized energy function. We convert the task-id prediction scores for all tasks (i.e., $\{S_{\textit{TPL}}^{(t)}(\boldsymbol{x})\}_{t=1}^T$) to normalized probabilities via softmax:
\begin{equation}
    \textbf{P}(t|\boldsymbol{x}) = \textit{softmax}\left(\left[S_{\textit{TPL}}^{(1)}(\boldsymbol{x}),S_{\textit{TPL}}^{(2)}(\boldsymbol{x}),\cdots,S_{\textit{TPL}}^{(T)}(\boldsymbol{x})\right] / \gamma\right)_t,
\end{equation}
where $\gamma$ is a temperature parameter. To encourage confident task-id prediction, we set a low temperature $\gamma=0.05$ to produce a low entropy task-id preidction distribution for all our experiments.

\section{Experiments}
\label{experiment}

\vspace{-1mm}
\subsection{Experimental Setup}
\label{sec:exp setup}

\textbf{CIL Baselines.} We use \textbf{17 \textit{baselines}}, including \textbf{11 \textit{replay methods}}: \texttt{iCaRL}~\citep{rebuffi2017icarl}, \texttt{A-GEM}~\citep{chaudhry2018efficient}, \texttt{EEIL}~\citep{castro2018end}, \texttt{GD}~\citep{lee2019overcoming}, \texttt{DER++}~\citep{buzzega2020dark}, \texttt{HAL}~\citep{chaudhry2021using}, \texttt{DER}~\citep{yan2021dynamically}, \texttt{FOSTER}~\citep{wang2022foster}, {AFC~\citep{kang2022class}},
\texttt{BEEF}~\citep{wang2022beef},
\texttt{MORE}~\citep{kim2022multi}, \texttt{ROW}~\citep{kim2023learnability}, and \textbf{6 \textit{non-replay methods}}: \texttt{HAT}~\citep{serra2018overcoming}, 
\texttt{ADAM}~\citep{zhou2023revisiting},
\texttt{OWM}~\citep{zeng2019continual}, \texttt{PASS}~\citep{zhu2021prototype}, \texttt{SLDA}~\citep{hayes2020lifelong}, and \texttt{L2P}~\citep{wang2022learning}.\footnote{~The systems \texttt{HAT+CSI} and \texttt{Sup+CSI} in~\citep{kimtheoretical} (which are based on the TIL+OOD paradigm but do not use a pre-trained model) are not included as they are much weaker 
because their contrastive learning and data augmentations do not work well with a pre-trained model.} {\color{black}We follow~\citep{kimtheoretical} to adapt \texttt{HAT} (which is a TIL method) for CIL and call it \texttt{HAT}$_{CIL}$.}
\textbf{Implementation details, network size and running time} are given in~\Cref{sec:implementation}.

\textbf{Datasets.} To form a sequence of tasks in CIL experiments, we follow the common CIL setting. We split CIFAR-10 into 5 tasks (2 classes per task) (\textbf{C10-5T}). For CIFAR-100, we conduct two experiments: 10 tasks (10 classes per task) (\textbf{C100-10T}) and 20 tasks (5 classes per task) (\textbf{C100-20T}). For TinyImageNet, we split 200 classes into 5 tasks (40 classes per task) (\textbf{T-5T}) and 10 tasks (20 classes per task) (\textbf{T-10T}). We set the replay buffer size for CIFAR-10 as 200 samples, and CIFAR-100 and TinyImageNet as 2000 samples following~\cite{kim2023learnability}. Following the random class order protocol in~\cite{rebuffi2017icarl}, we randomly generate five different class orders for each experiment and report the averaged metrics over the 5 random orders. For a fair comparison, the class orderings are kept the same for all systems. Results on a larger dataset are given in~\Cref{app:imagenet}.

\textbf{Backbone Architectures.} We conducted two sets of experiments, one \textbf{using a pre-trained model} and one \textbf{without using a pre-trained model}. Here we focus on using a pre-trained model as that is getting more popular. Following the TIL+OOD works~\citep{kim2022multi,kim2023learnability}, TPL uses the same DeiT-S/16 model~\citep{touvron2021training} pre-trained using 611 classes of ImageNet after removing 389 classes that are similar or identical to the classes of the experiment data CIFAR and TinyImageNet to prevent information leak~\citep{kim2022multi,kim2023learnability}. To leverage the pre-trained model while adapting to new knowledge, 
we insert an adapter module~\citep{houlsby2019parameter} at each transformer layer except \texttt{SLDA} and \texttt{L2P}.\footnote{~\texttt{SLDA} fine-tunes only the classifier with a fixed feature extractor and \texttt{L2P} trains learnable prompts.} The adapter modules, classifiers, and layer norms are trained using \texttt{HAT} while the transformer parameters are fixed to prevent CF. The hidden dimension of adapters is 64 for CIFAR-10, and 128 for CIFAR-100 and TinyImageNet. {For completeness, we also report the results of TPL using DeiT-S/16 \textbf{P}re-trained with the \textbf{F}ull \textbf{I}mageNet (called \textbf{TPL}$_{\textbf{PFI}}$) in the pink rows of~\Cref{tab.results}}. The results without using a pre-trained model are given in~\Cref{app:cil_resnet}.

{
\textbf{Evaluation Metrics.} We use threepopular  metrics: (1) \emph{accuracy after learning the final task} (\textbf{Last} in~\Cref{tab.results}), (2) \emph{average incremental accuracy} (\textbf{AIA} in~\Cref{tab.results}), and (3) \emph{forgetting rate} (see~\Cref{fig:fr} in \Cref{sec:forgetting_rate}, where we also discuss why the \textbf{current forgetting rate formula} is \textbf{not appropriate} for CIL, but only for TIL.  The definitions of all these metrics are given in~\Cref{sec:metrics}.
}

\subsection{Results and Comparisons}
\label{sec:results}
\begin{table} 
\centering
\vspace{-3.1em}
\caption{CIL ACC (\%). ``-XT": X number of tasks. The best result in each column is highlighted in bold. The baselines are divided into two groups via the dashed line. The first group contains non-replay methods, and the second group contains replay-based methods. \textbf{Non-CL} (non-continual learning) denotes pooling all tasks together to learn all classes as one task, which gives the performance \textbf{upper bound} for CIL. {\textbf{AIA} is the \textit{average incremental} ACC (\%). \textbf{Last} is the ACC after learning the final task.} See \textbf{forgetting rate results} in~\Cref{sec:forgetting_rate}. The {\color{pink}\textbf{pink}} rows also show the results of Non-CL$_{\text{PFI}}$ and TPL$_{\text{PFI}}$, which use DeiT \textbf{P}re-trained with \textbf{F}ull \textbf{I}mageNet.}
\Large
\resizebox{\linewidth}{!}{
\begin{tabular}{c!{\vrule width \lightrulewidth}cccccccccc!{\vrule width \lightrulewidth}cc} 
\toprule
         & \multicolumn{2}{c}{\textbf{~~ C10-5T~~}} & \multicolumn{2}{c}{\textbf{C100-10T}} & \multicolumn{2}{c}{\textbf{C100-20T}} & \multicolumn{2}{c}{\textbf{~ ~T-5T~ ~}} & \multicolumn{2}{c}{\textbf{~ ~T-10T~ ~}} & \multicolumn{2}{c}{\textbf{~Average~}}  \\ 
         & \textbf{Last} & \textbf{AIA} & \textbf{Last} & \textbf{AIA} & \textbf{Last} & \textbf{AIA} & \textbf{Last} & \textbf{AIA} & \textbf{Last} & \textbf{AIA} & \textbf{Last} & \textbf{AIA} \\
\midrule
\Bstrut
\textbf{Non-CL} &  $\text{95.79}^{\pm\text{0.15}}$&  $\text{97.01}^{\pm\text{0.14}}$                 &   {$\text{82.76}^{\pm\text{0.22}}$}& {$\text{87.20}^{\pm\text{0.29}}$}               &  {$\text{82.76}^{\pm\text{0.22}}$}&$\text{87.53}^{\pm\text{0.31}}$                 &   {$\text{72.52}^{\pm\text{0.41}}$}&$\text{77.03}^{\pm\text{0.47}}$                  &     {$\text{72.52}^{\pm\text{0.41}}$}&$\text{77.03}^{\pm\text{0.41}}$                 &  {$\text{81.27}$}& $\text{85.16}$                   \\
\hline
\Tstrut
OWM &
$\text{41.69}^{\pm\text{6.34}}$& 
$\text{56.00}^{\pm\text{3.46}}$& 
$\text{21.39}^{\pm\text{3.18}}$&
$\text{40.10}^{\pm\text{1.86}}$& 
$\text{16.98}^{\pm\text{4.44}}$&
$\text{32.58}^{\pm\text{1.58}}$& 
$\text{24.55}^{\pm\text{2.48}}$&
$\text{45.18}^{\pm\text{0.33}}$& 
$\text{17.52}^{\pm\text{3.45}}$&
$\text{35.75}^{\pm\text{2.21}}$& 
$\text{24.43}$&$\text{41.92}$   \\
ADAM & 
$\text{83.92}^{\pm\text{0.51}}$&  
$\text{90.33}^{\pm\text{0.42}}$& 
$\text{61.21}^{\pm\text{0.36}}$&  
$\text{72.55}^{\pm\text{0.41}}$& 
$\text{58.99}^{\pm\text{0.61}}$&  
$\text{70.89}^{\pm\text{0.51}}$& 
$\text{50.11}^{\pm\text{0.46}}$&  
$\text{61.85}^{\pm\text{0.51}}$& 
$\text{49.68}^{\pm\text{0.40}}$&
$\text{61.44}^{\pm\text{0.44}}$& 
$\text{60.78}$&$\text{71.41}$               \\
PASS & 
$\text{86.21}^{\pm\text{1.10}}$&               
$\text{89.03}^{\pm\text{7.13}}$& 
$\text{68.90}^{\pm\text{0.94}}$&             
$\text{77.01}^{\pm\text{2.44}}$& 
$\text{66.77}^{\pm\text{1.18}}$&             
$\text{76.42}^{\pm\text{1.23}}$& 
$\text{61.03}^{\pm\text{0.38}}$&               
$\text{67.12}^{\pm\text{6.26}}$& 
$\text{58.34}^{\pm\text{0.42}}$&                
$\text{67.33}^{\pm\text{3.63}}$& 
$\text{68.25}$&$\text{75.38}$               \\
\text{HAT}$_{CIL}$ & 
$\text{82.40}^{\pm\text{0.12}}$&
$\text{91.06}^{\pm\text{0.36}}$&
$\text{62.91}^{\pm\text{0.24}}$&
$\text{73.99}^{\pm\text{0.86}}$&
$\text{59.54}^{\pm\text{0.41}}$&
$\text{69.12}^{\pm\text{1.06}}$&
$\text{59.22}^{\pm\text{0.10}}$&
$\text{69.38}^{\pm\text{1.14}}$&
$\text{54.03}^{\pm\text{0.21}}$&
$\text{65.63}^{\pm\text{1.64}}$&
$\text{63.62}$&$\text{73.84}$           \\
SLDA & 
$\text{88.64}^{\pm\text{0.05}}$&
$\text{93.54}^{\pm\text{0.66}}$& 
$\text{67.82}^{\pm\text{0.05}}$&
$\text{77.72}^{\pm\text{0.58}}$& 
$\text{67.80}^{\pm\text{0.05}}$&
$\text{78.51}^{\pm\text{0.58}}$& 
$\text{57.93}^{\pm\text{0.05}}$&
$\text{66.03}^{\pm\text{1.35}}$& 
$\text{57.93}^{\pm\text{0.06}}$&
$\text{67.39}^{\pm\text{1.81}}$& 
$\text{68.02}$&$\text{76.64}$               \\
\Bstrut L2P &
$\text{73.59}^{\pm\text{4.15}}$&
$\text{84.60}^{\pm\text{2.28}}$&
$\text{61.72}^{\pm\text{0.81}}$&
$\text{72.88}^{\pm\text{1.18}}$& 
$\text{53.84}^{\pm\text{1.59}}$&
$\text{66.52}^{\pm\text{1.61}}$& 
$\text{59.12}^{\pm\text{0.96}}$&
$\text{67.81}^{\pm\text{1.25}}$& 
$\text{54.09}^{\pm\text{1.14}}$&
$\text{64.59}^{\pm\text{1.59}}$& 
$\text{60.47}$&$\text{71.28}$               \\
\hdashline
\Tstrut iCaRL & 
$\text{87.55}^{\pm\text{0.99}}$&
$\text{89.74}^{\pm\text{6.63}}$& 
$\text{68.90}^{\pm\text{0.47}}$&
$\text{76.50}^{\pm\text{3.56}}$& 
$\text{69.15}^{\pm\text{0.99}}$&
$\text{77.06}^{\pm\text{2.36}}$& 
$\text{53.13}^{\pm\text{1.04}}$&
$\text{61.36}^{\pm\text{6.21}}$& 
$\text{51.88}^{\pm\text{2.36}}$&
$\text{63.56}^{\pm\text{3.08}}$& 
$\text{66.12}$&$\text{73.64}$              \\
A-GEM & 
$\text{56.33}^{\pm\text{7.77}}$&    
$\text{68.19}^{\pm\text{3.24}}$& 
$\text{25.21}^{\pm\text{4.00}}$&
$\text{43.83}^{\pm\text{0.69}}$& 
$\text{21.99}^{\pm\text{4.01}}$&
$\text{35.97}^{\pm\text{1.15}}$& 
$\text{30.53}^{\pm\text{3.99}}$&
$\text{49.26}^{\pm\text{0.64}}$& 
$\text{21.90}^{\pm\text{5.52}}$&
$\text{39.58}^{\pm\text{3.32}}$& 
$\text{31.19}$&$\text{47.37}$               \\
EEIL & 
$\text{82.34}^{\pm\text{3.13}}$&
$\text{90.50}^{\pm\text{0.72}}$& 
$\text{68.08}^{\pm\text{0.51}}$&
$\text{81.10}^{\pm\text{0.37}}$& 
$\text{63.79}^{\pm\text{0.66}}$&
$\text{79.54}^{\pm\text{0.69}}$& 
$\text{53.34}^{\pm\text{0.54}}$&
$\text{66.63}^{\pm\text{0.40}}$& 
$\text{50.38}^{\pm\text{0.97}}$&
$\text{66.54}^{\pm\text{0.61}}$& 
$\text{63.59}$&$\text{76.86}$               \\
GD & 
$\text{89.16}^{\pm\text{0.53}}$&
$\text{94.22}^{\pm\text{0.75}}$& 
$\text{64.36}^{\pm\text{0.57}}$&
$\text{80.51}^{\pm\text{0.57}}$& 
$\text{60.10}^{\pm\text{0.74}}$&
$\text{78.43}^{\pm\text{0.76}}$& 
$\text{53.01}^{\pm\text{0.97}}$&
$\text{67.51}^{\pm\text{0.38}}$& 
$\text{42.48}^{\pm\text{2.53}}$&
$\text{63.91}^{\pm\text{0.40}}$& 
$\text{61.82}$&$\text{76.92}$               \\
DER++ & 
$\text{84.63}^{\pm\text{2.91}}$&
$\text{89.01}^{\pm\text{6.29}}$& 
$\text{69.73}^{\pm\text{0.99}}$&
$\text{80.64}^{\pm\text{2.74}}$& 
$\text{70.03}^{\pm\text{1.46}}$&
$\text{81.72}^{\pm\text{1.76}}$& 
$\text{55.84}^{\pm\text{2.21}}$&
$\text{66.55}^{\pm\text{3.73}}$& 
$\text{54.20}^{\pm\text{3.28}}$&
$\text{67.14}^{\pm\text{1.40}}$& 
$\text{66.89}$&$\text{77.01}$               \\
HAL & 
$\text{84.38}^{\pm\text{2.70}}$&
$\text{87.00}^{\pm\text{7.27}}$& 
$\text{67.17}^{\pm\text{1.50}}$&
$\text{77.42}^{\pm\text{2.73}}$& 
$\text{67.37}^{\pm\text{1.45}}$&
$\text{77.85}^{\pm\text{1.71}}$& 
$\text{52.80}^{\pm\text{2.37}}$&
$\text{65.31}^{\pm\text{3.68}}$& 
$\text{55.25}^{\pm\text{3.60}}$&
$\text{64.48}^{\pm\text{1.45}}$& 
$\text{65.39}$&$\text{74.41}$               \\
DER & 
$\text{86.79}^{\pm\text{1.20}}$&
$\text{92.83}^{\pm\text{1.10}}$& 
$\text{73.30}^{\pm\text{0.58}}$&
$\text{82.89}^{\pm\text{0.45}}$& 
$\text{72.00}^{\pm\text{0.57}}$&
$\text{82.79}^{\pm\text{0.76}}$& 
$\text{59.57}^{\pm\text{0.89}}$&
$\text{70.32}^{\pm\text{0.57}}$& 
$\text{57.18}^{\pm\text{1.40}}$&
$\text{70.21}^{\pm\text{0.86}}$& 
$\text{69.77}$&$\text{79.81}$               \\
FOSTER & 
$\text{86.09}^{\pm\text{0.38}}$&
$\text{91.54}^{\pm\text{0.65}}$&
$\text{71.69}^{\pm\text{0.24}}$&
$\text{81.16}^{\pm\text{0.39}}$&
$\text{72.91}^{\pm\text{0.45}}$&
$\text{83.02}^{\pm\text{0.86}}$&
$\text{54.44}^{\pm\text{0.28}}$&
$\text{69.95}^{\pm\text{0.28}}$&
$\text{55.70}^{\pm\text{0.40}}$& 
$\text{70.00}^{\pm\text{0.26}}$&
$\text{68.17}$&$\text{79.13}$\\
BEEF &
$\text{87.10}^{\pm\text{1.38}}$&
$\text{93.10}^{\pm\text{1.21}}$&
$\text{72.09}^{\pm\text{0.33}}$&
$\text{81.91}^{\pm\text{0.58}}$&
$\text{71.88}^{\pm\text{0.54}}$&
$\text{81.45}^{\pm\text{0.74}}$&
$\text{61.41}^{\pm\text{0.38}}$&
$\text{71.21}^{\pm\text{0.57}}$&
$\text{58.16}^{\pm\text{0.60}}$& 
$\text{71.16}^{\pm\text{0.82}}$&
$\text{70.13}$&$\text{79.77}$ \\
MORE & 
$\text{89.16}^{\pm\text{0.96}}$&               
$\text{94.23}^{\pm\text{0.82}}$& 
$\text{70.23}^{\pm\text{2.27}}$&
$\text{81.24}^{\pm\text{1.24}}$& 
$\text{70.53}^{\pm\text{1.09}}$&
$\text{81.59}^{\pm\text{0.98}}$& 
$\text{64.97}^{\pm\text{1.28}}$&
$\text{74.03}^{\pm\text{1.61}}$& 
$\text{63.06}^{\pm\text{1.26}}$&
$\text{72.74}^{\pm\text{1.04}}$& 
$\text{71.59}$&$\text{80.77}$               \\  
ROW & 
$\text{90.97}^{\pm\text{0.19}}$&
$\text{94.45}^{\pm\text{0.21}}$& 
$\text{74.72}^{\pm\text{0.48}}$&
$\text{82.87}^{\pm\text{0.41}}$& 
$\text{74.60}^{\pm\text{0.12}}$&
$\text{83.12}^{\pm\text{0.23}}$& 
$\text{65.11}^{\pm\text{1.97}}$&
$\text{74.16}^{\pm\text{1.34}}$& 
$\text{63.21}^{\pm\text{2.53}}$&
$\text{72.91}^{\pm\text{2.12}}$& 
$\text{73.72}$&$\text{81.50}$               \\ 
\hline
\Tstrut
\textbf{TPL (ours)} &  $\textbf{92.33}^{\pm\text{0.32}}$& $\textbf{95.11}^{\pm \text{0.44}}$                   &$\textbf{76.53}^{\pm\text{0.27}}$&$\textbf{84.10}^{\pm\text{0.34}}$                &  $\textbf{76.34}^{\pm\text{0.38}}$&$\textbf{84.46}^{\pm\text{0.28}}$                 &   $\textbf{68.64}^{\pm\text{0.44}}$&$\textbf{76.77}^{\pm\text{0.23}}$                  &     $\textbf{67.20}^{\pm\text{0.51}}$&$\textbf{75.72}^{\pm\text{0.37}}$                 &  $\textbf{76.21}$&$\textbf{83.23}$                   \Bstrut \\
\bottomrule
\rowcolor{pink}
\Tstrut
Non-CL$_{\text{PFI}}$
& $\text{96.90}^{\pm\text{0.07}}$ 
& $\text{97.96}^{\pm\text{0.05}}$ 
& $\text{83.61}^{\pm\text{0.33}}$
& $\text{89.72}^{\pm\text{0.10}}$
& $\text{83.61}^{\pm\text{0.33}}$
&
$\text{88.89}^{\pm\text{0.06}}$
& $\text{85.55}^{\pm\text{0.07}}$
&
$\text{88.26}^{\pm\text{0.08}}$
& $\text{85.71}^{\pm\text{0.14}}$
&
$\text{88.66}^{\pm\text{0.01}}$
& 87.08
& 90.70              \\ 
\rowcolor{pink} TPL$_{\text{PFI}}$  
& $\text{94.86}^{\pm\text{0.02}}$
& $\text{96.89}^{\pm\text{0.02}}$
& $\text{82.43}^{\pm\text{0.12}}$
& $\text{88.28}^{\pm\text{0.17}}$
& $\text{80.86}^{\pm\text{0.07}}$
& $\text{87.32}^{\pm\text{0.07}}$
& $\text{84.06}^{\pm\text{0.11}}$
& $\text{87.19}^{\pm\text{0.11}}$
& $\text{83.87}^{\pm\text{0.07}}$
& $\text{87.40}^{\pm\text{0.16}}$
& 85.22
& 89.42                  \\
\bottomrule
\end{tabular}}
\label{tab.results}
\vspace{-1.em}
\end{table}

Table~\ref{tab.results} shows the CIL accuracy (ACC) results. The last two columns give the row averages. {Our \texttt{TPL} performs the best in both \textit{average incremental ACC} (AIA) and ACC \textit{after the last task} (Last). Based on {AIA}, TPL's forgetting (CF) is almost negligible. When the full ImageNet data is used in pre-training (pink rows), TPL$_{\text{PFI}}$ has \textbf{almost no forgetting} in both AIA and Last ACC.} 


\textbf{Comparison with CIL baselines with pre-training.} {The best-performing replay-based baseline is \texttt{ROW}, which also follows the TIL+OOD paradigm~\citep{kimtheoretical}. 
Since its OOD score is inferior to our principled $S_{\textit{LR}}(\boldsymbol{x})$, \texttt{ROW} is greatly outperformed by \texttt{TPL}. The ACC gap between our \texttt{TPL} and the best exemplar-free method \texttt{PASS} is even greater, 68.25\% (\texttt{PASS}) vs. 76.21\% (\texttt{TPL}) in Last ACC. \texttt{TPL} also markedly outperforms the strong \emph{network expansion} methods \texttt{DER}, \texttt{FOSTER}, and \texttt{BEEF}.} 


\begin{wraptable}{R}{0.55\linewidth}
\centering
\vspace{-1.4em}
\caption{CIL ACC (\%) after learning the final task without pre-training (average over the five datasets used in~\Cref{tab.results}). The detailed results are shown in~\Cref{tab.CIL_non_pretrain} of \Cref{app:cil_resnet}.}
\resizebox{\linewidth}{!}{
\begin{tabular}{cccccccc}  
\toprule
OWM & PASS & EEIL & GD & HAL & A-GEM & HAT & iCaRL\\
\hline
\Tstrut{24.7} & {30.6} & {46.3} & {47.1} & {46.4} & {45.8} & {39.5} & {45.6} \\
\midrule
DER++ & DER &FOSTER &BEEF & MORE &ROW & \textbf{TPL (ours)} & \\
\hline
\Tstrut\text{46.5} & {54.2} & {52.2} & {53.4} & {51.2} & {53.1} &\textbf{57.5} & \\
\bottomrule
\end{tabular}}
\label{tab:non_pretrain_summary}
\vspace{-1.2em}
\end{wraptable}

\textbf{Without pre-training.} The accuracy results after learning the final task without pre-training are given in~\Cref{tab.CIL_non_pretrain} of \Cref{app:cil_resnet}. We provide a summary  in~\Cref{tab:non_pretrain_summary} here. As \texttt{L2P}, \texttt{SLDA}, and \texttt{ADAM} are designed specifically for pre-trained backbones, they cannot be adapted to the non-pre-training setting and thus are excluded here. Similar to the observation in~\Cref{tab.results}, our \texttt{TPL} achieves the overall best results (with ACC of 57.5\%), while \texttt{DER} ranks the second (54.2\%).

\subsection{Ablation Study}
\label{sec:ablation}

\begin{figure}[t]
  \centering
  \begin{subfigure}[b]{0.33\textwidth}
    \centering
    \includegraphics[width=1.0\textwidth]{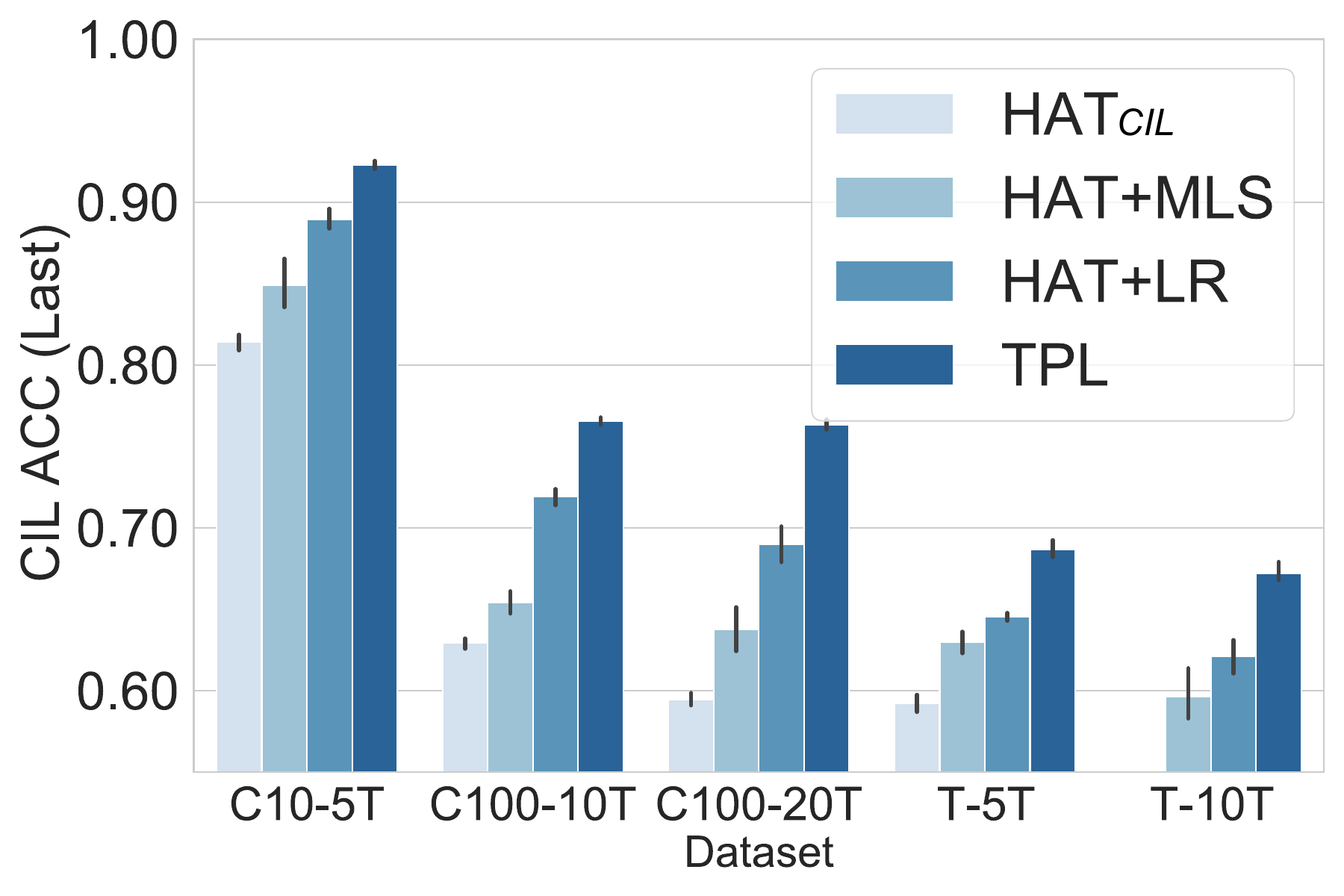}
    \vspace{-1.5em}
    \label{fig:performance_gain}
    \caption{\footnotesize{Performance Gain}}
  \end{subfigure}%
  \hfill
  \begin{subfigure}[b]{0.33\textwidth}
    \centering
    \includegraphics[width=1.0\textwidth]{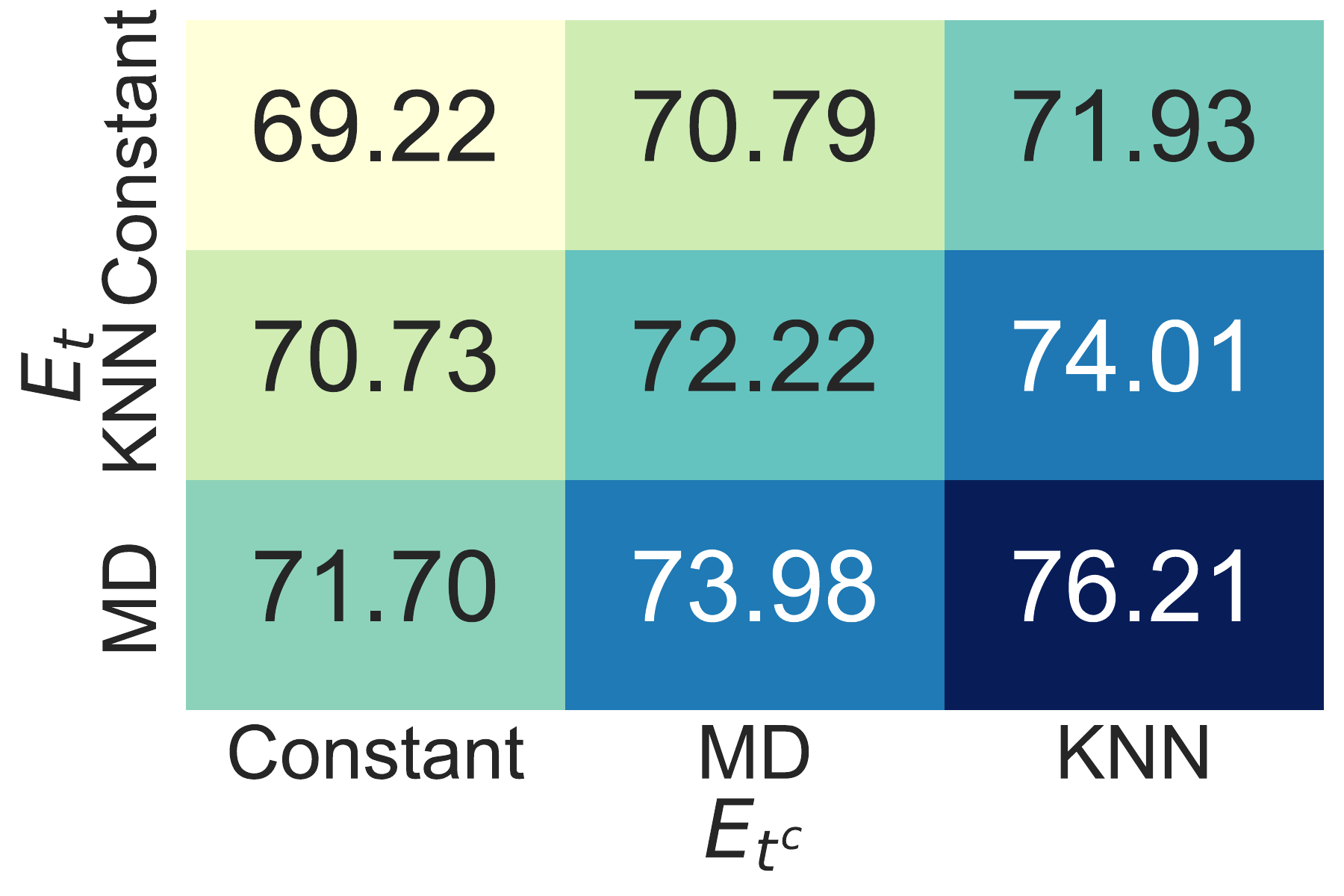}
    \vspace{-1.5em}
    \label{fig:different_ind_ood}
   \caption{\footnotesize{Different $E_{t}$ v.s. $E_{t^c}$}}
  \end{subfigure}%
  \hfill
  \begin{subfigure}[b]{0.33\textwidth}
    \centering
    \includegraphics[width=1.0\textwidth]{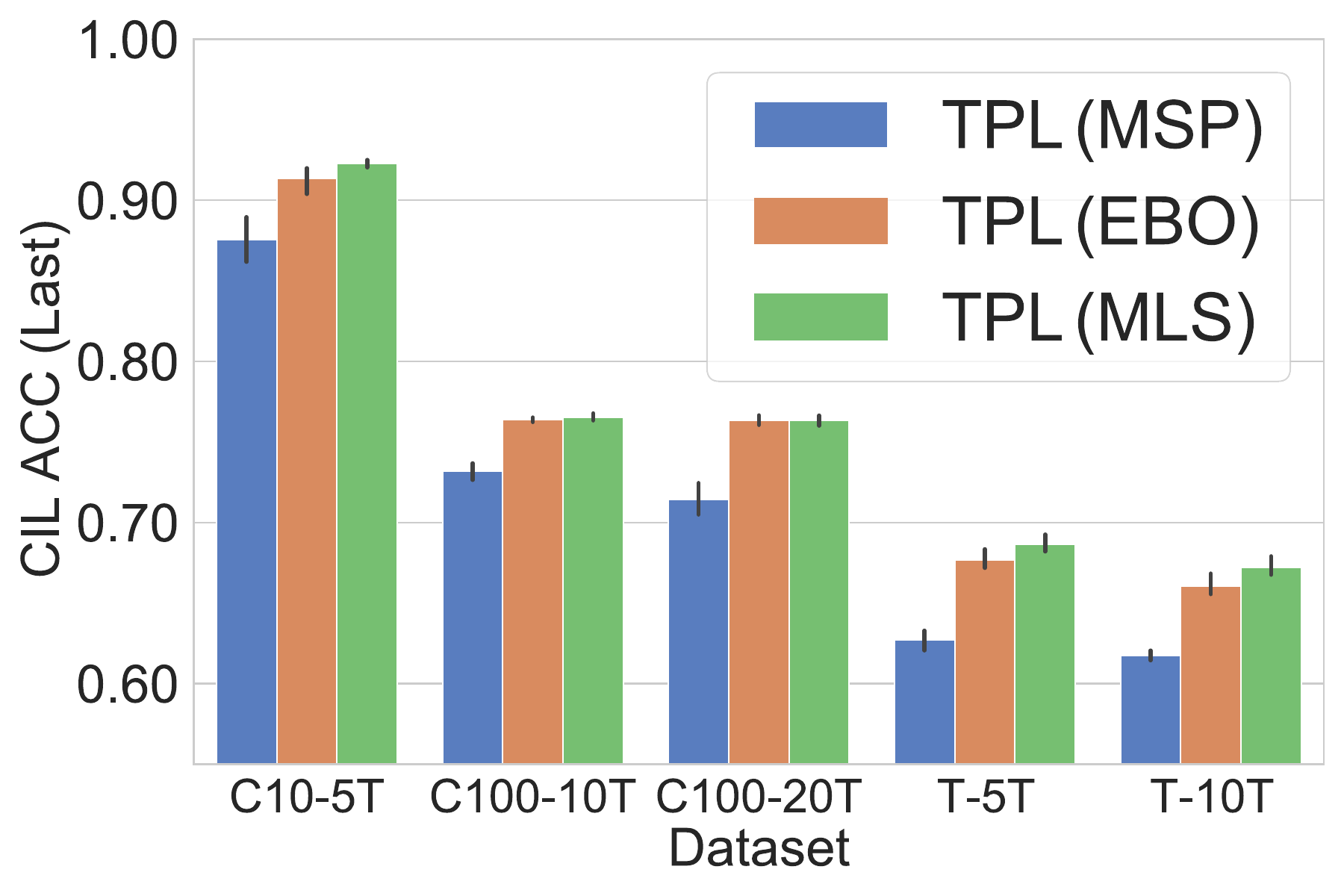}
    \vspace{-1.5em}
    \label{fig:different_e_logit}
    \caption{\footnotesize{Different $E_{\textit{logit}}$}}
  \end{subfigure}
  \caption{Ablation Studies. Fig (a) illustrates the achieved ACC gain for each of the designed techniques on the five datasets; Fig (b) displays the average ACC results obtained from different choices of  $E_{t}$ and $E_{t^c}$ for~\cref{eq:LR}; Fig (c) showcases the results for various selections of $E_{\textit{logit}}$ for \texttt{TPL} in~\cref{eq:final_score}.}
  \label{fig:all}
  \vspace{-0.7em}
\end{figure}

\textbf{Performance gain}.~\Cref{fig:all}(a) shows the performance gain achieved by adding each proposed technique. Starting from vanilla \texttt{HAT}$_{CIL}$ with an average Last ACC of 63.41\% over all datasets, the proposed likelihood ratio \texttt{LR}} score (HAT+LR) boosts the average Last ACC to 71.25\%. 
Utilizing the OOD detection method \texttt{MLS} (HAT+MLS) only improves the ACC to 68.69\%.
The final composition of LR and MLS boosted the performance to 76.21\%. 


\textbf{Different $E_{t}$ v.s. $E_{t^c}$}. Recall that the key insight behind the \texttt{LR} score lies in the estimation of likelihood ratio.~\Cref{fig:all}(b) presents the average Last ACC results across 5 datasets, employing various approaches to estimate $\mathcal P_{t}$ and $\mathcal P_{t^c}$. In this context, the term \emph{Constant} refers to the use of a uniform distribution as the distribution of $\mathcal P_{t^c}$, where the energy function is a constant mapping. Our \texttt{TPL} approach is equivalent to employing ($E_{t}=\texttt{MD}$, $E_{t^c}=\texttt{KNN}$). The results reveal the following: (1) The incorporation of the $\mathcal P_{t^c}$ distribution estimation is beneficial compared to assuming a uniform distribution. (2) As $\mathcal P_{t^c}$ can only be estimated using the replay data, the high-performing \texttt{KNN} method outperforms \texttt{MD}. However, since \texttt{MD} can estimate $\mathcal P_{t}$ without task $t$'s training data during the test phase, it proves to be more effective than \texttt{KNN} when serving as $E_{t}$.

\textbf{Different logit-based scores}. Although $S_{\textit{MLS}}(\boldsymbol x)$ is used as the logit-based score in Section~\ref{sec:composition}, alternative logit-based scores can also be considered. In this study, we conduct experiments using 3 popular logit-based scores \texttt{MSP}~\citep{hendrycks2016baseline}, \texttt{EBO}~\citep{liu2020energy}, and \texttt{MLS} (their definitions are given in~\Cref{app:logit_score_compute}). The results presented in~\Cref{fig:all}(c) indicate that \texttt{EBO} and \texttt{MLS} yield comparable results, with average Last ACC of 75.76\%, and 76.21\% respectively, while \texttt{MSP} has inferior performance with average Last ACC of 71.32\%. 

\begin{wraptable}{R}{0.50\linewidth}
\centering
\vspace{-1.1em}
\caption{ACC (\%) after learning the final task (Last) with smaller replay buffer sizes (average over the five datasets in~\Cref{tab.results}). The detailed results are shown in~\Cref{tab.small_replay_full} of \Cref{sec.small_replay}. The replay buffer size is set as 100 for CIFAR-10, and 1000 for CIFAR-100 and TinyImageNet.}
\resizebox{\linewidth}{!}{
\begin{tabular}{cccccc}  
\toprule
iCaRL & A-GEM & EEIL & GD & DER++ & HAL \\ \hline
\Tstrut 63.60 & 31.15 & 58.24 & 54.39 & 62.16 & 60.21 \\
\midrule
DER & FOSTER & BEEF & MORE & ROW & TPL \\ \hline
\Tstrut
 68.32 & 66.86 & 68.94 & 71.44 & 72.70 & \textbf{75.56} \\
\bottomrule
\end{tabular}}
\label{tab:small_replay_summary}
\vspace{-1.2em}
\end{wraptable}

\textbf{Smaller replay buffer sizes}. The accuracy after learning the final task with smaller replay buffer sizes are given in~\Cref{tab.small_replay_full} of \Cref{sec.small_replay}. We provide a summary as~\Cref{tab:small_replay_summary}, which shows that when using a smaller replay buffer, the performance drop of \texttt{TPL} is small. The goal of using the replay data in \texttt{TPL} is to compute the likelihood ratio (LR) score, while traditional replay methods focus on preventing forgetting (CF). Note that CF is already addressed by the TIL method \texttt{HAT} in our case. Thus our method \texttt{TPL} is robust with fewer replay samples.

{
\textbf{More OOD methods.} To understand the effect of OOD detection on CIL, we applied 20 OOD detection methods to CIL and drew some interesting conclusions (see~\Cref{app.ood+til}). (1) There exists a linear relationship between OOD detection AUC and CIL ACC performances. (2) Different OOD detection methods result in similar TIL (task-incremental learning) ACC when applying HAT.

\textbf{More pre-trained models (visual encoders).} We also study TPL with different pre-trained models in~\Cref{app.visual_encoder} (MAE, Dino, ViT and DeiT of different sizes). We found the pre-trained models based on supervised learning outperform self-supervised models in both CIL and TIL. 
}

\section{Conclusion}
\label{conclusion}
\vspace{-1mm}
{In this paper, we developed a novel approach for class incremental learning (CIL) via task-id prediction based on likelihood ratio. Recent studies~\citep{kim2022multi,kimtheoretical,kim2023learnability} suggested that OOD detection methods can be applied to perform task-id prediction in CIL and thus achieve the \emph{state-of-the-art} performance. However, we argue that traditional OOD detection is not optimal for CIL as additional information in CIL can be leveraged to design a better and principled method for task-id prediction. 
Our experimental results show that our \texttt{TPL} outperforms strong baselines and has almost negligible catastrophic forgetting. 
Limitations of our approach are discussed in \Cref{sec.limitations}.}

\section*{Acknowledgements}

We sincerely thank Baizhou Huang of Peking University, Shanda Li of Carnegie Mellon University, and the anonymous reviewers of ICLR 2024 for providing valuable
suggestions on this work.

\section*{Ethics Statement}

Since this research involves only classification learning using existing datasets downloaded from the public domain and our algorithms are not for any specific application but for solving the general problem of continual learning, we do not feel there are any possible ethical issues in this research. 

\section*{Reproducibility Statement}

The source code of \texttt{TPL} has been public at \url{https://github.com/linhaowei1/TPL}. The proofs of~\Cref{theorem:lr,theorem:max_auc} are provided in~\Cref{proof1}. The training details and dataset details are given in Sec.~\ref{sec:exp setup} and~\Cref{app:training_details}.
 
\bibliography{iclr2024_conference}

\begin{thebibliography}{88}
\providecommand{\natexlab}[1]{#1}
\providecommand{\url}[1]{\texttt{#1}}
\expandafter\ifx\csname urlstyle\endcsname\relax
  \providecommand{\doi}[1]{doi: #1}\else
  \providecommand{\doi}{doi: \begingroup \urlstyle{rm}\Url}\fi

\bibitem[Abati et~al.(2020)Abati, Tomczak, Blankevoort, Calderara, Cucchiara,
  and Bejnordi]{abati2020conditional}
Davide Abati, Jakub Tomczak, Tijmen Blankevoort, Simone Calderara, Rita
  Cucchiara, and Babak~Ehteshami Bejnordi.
\newblock Conditional channel gated networks for task-aware continual learning.
\newblock In \emph{Proceedings of the IEEE/CVF Conference on Computer Vision
  and Pattern Recognition}, pp.\  3931--3940, 2020.

\bibitem[Ahn et~al.(2021)Ahn, Kwak, Lim, Bang, Kim, and Moon]{ahn2021ss}
Hongjoon Ahn, Jihwan Kwak, Subin Lim, Hyeonsu Bang, Hyojun Kim, and Taesup
  Moon.
\newblock Ss-il: Separated softmax for incremental learning.
\newblock In \emph{Proceedings of the IEEE/CVF International conference on
  computer vision}, pp.\  844--853, 2021.

\bibitem[Aljundi et~al.(2017)Aljundi, Chakravarty, and
  Tuytelaars]{aljundi2017expert}
Rahaf Aljundi, Punarjay Chakravarty, and Tinne Tuytelaars.
\newblock Expert gate: Lifelong learning with a network of experts.
\newblock In \emph{Proceedings of the IEEE Conference on Computer Vision and
  Pattern Recognition}, pp.\  3366--3375, 2017.

\bibitem[Bang et~al.(2022)Bang, Koh, Park, Song, Ha, and Choi]{bang2022online}
Jihwan Bang, Hyunseo Koh, Seulki Park, Hwanjun Song, Jung-Woo Ha, and Jonghyun
  Choi.
\newblock Online continual learning on a contaminated data stream with blurry
  task boundaries.
\newblock In \emph{Proceedings of the IEEE/CVF Conference on Computer Vision
  and Pattern Recognition}, pp.\  9275--9284, 2022.

\bibitem[Bendale \& Boult(2016)Bendale and Boult]{bendale2016towards}
Abhijit Bendale and Terrance~E Boult.
\newblock Towards open set deep networks.
\newblock In \emph{Proceedings of the IEEE conference on computer vision and
  pattern recognition}, pp.\  1563--1572, 2016.

\bibitem[Bhat et~al.(2022)Bhat, Zonooz, and Arani]{Bhat2022ConsistencyIT}
Prashant Bhat, Bahram Zonooz, and E.~Arani.
\newblock Consistency is the key to further mitigating catastrophic forgetting
  in continual learning.
\newblock In \emph{CoLLAs}, 2022.
\newblock URL \url{https://api.semanticscholar.org/CorpusID:250425816}.

\bibitem[Buzzega et~al.(2020)Buzzega, Boschini, Porrello, Abati, and
  Calderara]{buzzega2020dark}
Pietro Buzzega, Matteo Boschini, Angelo Porrello, Davide Abati, and Simone
  Calderara.
\newblock Dark experience for general continual learning: a strong, simple
  baseline.
\newblock \emph{Advances in neural information processing systems},
  33:\penalty0 15920--15930, 2020.

\bibitem[Caron et~al.(2021)Caron, Touvron, Misra, J\'egou, Mairal, Bojanowski,
  and Joulin]{caron2021emerging}
Mathilde Caron, Hugo Touvron, Ishan Misra, Herv\'e J\'egou, Julien Mairal,
  Piotr Bojanowski, and Armand Joulin.
\newblock Emerging properties in self-supervised vision transformers.
\newblock In \emph{Proceedings of the International Conference on Computer
  Vision (ICCV)}, 2021.

\bibitem[Castro et~al.(2018)Castro, Mar{\'\i}n-Jim{\'e}nez, Guil, Schmid, and
  Alahari]{castro2018end}
Francisco~M Castro, Manuel~J Mar{\'\i}n-Jim{\'e}nez, Nicol{\'a}s Guil, Cordelia
  Schmid, and Karteek Alahari.
\newblock End-to-end incremental learning.
\newblock In \emph{Proceedings of the European conference on computer vision
  (ECCV)}, pp.\  233--248, 2018.

\bibitem[Chaudhry et~al.(2018)Chaudhry, Ranzato, Rohrbach, and
  Elhoseiny]{chaudhry2018efficient}
Arslan Chaudhry, Marc'Aurelio Ranzato, Marcus Rohrbach, and Mohamed Elhoseiny.
\newblock Efficient lifelong learning with a-gem.
\newblock \emph{arXiv preprint arXiv:1812.00420}, 2018.

\bibitem[Chaudhry et~al.(2021)Chaudhry, Gordo, Dokania, Torr, and
  Lopez-Paz]{chaudhry2021using}
Arslan Chaudhry, Albert Gordo, Puneet Dokania, Philip Torr, and David
  Lopez-Paz.
\newblock Using hindsight to anchor past knowledge in continual learning.
\newblock In \emph{Proceedings of the AAAI Conference on Artificial
  Intelligence}, volume~35, pp.\  6993--7001, 2021.

\bibitem[De~Lange et~al.(2021)De~Lange, Aljundi, Masana, Parisot, Jia,
  Leonardis, Slabaugh, and Tuytelaars]{de2021continual}
Matthias De~Lange, Rahaf Aljundi, Marc Masana, Sarah Parisot, Xu~Jia,
  Ale{\v{s}} Leonardis, Gregory Slabaugh, and Tinne Tuytelaars.
\newblock A continual learning survey: Defying forgetting in classification
  tasks.
\newblock \emph{IEEE transactions on pattern analysis and machine
  intelligence}, 44\penalty0 (7):\penalty0 3366--3385, 2021.

\bibitem[Du et~al.(2021)Du, Wang, Cai, and Li]{du2021vos}
Xuefeng Du, Zhaoning Wang, Mu~Cai, and Yixuan Li.
\newblock Vos: Learning what you don't know by virtual outlier synthesis.
\newblock In \emph{International Conference on Learning Representations}, 2021.

\bibitem[Du et~al.(2020)Du, Li, and Mordatch]{du2020compositional}
Yilun Du, Shuang Li, and Igor Mordatch.
\newblock Compositional visual generation with energy based models.
\newblock \emph{Advances in Neural Information Processing Systems},
  33:\penalty0 6637--6647, 2020.

\bibitem[Gal \& Ghahramani(2016)Gal and Ghahramani]{gal2016dropout}
Yarin Gal and Zoubin Ghahramani.
\newblock Dropout as a bayesian approximation: Representing model uncertainty
  in deep learning.
\newblock In \emph{international conference on machine learning}, pp.\
  1050--1059. PMLR, 2016.

\bibitem[Guo et~al.(2022)Guo, Liu, and Zhao]{guo2022online}
Yiduo Guo, Bing Liu, and Dongyan Zhao.
\newblock Online continual learning through mutual information maximization.
\newblock In \emph{International Conference on Machine Learning}, pp.\
  8109--8126. PMLR, 2022.

\bibitem[Guo et~al.(2023)Guo, Liu, and Zhao]{guo2023dealing}
Yiduo Guo, Bing Liu, and Dongyan Zhao.
\newblock Dealing with cross-task class discrimination in online continual
  learning.
\newblock In \emph{Proceedings of the IEEE/CVF Conference on Computer Vision
  and Pattern Recognition}, pp.\  11878--11887, 2023.

\bibitem[Hadsell et~al.(2020)Hadsell, Rao, Rusu, and
  Pascanu]{hadsell2020embracing}
Raia Hadsell, Dushyant Rao, Andrei~A Rusu, and Razvan Pascanu.
\newblock Embracing change: Continual learning in deep neural networks.
\newblock \emph{Trends in cognitive sciences}, 24\penalty0 (12):\penalty0
  1028--1040, 2020.

\bibitem[Hayes \& Kanan(2020)Hayes and Kanan]{hayes2020lifelong}
Tyler~L Hayes and Christopher Kanan.
\newblock Lifelong machine learning with deep streaming linear discriminant
  analysis.
\newblock In \emph{Proceedings of the IEEE/CVF conference on computer vision
  and pattern recognition workshops}, pp.\  220--221, 2020.

\bibitem[He et~al.(2016)He, Zhang, Ren, and Sun]{he2016deep}
Kaiming He, Xiangyu Zhang, Shaoqing Ren, and Jian Sun.
\newblock Deep residual learning for image recognition.
\newblock In \emph{Proceedings of the IEEE conference on computer vision and
  pattern recognition}, pp.\  770--778, 2016.

\bibitem[He et~al.(2021)He, Chen, Xie, Li, Doll{\'{a}}r, and Girshick]{mae}
Kaiming He, Xinlei Chen, Saining Xie, Yanghao Li, Piotr Doll{\'{a}}r, and
  Ross~B. Girshick.
\newblock Masked autoencoders are scalable vision learners.
\newblock \emph{CoRR}, abs/2111.06377, 2021.
\newblock URL \url{https://arxiv.org/abs/2111.06377}.

\bibitem[Hendrycks \& Gimpel(2016)Hendrycks and Gimpel]{hendrycks2016baseline}
Dan Hendrycks and Kevin Gimpel.
\newblock A baseline for detecting misclassified and out-of-distribution
  examples in neural networks.
\newblock \emph{arXiv preprint arXiv:1610.02136}, 2016.

\bibitem[Hendrycks et~al.(2018)Hendrycks, Mazeika, and
  Dietterich]{hendrycks2018deep}
Dan Hendrycks, Mantas Mazeika, and Thomas Dietterich.
\newblock Deep anomaly detection with outlier exposure.
\newblock \emph{arXiv preprint arXiv:1812.04606}, 2018.

\bibitem[Hendrycks et~al.(2019)Hendrycks, Basart, Mazeika, Mostajabi,
  Steinhardt, and Song]{hendrycks2019scaling}
Dan Hendrycks, Steven Basart, Mantas Mazeika, Mohammadreza Mostajabi, Jacob
  Steinhardt, and Dawn Song.
\newblock Scaling out-of-distribution detection for real-world settings.
\newblock \emph{arXiv preprint arXiv:1911.11132}, 2019.

\bibitem[Henning et~al.(2021)Henning, Cervera, D'Angelo, Von~Oswald, Traber,
  Ehret, Kobayashi, Grewe, and Sacramento]{henning2021posterior}
Christian Henning, Maria Cervera, Francesco D'Angelo, Johannes Von~Oswald,
  Regina Traber, Benjamin Ehret, Seijin Kobayashi, Benjamin~F Grewe, and
  Jo{\~a}o Sacramento.
\newblock Posterior meta-replay for continual learning.
\newblock \emph{Advances in Neural Information Processing Systems},
  34:\penalty0 14135--14149, 2021.

\bibitem[Houlsby et~al.(2019)Houlsby, Giurgiu, Jastrzebski, Morrone,
  De~Laroussilhe, Gesmundo, Attariyan, and Gelly]{houlsby2019parameter}
Neil Houlsby, Andrei Giurgiu, Stanislaw Jastrzebski, Bruna Morrone, Quentin
  De~Laroussilhe, Andrea Gesmundo, Mona Attariyan, and Sylvain Gelly.
\newblock Parameter-efficient transfer learning for nlp.
\newblock In \emph{International Conference on Machine Learning}, pp.\
  2790--2799. PMLR, 2019.

\bibitem[Huang et~al.(2021)Huang, Geng, and Li]{huang2021importance}
Rui Huang, Andrew Geng, and Yixuan Li.
\newblock On the importance of gradients for detecting distributional shifts in
  the wild.
\newblock \emph{Advances in Neural Information Processing Systems},
  34:\penalty0 677--689, 2021.

\bibitem[Jeeveswaran et~al.(2023)Jeeveswaran, Bhat, Zonooz, and
  Arani]{Jeeveswaran2023BiRTBR}
Kishaan Jeeveswaran, Prashant Bhat, Bahram Zonooz, and E.~Arani.
\newblock Birt: Bio-inspired replay in vision transformers for continual
  learning.
\newblock \emph{ArXiv}, abs/2305.04769, 2023.
\newblock URL \url{https://api.semanticscholar.org/CorpusID:258557568}.

\bibitem[Kang et~al.(2022)Kang, Park, and Han]{kang2022class}
Minsoo Kang, Jaeyoo Park, and Bohyung Han.
\newblock Class-incremental learning by knowledge distillation with adaptive
  feature consolidation.
\newblock In \emph{Proceedings of the IEEE/CVF conference on computer vision
  and pattern recognition}, pp.\  16071--16080, 2022.

\bibitem[Ke \& Liu(2022)Ke and Liu]{ke2022continualsurvey}
Zixuan Ke and Bing Liu.
\newblock Continual learning of natural language processing tasks: A survey.
\newblock \emph{arXiv preprint arXiv:2211.12701}, 2022.

\bibitem[Ke et~al.(2021{\natexlab{a}})Ke, Liu, Ma, Xu, and
  Shu]{ke2021achieving}
Zixuan Ke, Bing Liu, Nianzu Ma, Hu~Xu, and Lei Shu.
\newblock Achieving forgetting prevention and knowledge transfer in continual
  learning.
\newblock \emph{Advances in Neural Information Processing Systems},
  34:\penalty0 22443--22456, 2021{\natexlab{a}}.

\bibitem[Ke et~al.(2021{\natexlab{b}})Ke, Xu, and Liu]{ke2021adapting}
Zixuan Ke, Hu~Xu, and Bing Liu.
\newblock Adapting bert for continual learning of a sequence of aspect
  sentiment classification tasks.
\newblock In \emph{Proceedings of the 2021 Conference of the North American
  Chapter of the Association for Computational Linguistics: Human Language
  Technologies}, pp.\  4746--4755, 2021{\natexlab{b}}.

\bibitem[Ke et~al.(2023)Ke, Shao, Lin, Konishi, Kim, and Liu]{ke2023continual}
Zixuan Ke, Yijia Shao, Haowei Lin, Tatsuya Konishi, Gyuhak Kim, and Bing Liu.
\newblock Continual pre-training of language models.
\newblock In \emph{The Eleventh International Conference on Learning
  Representations (ICLR-2023)}, 2023.

\bibitem[Kemker \& Kanan(2017)Kemker and Kanan]{kemker2017fearnet}
Ronald Kemker and Christopher Kanan.
\newblock Fearnet: Brain-inspired model for incremental learning.
\newblock \emph{arXiv preprint arXiv:1711.10563}, 2017.

\bibitem[Kim et~al.(2022{\natexlab{a}})Kim, Liu, and Ke]{kim2022multi}
Gyuhak Kim, Bing Liu, and Zixuan Ke.
\newblock A multi-head model for continual learning via out-of-distribution
  replay.
\newblock In \emph{Conference on Lifelong Learning Agents}, pp.\  548--563.
  PMLR, 2022{\natexlab{a}}.

\bibitem[Kim et~al.(2022{\natexlab{b}})Kim, Xiao, Konishi, Ke, and
  Liu]{kimtheoretical}
Gyuhak Kim, Changnan Xiao, Tatsuya Konishi, Zixuan Ke, and Bing Liu.
\newblock A theoretical study on solving continual learning.
\newblock In \emph{Advances in Neural Information Processing Systems},
  2022{\natexlab{b}}.

\bibitem[Kim et~al.(2023)Kim, Xiao, Konishi, and Liu]{kim2023learnability}
Gyuhak Kim, Changnan Xiao, Tatsuya Konishi, and Bing Liu.
\newblock Learnability and algorithm for continual learning.
\newblock \emph{ICML-2023}, 2023.

\bibitem[Kirkpatrick et~al.(2017)Kirkpatrick, Pascanu, Rabinowitz, Veness,
  Desjardins, Rusu, Milan, Quan, Ramalho, Grabska-Barwinska,
  et~al.]{kirkpatrick2017overcoming}
James Kirkpatrick, Razvan Pascanu, Neil Rabinowitz, Joel Veness, Guillaume
  Desjardins, Andrei~A Rusu, Kieran Milan, John Quan, Tiago Ramalho, Agnieszka
  Grabska-Barwinska, et~al.
\newblock Overcoming catastrophic forgetting in neural networks.
\newblock \emph{Proceedings of the national academy of sciences}, 114\penalty0
  (13):\penalty0 3521--3526, 2017.

\bibitem[Krizhevsky \& Hinton(2010)Krizhevsky and
  Hinton]{krizhevsky2010convolutional}
Alex Krizhevsky and Geoff Hinton.
\newblock Convolutional deep belief networks on cifar-10.
\newblock \emph{Unpublished manuscript}, 40\penalty0 (7):\penalty0 1--9, 2010.

\bibitem[Krizhevsky et~al.(2009)Krizhevsky, Hinton,
  et~al.]{krizhevsky2009learning}
Alex Krizhevsky, Geoffrey Hinton, et~al.
\newblock Learning multiple layers of features from tiny images.
\newblock 2009.

\bibitem[Le \& Yang(2015)Le and Yang]{le2015tiny}
Ya~Le and Xuan Yang.
\newblock Tiny imagenet visual recognition challenge.
\newblock \emph{CS 231N}, 7:\penalty0 7, 2015.

\bibitem[Lee et~al.(2019)Lee, Lee, Shin, and Lee]{lee2019overcoming}
Kibok Lee, Kimin Lee, Jinwoo Shin, and Honglak Lee.
\newblock Overcoming catastrophic forgetting with unlabeled data in the wild.
\newblock In \emph{Proceedings of the IEEE/CVF International Conference on
  Computer Vision}, pp.\  312--321, 2019.

\bibitem[Lee et~al.(2018{\natexlab{a}})Lee, Lee, Lee, and
  Shin]{lee2018training}
Kimin Lee, Honglak Lee, Kibok Lee, and Jinwoo Shin.
\newblock Training confidence-calibrated classifiers for detecting
  out-of-distribution samples.
\newblock In \emph{International Conference on Learning Representations},
  2018{\natexlab{a}}.

\bibitem[Lee et~al.(2018{\natexlab{b}})Lee, Lee, Lee, and Shin]{lee2018simple}
Kimin Lee, Kibok Lee, Honglak Lee, and Jinwoo Shin.
\newblock A simple unified framework for detecting out-of-distribution samples
  and adversarial attacks.
\newblock \emph{Advances in neural information processing systems}, 31,
  2018{\natexlab{b}}.

\bibitem[Li et~al.(2022)Li, Zhai, Chen, Gao, Zhang, and Zhang]{li2022continual}
Guodun Li, Yuchen Zhai, Qianglong Chen, Xing Gao, Ji~Zhang, and Yin Zhang.
\newblock Continual few-shot intent detection.
\newblock In \emph{Proceedings of the 29th International Conference on
  Computational Linguistics}, pp.\  333--343, 2022.

\bibitem[Liang et~al.(2017)Liang, Li, and Srikant]{liang2017enhancing}
Shiyu Liang, Yixuan Li, and Rayadurgam Srikant.
\newblock Enhancing the reliability of out-of-distribution image detection in
  neural networks.
\newblock \emph{arXiv preprint arXiv:1706.02690}, 2017.

\bibitem[Lin \& Gu(2023)Lin and Gu]{Lin2023FLatSPO}
Haowei Lin and Yuntian Gu.
\newblock Flats: Principled out-of-distribution detection with feature-based
  likelihood ratio score.
\newblock \emph{ArXiv}, abs/2310.05083, 2023.
\newblock URL \url{https://api.semanticscholar.org/CorpusID:263831173}.

\bibitem[Liu et~al.(2020{\natexlab{a}})Liu, Wang, Owens, and Li]{liu2020energy}
Weitang Liu, Xiaoyun Wang, John Owens, and Yixuan Li.
\newblock Energy-based out-of-distribution detection.
\newblock \emph{Advances in Neural Information Processing Systems},
  33:\penalty0 21464--21475, 2020{\natexlab{a}}.

\bibitem[Liu et~al.(2020{\natexlab{b}})Liu, Liu, Su, Schiele, and
  Sun]{2020Mnemonics}
Y.~Liu, A.~A. Liu, Y.~Su, B.~Schiele, and Q.~Sun.
\newblock Mnemonics training: Multi-class incremental learning without
  forgetting.
\newblock In \emph{2020 IEEE/CVF Conference on Computer Vision and Pattern
  Recognition (CVPR)}, 2020{\natexlab{b}}.

\bibitem[Lopez-Paz \& Ranzato(2017)Lopez-Paz and Ranzato]{lopez2017gradient}
David Lopez-Paz and Marc'Aurelio Ranzato.
\newblock Gradient episodic memory for continual learning.
\newblock \emph{Advances in neural information processing systems}, 30, 2017.

\bibitem[Mallya \& Lazebnik(2018)Mallya and Lazebnik]{mallya2018packnet}
Arun Mallya and Svetlana Lazebnik.
\newblock Packnet: Adding multiple tasks to a single network by iterative
  pruning.
\newblock In \emph{Proceedings of the IEEE Conference on Computer Vision and
  Pattern Recognition}, pp.\  7765--7773, 2018.

\bibitem[Masana et~al.(2021)Masana, Tuytelaars, and Van~de
  Weijer]{masana2021ternary}
Marc Masana, Tinne Tuytelaars, and Joost Van~de Weijer.
\newblock Ternary feature masks: zero-forgetting for task-incremental learning.
\newblock In \emph{Proceedings of the IEEE/CVF conference on computer vision
  and pattern recognition}, pp.\  3570--3579, 2021.

\bibitem[McCloskey \& Cohen(1989)McCloskey and Cohen]{McCloskey1989}
Michael McCloskey and Neal~J Cohen.
\newblock {Catastrophic interference in connectionist networks: The sequential
  learning problem}.
\newblock In \emph{Psychology of learning and motivation}, volume~24, pp.\
  109--165. Elsevier, 1989.

\bibitem[Nalisnick et~al.(2018)Nalisnick, Matsukawa, Teh, Gorur, and
  Lakshminarayanan]{nalisnick2018deep}
Eric Nalisnick, Akihiro Matsukawa, Yee~Whye Teh, Dilan Gorur, and Balaji
  Lakshminarayanan.
\newblock Do deep generative models know what they don't know?
\newblock \emph{arXiv preprint arXiv:1810.09136}, 2018.

\bibitem[Ndiour et~al.(2020)Ndiour, Ahuja, and Tickoo]{ndiour2020out}
Ibrahima Ndiour, Nilesh Ahuja, and Omesh Tickoo.
\newblock Out-of-distribution detection with subspace techniques and
  probabilistic modeling of features.
\newblock \emph{arXiv preprint arXiv:2012.04250}, 2020.

\bibitem[Neyman \& Pearson(1933)Neyman and Pearson]{neyman1933ix}
Jerzy Neyman and Egon~Sharpe Pearson.
\newblock Ix. on the problem of the most efficient tests of statistical
  hypotheses.
\newblock \emph{Philosophical Transactions of the Royal Society of London.
  Series A, Containing Papers of a Mathematical or Physical Character},
  231\penalty0 (694-706):\penalty0 289--337, 1933.

\bibitem[Rajasegaran et~al.(2020)Rajasegaran, Khan, Hayat, Khan, and
  Shah]{rajasegaran2020itaml}
Jathushan Rajasegaran, Salman Khan, Munawar Hayat, Fahad~Shahbaz Khan, and
  Mubarak Shah.
\newblock itaml: An incremental task-agnostic meta-learning approach.
\newblock In \emph{Proceedings of the IEEE/CVF Conference on Computer Vision
  and Pattern Recognition}, pp.\  13588--13597, 2020.

\bibitem[Rebuffi et~al.(2017)Rebuffi, Kolesnikov, Sperl, and
  Lampert]{rebuffi2017icarl}
Sylvestre-Alvise Rebuffi, Alexander Kolesnikov, Georg Sperl, and Christoph~H
  Lampert.
\newblock icarl: Incremental classifier and representation learning.
\newblock In \emph{Proceedings of the IEEE conference on Computer Vision and
  Pattern Recognition}, pp.\  2001--2010, 2017.

\bibitem[Russakovsky et~al.(2015)Russakovsky, Deng, Su, Krause, Satheesh, Ma,
  Huang, Karpathy, Khosla, Bernstein, Berg, and Fei-Fei]{imagenet15russakovsky}
Olga Russakovsky, Jia Deng, Hao Su, Jonathan Krause, Sanjeev Satheesh, Sean Ma,
  Zhiheng Huang, Andrej Karpathy, Aditya Khosla, Michael Bernstein,
  Alexander~C. Berg, and Li~Fei-Fei.
\newblock {ImageNet Large Scale Visual Recognition Challenge}.
\newblock \emph{International Journal of Computer Vision (IJCV)}, 115\penalty0
  (3):\penalty0 211--252, 2015.
\newblock \doi{10.1007/s11263-015-0816-y}.

\bibitem[Rusu et~al.(2016)Rusu, Rabinowitz, Desjardins, Soyer, Kirkpatrick,
  Kavukcuoglu, Pascanu, and Hadsell]{rusu2016progressive}
Andrei~A Rusu, Neil~C Rabinowitz, Guillaume Desjardins, Hubert Soyer, James
  Kirkpatrick, Koray Kavukcuoglu, Razvan Pascanu, and Raia Hadsell.
\newblock Progressive neural networks.
\newblock \emph{arXiv preprint arXiv:1606.04671}, 2016.

\bibitem[Serra et~al.(2018)Serra, Suris, Miron, and
  Karatzoglou]{serra2018overcoming}
Joan Serra, Didac Suris, Marius Miron, and Alexandros Karatzoglou.
\newblock Overcoming catastrophic forgetting with hard attention to the task.
\newblock In \emph{International Conference on Machine Learning}, pp.\
  4548--4557. PMLR, 2018.

\bibitem[Shao et~al.(2023)Shao, Guo, Zhao, and Liu]{shao2023class}
Yijia Shao, Yiduo Guo, Dongyan Zhao, and Bing Liu.
\newblock Class-incremental learning based on label generation.
\newblock \emph{arXiv preprint arXiv:2306.12619}, 2023.

\bibitem[Sun et~al.(2021)Sun, Guo, and Li]{sun2021react}
Yiyou Sun, Chuan Guo, and Yixuan Li.
\newblock React: Out-of-distribution detection with rectified activations.
\newblock \emph{Advances in Neural Information Processing Systems},
  34:\penalty0 144--157, 2021.

\bibitem[Sun et~al.(2022)Sun, Ming, Zhu, and Li]{sun2022out}
Yiyou Sun, Yifei Ming, Xiaojin Zhu, and Yixuan Li.
\newblock Out-of-distribution detection with deep nearest neighbors.
\newblock \emph{arXiv preprint arXiv:2204.06507}, 2022.

\bibitem[Tack et~al.(2020)Tack, Mo, Jeong, and Shin]{tack2020csi}
Jihoon Tack, Sangwoo Mo, Jongheon Jeong, and Jinwoo Shin.
\newblock Csi: Novelty detection via contrastive learning on distributionally
  shifted instances.
\newblock \emph{Advances in neural information processing systems},
  33:\penalty0 11839--11852, 2020.

\bibitem[Terrell \& Scott(1992)Terrell and Scott]{terrell1992variable}
George~R Terrell and David~W Scott.
\newblock Variable kernel density estimation.
\newblock \emph{The Annals of Statistics}, pp.\  1236--1265, 1992.

\bibitem[Thulasidasan et~al.(2019)Thulasidasan, Chennupati, Bilmes,
  Bhattacharya, and Michalak]{thulasidasan2019mixup}
Sunil Thulasidasan, Gopinath Chennupati, Jeff~A Bilmes, Tanmoy Bhattacharya,
  and Sarah Michalak.
\newblock On mixup training: Improved calibration and predictive uncertainty
  for deep neural networks.
\newblock \emph{Advances in Neural Information Processing Systems}, 32, 2019.

\bibitem[Touvron et~al.(2021)Touvron, Cord, Douze, Massa, Sablayrolles, and
  J{\'e}gou]{touvron2021training}
Hugo Touvron, Matthieu Cord, Matthijs Douze, Francisco Massa, Alexandre
  Sablayrolles, and Herv{\'e} J{\'e}gou.
\newblock Training data-efficient image transformers \& distillation through
  attention.
\newblock In \emph{International Conference on Machine Learning}, pp.\
  10347--10357. PMLR, 2021.

\bibitem[Vaze et~al.(2022)Vaze, Han, Vedaldi, and Zisserman]{vaze2022open}
S~Vaze, K~Han, A~Vedaldi, and A~Zisserman.
\newblock Open-set recognition: A good closed-set classifier is all you need?
\newblock In \emph{International Conference on Learning Representations
  (ICLR)}, 2022.

\bibitem[Von~Oswald et~al.(2019)Von~Oswald, Henning, Grewe, and
  Sacramento]{von2019continual}
Johannes Von~Oswald, Christian Henning, Benjamin~F Grewe, and Jo{\~a}o
  Sacramento.
\newblock Continual learning with hypernetworks.
\newblock \emph{arXiv preprint arXiv:1906.00695}, 2019.

\bibitem[Wang et~al.(2022{\natexlab{a}})Wang, Zhou, Liu, Ye, Bian, Zhan, and
  Zhao]{wang2022beef}
Fu-Yun Wang, Da-Wei Zhou, Liu Liu, Han-Jia Ye, Yatao Bian, De-Chuan Zhan, and
  Peilin Zhao.
\newblock Beef: Bi-compatible class-incremental learning via energy-based
  expansion and fusion.
\newblock In \emph{The Eleventh International Conference on Learning
  Representations}, 2022{\natexlab{a}}.

\bibitem[Wang et~al.(2022{\natexlab{b}})Wang, Zhou, Ye, and
  Zhan]{wang2022foster}
Fu-Yun Wang, Da-Wei Zhou, Han-Jia Ye, and De-Chuan Zhan.
\newblock Foster: Feature boosting and compression for class-incremental
  learning.
\newblock In \emph{European conference on computer vision}, pp.\  398--414.
  Springer, 2022{\natexlab{b}}.

\bibitem[Wang et~al.(2022{\natexlab{c}})Wang, Li, Feng, and Zhang]{wang2022vim}
Haoqi Wang, Zhizhong Li, Litong Feng, and Wayne Zhang.
\newblock Vim: Out-of-distribution with virtual-logit matching.
\newblock In \emph{Proceedings of the IEEE/CVF Conference on Computer Vision
  and Pattern Recognition}, pp.\  4921--4930, 2022{\natexlab{c}}.

\bibitem[Wang et~al.(2023)Wang, Zhang, Su, and Zhu]{wang2023comprehensive}
Liyuan Wang, Xingxing Zhang, Hang Su, and Jun Zhu.
\newblock A comprehensive survey of continual learning: Theory, method and
  application, 2023.

\bibitem[Wang et~al.(2022{\natexlab{d}})Wang, Shao, Lin, Hu, and
  Liu]{wang2022cmg}
Mengyu Wang, Yijia Shao, Haowei Lin, Wenpeng Hu, and Bing Liu.
\newblock Cmg: A class-mixed generation approach to out-of-distribution
  detection.
\newblock \emph{Proceedings of ECML/PKDD-2022}, 2022{\natexlab{d}}.

\bibitem[Wang et~al.(2022{\natexlab{e}})Wang, Zhang, Lee, Zhang, Sun, Ren, Su,
  Perot, Dy, and Pfister]{wang2022learning}
Zifeng Wang, Zizhao Zhang, Chen-Yu Lee, Han Zhang, Ruoxi Sun, Xiaoqi Ren,
  Guolong Su, Vincent Perot, Jennifer Dy, and Tomas Pfister.
\newblock Learning to prompt for continual learning.
\newblock In \emph{Proceedings of the IEEE/CVF Conference on Computer Vision
  and Pattern Recognition}, pp.\  139--149, 2022{\natexlab{e}}.

\bibitem[Wei et~al.(2022)Wei, Xie, Cheng, Feng, An, and Li]{wei2022mitigating}
Hongxin Wei, Renchunzi Xie, Hao Cheng, Lei Feng, Bo~An, and Yixuan Li.
\newblock Mitigating neural network overconfidence with logit normalization.
\newblock 2022.

\bibitem[Wightman(2019)]{rw2019timm}
Ross Wightman.
\newblock Pytorch image models.
\newblock \url{https://github.com/rwightman/pytorch-image-models}, 2019.

\bibitem[Wold et~al.(1987)Wold, Esbensen, and Geladi]{wold1987principal}
Svante Wold, Kim Esbensen, and Paul Geladi.
\newblock Principal component analysis.
\newblock \emph{Chemometrics and intelligent laboratory systems}, 2\penalty0
  (1-3):\penalty0 37--52, 1987.

\bibitem[Wortsman et~al.(2020)Wortsman, Ramanujan, Liu, Kembhavi, Rastegari,
  Yosinski, and Farhadi]{wortsman2020supermasks}
Mitchell Wortsman, Vivek Ramanujan, Rosanne Liu, Aniruddha Kembhavi, Mohammad
  Rastegari, Jason Yosinski, and Ali Farhadi.
\newblock Supermasks in superposition.
\newblock \emph{Advances in Neural Information Processing Systems},
  33:\penalty0 15173--15184, 2020.

\bibitem[Xiang \& Shlizerman(2023)Xiang and Shlizerman]{xiang2023tkil}
Jinlin Xiang and Eli Shlizerman.
\newblock Tkil: Tangent kernel optimization for class balanced incremental
  learning.
\newblock In \emph{Proceedings of the IEEE/CVF International Conference on
  Computer Vision}, pp.\  3529--3539, 2023.

\bibitem[Yan et~al.(2021)Yan, Xie, and He]{yan2021dynamically}
Shipeng Yan, Jiangwei Xie, and Xuming He.
\newblock Der: Dynamically expandable representation for class incremental
  learning.
\newblock In \emph{Proceedings of the IEEE/CVF Conference on Computer Vision
  and Pattern Recognition}, pp.\  3014--3023, 2021.

\bibitem[Yang et~al.(2022)Yang, Wang, Zou, Zhou, Ding, Peng, Wang, Chen, Li,
  Sun, Du, Zhou, Zhang, Hendrycks, Li, and Liu]{yang2022openood}
Jingkang Yang, Pengyun Wang, Dejian Zou, Zitang Zhou, Kunyuan Ding, Wenxuan
  Peng, Haoqi Wang, Guangyao Chen, Bo~Li, Yiyou Sun, Xuefeng Du, Kaiyang Zhou,
  Wayne Zhang, Dan Hendrycks, Yixuan Li, and Ziwei Liu.
\newblock Openood: Benchmarking generalized out-of-distribution detection.
\newblock 2022.

\bibitem[Yun et~al.(2019)Yun, Han, Oh, Chun, Choe, and Yoo]{yun2019cutmix}
Sangdoo Yun, Dongyoon Han, Seong~Joon Oh, Sanghyuk Chun, Junsuk Choe, and
  Youngjoon Yoo.
\newblock Cutmix: Regularization strategy to train strong classifiers with
  localizable features.
\newblock In \emph{Proceedings of the IEEE/CVF international conference on
  computer vision}, pp.\  6023--6032, 2019.

\bibitem[Zeng et~al.(2019)Zeng, Chen, Cui, and Yu]{zeng2019continual}
Guanxiong Zeng, Yang Chen, Bo~Cui, and Shan Yu.
\newblock Continual learning of context-dependent processing in neural
  networks.
\newblock \emph{Nature Machine Intelligence}, 1\penalty0 (8):\penalty0
  364--372, 2019.

\bibitem[Zhang et~al.(2023)Zhang, Inkawhich, Linderman, Chen, and
  Li]{zhang2023mixture}
Jingyang Zhang, Nathan Inkawhich, Randolph Linderman, Yiran Chen, and Hai Li.
\newblock Mixture outlier exposure: Towards out-of-distribution detection in
  fine-grained environments.
\newblock In \emph{Proceedings of the IEEE/CVF Winter Conference on
  Applications of Computer Vision}, pp.\  5531--5540, 2023.

\bibitem[Zhou et~al.(2023)Zhou, Ye, Zhan, and Liu]{zhou2023revisiting}
Da-Wei Zhou, Han-Jia Ye, De-Chuan Zhan, and Ziwei Liu.
\newblock Revisiting class-incremental learning with pre-trained models:
  Generalizability and adaptivity are all you need.
\newblock \emph{arXiv preprint arXiv:2303.07338}, 2023.

\bibitem[Zhu et~al.(2021)Zhu, Zhang, Wang, Yin, and Liu]{zhu2021prototype}
Fei Zhu, Xu-Yao Zhang, Chuang Wang, Fei Yin, and Cheng-Lin Liu.
\newblock Prototype augmentation and self-supervision for incremental learning.
\newblock In \emph{Proceedings of the IEEE/CVF Conference on Computer Vision
  and Pattern Recognition}, pp.\  5871--5880, 2021.

\end{thebibliography}
\bibliographystyle{iclr2024_conference}

\newpage
\appendix

\addcontentsline{toc}{section}{Appendix} 
\part{Appendix of TPL} 
\parttoc 
\newpage

\section{A Comprehensive Study on TIL+OOD based methods}
\label{app.ood+til}
Based on the previous research on CIL that combines TIL method and OOD detection~\citep{kimtheoretical,kim2022multi,kim2023learnability}, we make a thorough study on this paradigm by benchmarking 20 popular OOD detection methods. This study provides a comprehensive understanding of these methods and draws some interesting conclusions, which we believe are beneficial to the future study of CIL.

\subsection{A unified CIL method based on TIL+OOD}

We first describe how to design a CIL method based on TIL+OOD paradigm. As \texttt{HAT} is already a near-optimal solution for TIL~\citep{kim2023learnability} which has almost no CF, we choose \texttt{HAT} as the TIL method and only introduce variants in the OOD detection parts.

\textbf{Categorization of OOD detection methods.} Usually, an OOD detection technique consists of two parts: (1) Training a classifier that can better distinguish IND data and OOD data in the feature / logits space; (2) Applying an inference-time OOD score to compute the INDness of the test case $\boldsymbol{x}$. For training techniques, a group of methods such as \texttt{LogitNorm}~\cite{wei2022mitigating}, \texttt{MCDropout}~\cite{gal2016dropout}, \texttt{Mixup}~\cite{thulasidasan2019mixup}, 
\texttt{CutMix}~\cite{yun2019cutmix} apply data augmentation or confidence regularization to improve OOD detection; another series of methods such as \texttt{OE}~\cite{hendrycks2018deep}, \texttt{MixOE}~\cite{zhang2023mixture}, \texttt{CMG}~\cite{wang2022cmg}, \texttt{VOS}~\cite{du2021vos} use auxiliary OOD data in training. For inference-time techniques, the OOD score can be computed based on \emph{feature} (the hidden representations) or \emph{logits} (the unnormalized softmax score). \texttt{GradNorm}~\cite{huang2021importance},
\texttt{KL-Matching}~\cite{hendrycks2019scaling},
\texttt{OpenMax}~\cite{bendale2016towards},
\texttt{ODIN}~\cite{liang2017enhancing}, 
\texttt{MSP}~\cite{hendrycks2016baseline}, 
\texttt{MLS}~\cite{hendrycks2019scaling},
\texttt{EBO}~\cite{liu2020energy}, \texttt{ReAct}~\cite{sun2021react} are logit-based methods, and \texttt{Residual}~\cite{ndiour2020out}, \texttt{KNN}~\cite{sun2022out},
\texttt{MDS}~\cite{lee2018simple} are feature-based methods. Recently, \texttt{VIM}~\citep{wang2022vim} is based on both feature and logits. We study the aforementioned 20 OOD detection methods in this section. For simplicity, we use \texttt{MSP} (maximum softmax probability) as the OOD score for the 8 training-time techniques, and adopt standard supervised learning with cross-entropy loss for the inference-time OOD scores.

\textbf{A unified method for CIL based on TIL+OOD.} We introduce a unified CIL method that is based on TIL+OOD paradigm, which can be compatible with \textbf{any} OOD detection techniques. The details are as follows. 

We first recall that \texttt{HAT} learns each task by performing two functions jointly  \textit{\textbf{in training}}: (1) learning a model for the task and (2) identifying the neurons that are important for the task and setting masks on them. When a new task is learned, the gradient flow through those masked neurons is blocked in the backward pass, which protects the models of the previous tasks to ensure no forgetting (CF). In the forward pass, no blocking is applied so that the tasks can share a lot of parameters or knowledge. Since (1) can be any supervised learning method, in our case, we replace it with an \textbf{OOD detection} method, which performs both in-distribution (IND) classification  and out-of-distribution (OOD) detection. 

\textbf{\textit{In testing}}, we first predict the task-id to which the test instance $\boldsymbol{x}$ belongs, and then perform the IND classification (within task prediction) in the task to obtain the predicted class. Let $S(\boldsymbol{x};t)$ be the OOD score of $\boldsymbol{x}$ in task $t$ based on task $t$'s model, the task-id $\hat t$ can be predicted by identifying the task with the highest OOD score: 

\[\hat{t}=\argmax_t S(\boldsymbol{x};t)\]

We use argmax here as OOD score is defined to measure the IND-ness in the literature. Note that OOD score is produced by any OOD detection method.

\subsection{Experimental Setup}

\textbf{Backbone Architecture.} 
Following the main experiment in our paper, we use DeiT-S/16~\citep{touvron2021training} that is pre-trained using 611 classes of ImageNet after removing 389 classes that are similar or identical to the classes of the experiment data CIFAR and TinyImageNet to prevent information leak. We insert an adapter module~\citep{houlsby2019parameter} at each transformer layer. The adapter modules, classifiers and the layer norms are trained using \texttt{HAT} while the transformer parameters are fixed to prevent forgetting in the pre-trained network.

\textbf{Evaluation Protocol.}  
We compute AUC for OOD detection on each task model. The classes of the task are the IND classes while the classes of all other tasks of the dataset are the OOD classes. The evaluation metric for CIL is accuracy (ACC), which is measured after all tasks are learned. We report the average AUC value over all the tasks in each dataset, and the ACC of each dataset. Note that as \texttt{KNN} needs the training data at test time, in the CIL setting, we can only use the saved replay data of each task for its OOD score computation as the full data of previous tasks are not accessible in CIL. We also compute TIL accuracy in using difference OOD training-time techniques.

\subsection{Results and Analysis}

\begin{table}

\caption{Average AUC and ACC results based on pre-trained DeiT on five datasets with five random seeds. Bold and underlined numbers indicate the best and second-best results, respectively. $\heartsuit$: logit-based OOD scores, $\spadesuit$: feature-based OOD scores, $\diamondsuit$: training-time techniques. The systems are divided into three categories by dashed lines. The first category includes post-hoc OOD detectors, the second category includes methods that exploit surrogate OOD data, and the last category includes methods that employ special training strategies.}

\resizebox{\linewidth}{!}{
\centering
\begin{tabular}{c|cccccccccc|cc} 
\toprule
          & \multicolumn{2}{c}{\textbf{C10-5T}} & \multicolumn{2}{c}{\textbf{C100-10T}} & \multicolumn{2}{c}{\textbf{C100-20T}} & \multicolumn{2}{c}{\textbf{T-5T}} & \multicolumn{2}{c|}{\textbf{T-10T}} & \multicolumn{2}{c}{\textbf{Average}}  \\
\Bstrut
Method    & OOD  & CIL                          & OOD  & CIL                            & OOD  & CIL                            & OOD  & CIL                        & OOD  & CIL                          & OOD  & CIL                            \\ 
\hline 
\Tstrut
GradNorm $\heartsuit$  & $\text{73.8}^{\pm \text{0.37}}$ & $\text{43.5}^{\pm \text{0.48}}$                         & $\text{84.0}^{\pm\text{ 0.33}}$ & $\text{48.6}^{\pm \text{0.81}}$                           & $\text{80.8}^{\pm \text{0.31}}$ & $\text{22.9}^{\pm \text{1.91}}$                           & $\text{80.8}^{\pm \text{0.21}}$ & $\text{57.0}^{\pm \text{0.49}}$                       & $\text{83.1}^{\pm \text{0.24}}$ & $\text{47.4}^{\pm \text{0.85}}$                         & $\text{80.5}$ & $\text{43.9}$                           \\
KL-Matching $\heartsuit$        & $\text{75.2}^{\pm \text{0.31}}$ & $\text{48.8}^{\pm \text{0.44}}$                         & $\text{84.1}^{\pm \text{0.35}}$ & $\text{48.7}^{\pm \text{0.74}}$                           & $\text{82.6}^{\pm\text{ 0.32}}$ & $\text{25.6}^{\pm \text{1.81}}$                           & $\text{77.8}^{\pm \text{0.25}}$ & $\text{47.6}^{\pm \text{0.58}}$                       & $\text{73.6}^{\pm \text{0.29}}$ & $\text{31.6}^{\pm\text{ 0.79}}$                         & $\text{80.7}$ & $\text{40.5}$                           \\
OpenMax $\heartsuit$   &$ \text{94.1}^{\pm\text{ 0.14}}$ &$ \text{84.1}^{\pm \text{0.29}}$                         & $\text{84.7}^{\pm \text{0.26}}$ & $\text{50.9}^{\pm \text{0.62}}$                           & $\text{90.6}^{\pm \text{0.24}}$ & $\text{57.9}^{\pm \text{0.47}}$                           & $\text{74.1}^{\pm \text{0.19}}$ & $\text{48.1}^{\pm \text{0.24}}$                       & $\text{71.7}^{\pm \text{0.09}}$ & $\text{32.0}^{\pm \text{0.33}}$                         & $\text{83.0}$ & $\text{54.6}$                           \\
ODIN $\heartsuit$      & $\text{92.3}^{\pm \text{0.15}}$ & $\text{75.6}^{\pm \text{0.24}}$                         & $\text{89.8}^{\pm \text{0.19}}$ & $\text{61.7}^{\pm \text{0.40}}$                           & $\text{92.5}^{\pm\text{ 0.28}}$ & $\text{53.2}^{\pm \text{0.41}}$                           & $\text{85.8}^{\pm \text{0.09}}$ & $\text{64.5}^{\pm \text{0.10}}$                       & $\underline{\text{88.5}}^{\pm \text{0.07}}$ & $\text{59.5}^{\pm \text{0.25}}$                         & $\text{86.4}$ & $\text{62.9}$                           \\
MSP $\heartsuit$      & $\text{93.5}^{\pm \text{0.03}}$ & $\text{82.4}^{\pm \text{0.12}}$                         & $\text{89.0}^{\pm \text{0.11}}$ & $\text{62.9}^{\pm \text{0.24}}$                           & $\text{91.7}^{\pm \text{0.25}}$ & $\text{59.5}^{\pm\text{ 0.49}}$                           & $\text{81.5}^{\pm \text{0.10}}$ & $\text{59.2}^{\pm\text{ 0.21}}$                       & $\text{84.4}^{\pm \text{0.09}}$ & $\text{54.0}^{\pm \text{0.21}}$                         & $\text{88.0}$ & $\text{63.6}$                           \\
MLS $\heartsuit$       & $\text{94.0}^{\pm \text{0.11}}$ & $\underline{\text{84.9}}^{\pm \text{0.19}}$                         &$ \underline{\text{91.0}}^{\pm \text{0.18}}$ & $\underline{\text{69.2}}^{\pm \text{0.21}}$                           & $\text{92.7}^{\pm \text{0.22}}$ &$ \text{64.1}^{\pm \text{0.45}}$                           & $\text{86.1}^{\pm \text{0.08}}$ & $\underline{\text{65.4}}^{\pm \text{0.21}}$                       & $\underline{\text{88.5}}^{\pm \text{0.08}}$ & $\underline{\text{61.4}}^{\pm \text{0.27}}$                         & $\underline{\text{90.4}}$ & $\underline{\text{69.0}}$                         \\

EBO $\heartsuit$       & $\text{94.0}^{\pm \text{0.11}}$ &$ \underline{\text{84.9}}^{\pm \text{0.19}}$                         & $\text{90.8}^{\pm \text{0.18}} $& $\text{69.1}^{\pm \text{0.21}}$                           & $\text{92.5}^{\pm \text{0.21}}$ & $\text{64.1}^{\pm \text{0.45}}$                           &$ \underline{\text{86.2}}^{\pm \text{0.08}}$ & $\underline{\text{65.4}}^{\pm \text{0.20}}$                       & $\text{88.4}^{\pm \text{0.07}}$&$ \underline{\text{61.4}}^{\pm \text{0.28}} $                        & $\underline{\text{90.4}}$ & $\underline{\text{69.0}}$                           \\
ReAct $\heartsuit$      & $\text{94.0}^{\pm \text{0.11}}$ & $\underline{\text{84.9}}^{\pm \text{0.20}}$                         & $\text{90.8}^{\pm \text{0.19}}$ & $\text{69.1}^{\pm \text{0.22}}$                           & $\text{92.5}^{\pm \text{0.19}}$ & $\text{64.1}^{\pm \text{0.39}}    $                       & $\underline{\text{86.2}}^{\pm \text{0.09}}$ & $\underline{\text{65.4}}^{\pm \text{0.21}} $                      & $\text{88.4}^{\pm \text{0.05}}$ &$ \underline{\text{61.4}}^{\pm \text{0.28}}$                         & $\underline{\text{90.4}}$ & $\underline{\text{69.0}}$                           \\
KNN $\spadesuit$       & $\text{92.8}^{\pm \text{0.13}}$ & $\text{76.7}^{\pm \text{0.25}}$                         & $\text{85.9}^{\pm \text{0.14}}$& $\text{61.5}^{\pm \text{0.21}}  $                         &$ \text{90.2}^{\pm \text{0.11}}$  &$ \text{54.8}^{\pm \text{0.30}}  $                         & $\text{74.7}^{\pm \text{0.10}}$ & $\text{49.9}^{\pm \text{0.21}}$                       & $\text{79.2}^{\pm \text{0.08}}$ & $\text{44.4}^{\pm \text{0.21}}  $                       & $\text{84.6}$ & $\text{57.5}$                           \\
Residual $\spadesuit$  &$ \text{92.4}^{\pm \text{0.14}}$ &$ \text{83.0}^{\pm \text{0.17}}$                         & $\text{85.0}^{\pm \text{0.16}}$ & $\text{64.7}^{\pm \text{0.35}}$                           & $\text{89.8}^{\pm \text{0.14}}$ & $\text{61.3}^{\pm \text{0.31}}        $                   & $\text{78.0}^{\pm \text{0.11}}$ & $\text{53.7}^{\pm\text{0.24}} $                      & $\text{81.6}^{\pm \text{0.09}}$ & $\text{51.0}^{\pm \text{0.30}} $                        & $\text{85.4}$ & $\text{62.8}$                           \\
MDS $\spadesuit$        & $\text{92.6}^{\pm \text{0.12}}$ & $\text{85.7}^{\pm \text{0.02}}$                         & $\text{86.9}^{\pm\text{ 0.14}}$ & $\text{69.0}^{\pm \text{0.24}}$                           & $\text{91.2}^{\pm \text{0.15}}$ & $\text{65.4}^{\pm\text{0.42}}$                           & $\text{82.0}^{\pm \text{0.06}}$ & $\text{60.8}^{\pm \text{0.28}}$                       & $\text{84.4}^{\pm \text{0.05}}$ & $\text{56.9}^{\pm \text{0.30}}$                         & $\text{87.4}$ & $\text{67.6}$                           \\
\Bstrut
VIM $\heartsuit\spadesuit$       & $\textbf{95.4}^{\pm \text{0.07}}$ & $\textbf{89.0}^{\pm \text{0.23}}$                         & $\underline{\text{91.0}}^{\pm \text{0.12}} $&$ \textbf{72.8}^{\pm \text{0.30}}$                           & $\text{93.5}^{\pm \text{0.13}}$ & $\textbf{69.8}^{\pm \text{0.55}}$                           & $\textbf{86.3}^{\pm \text{0.05}}$ &$ \textbf{65.9}^{\pm \text{0.23}} $                      & $\textbf{88.7}^{\pm \text{0.07}}$ & $\textbf{63.1}^{\pm \text{0.42}}$                         & $\textbf{91.0}$ & $\textbf{72.1}$                           \\ 

\hdashline
\Tstrut
CMG $\diamondsuit$       & $\text{92.3}^{\pm \text{0.21}}$  & $\text{80.7}^{\pm \text{0.40}}$                          & $\text{85.2}^{\pm \text{0.28}}$  & $\text{56.2}^{\pm \text{0.60}}$                            & $\text{90.1}^{\pm \text{0.24}}$  & $\text{53.1}^{\pm \text{0.81}}  $                         & $\text{80.5}^{\pm \text{0.12}}$  & $\text{56.9}^{\pm \text{0.25}} $                       &$ \text{83.1}^{\pm \text{0.10}}$  &  $\text{50.5}^{\pm \text{0.31}} $                         & $\text{86.2}$  & $\text{59.5}$                            \\
VOS $\diamondsuit$       & $\text{93.1}^{\pm \text{0.10}}$  & $\text{82.1}^{\pm \text{0.31}}$                          & $\text{86.7}^{\pm \text{0.22}}$  & $\text{59.4}^{\pm \text{0.58}}$                            & $\text{90.9}^{\pm \text{0.18}}$  & $\text{56.0}^{\pm \text{0.66}} $                           &$\text{82.4}^{\pm \text{0.10}} $ & $\text{59.7}^{\pm\text{0.21}}$                        & $\text{83.8}^{\pm \text{0.10}}$  & $\text{51.8}^{\pm \text{0.34}}$                          & $\text{87.8}$  & $\text{61.8}$                            \\
OE $\diamondsuit$        & $\underline{\text{94.5}}^{\pm \text{0.10}}$ &$ \text{84.3}^{\pm \text{0.24}}$                     & $\text{90.9}^{\pm \text{0.15}} $&$ \text{66.7}^{\pm \text{0.31}}$                           & $\textbf{94.0}^{\pm \text{0.14}}$ & $\underline{\text{66.2}}^{\pm \text{0.40}}$                           & $\text{82.4}^{\pm \text{0.08}}$ &$ \text{60.5}^{\pm \text{0.19}} $                      & $\text{85.7}^{\pm \text{0.10}}$ &$ \text{56.4}^{\pm \text{0.33}}  $                       & $\text{89.5}$ & $\text{66.8}$                           \\
\Bstrut
MixOE $\diamondsuit$     & $\text{93.6}^{\pm \text{0.12}}$ & $\text{82.1}^{\pm \text{0.28}}$                         & $\textbf{91.7}^{\pm \text{0.19}}$ & $\text{67.6}^{\pm \text{0.37}}$                           & $\underline{\text{93.9}}^{\pm \text{0.15}}$ & $\text{62.9}^{\pm \text{0.37}}$                           & $\text{85.4}^{\pm \text{0.09}}$ & $\text{61.9}^{\pm \text{0.19}}$                       & $\text{87.4}^{\pm \text{0.11}}$ & $\text{56.0}^{\pm \text{0.36}}$                        & $\underline{\text{90.4}}$ & $\text{66.1}$                           \\
\hdashline
\Tstrut
LogitNorm $\diamondsuit$ & $\text{93.1}^{\pm \text{0.11}}$  & $\text{82.2}^{\pm \text{0.21}}$                          & $\text{89.0}^{\pm \text{0.17}}$  & $\text{64.3}^{\pm \text{0.35}}$                           & $\text{91.9}^{\pm \text{0.14}}$  & $\text{59.7}^{\pm \text{0.31}}$                            & $\text{81.7}^{\pm \text{0.05}}$  & $\text{58.7}^{\pm \text{0.15}} $                       & $\text{84.6}^{\pm \text{0.06}}$  & $\text{53.4}^{\pm \text{0.37}}$                          & $\text{88.1}$  & $\text{63.7}$                            \\
MCDropout $\diamondsuit$ & $\text{92.4}^{\pm \text{0.17}}$  & $\text{80.7}^{\pm \text{0.27}}$                          & $\text{87.8}^{\pm \text{0.19}}$  & $\text{61.7}^{\pm \text{0.38}}$                            & $\text{91.3}^{\pm \text{0.16}}$  & $\text{57.5}^{\pm \text{0.44}}$                            & $\text{80.9}^{\pm \text{0.08}}$  & $\text{57.5}^{\pm \text{0.19}}$                        & $\text{84.0}^{\pm \text{0.08}}$  & $\text{52.2}^{\pm \text{0.32}}$                          & $\text{87.3}$  & $\text{65.5}$                            \\
Mixup $\diamondsuit$     & $\text{89.8}^{\pm \text{0.62}}$ & $\text{73.8}^{\pm \text{1.54}}$                         & $\text{89.6}^{\pm \text{0.19}}$ & $\text{64.0}^{\pm \text{0.39}}$                           & $\text{91.3}^{\pm \text{0.18}} $&$ \text{55.4}^{\pm\text{0.42}}$                           & $\text{83.2}^{\pm\text{0.10}}$ & $\text{61.5}^{\pm \text{0.25}}$                    & $\text{85.6}^{\pm \text{0.07}}$ & $\text{56.0}^{\pm \text{0.34}}  $                       & $\text{87.9}$ & $\text{66.5}$                           \\
CutMix $\diamondsuit$    & $\text{90.4}^{\pm \text{0.26}}$ & $\text{69.0}^{\pm \text{0.71}}$                         & $\text{89.5}^{\pm\text{0.16}}$ & $\text{62.4}^{\pm \text{0.34}}$                          & $\text{90.9}^{\pm \text{0.15}}$ & $\text{50.6}^{\pm \text{0.40}}$                           & $\text{82.6}^{\pm \text{0.12}}$ & $\text{61.3}^{\pm \text{0.25}}$                       & $\text{85.1}^{\pm \text{0.22}}$ & $\text{55.2}^{\pm\text{0.51}}$                         & $\text{87.7}$ & $\text{59.7}$                           \\
\bottomrule
\end{tabular}}
\label{tab:ood}

\end{table}

The experiment results are given in Table~\ref{tab:ood}, which allow us to make some important  observations.

\textbf{(1) OOD detection and CIL performances. } The OOD detection results in AUC here have similar trends as those in \cite{yang2022openood} except \texttt{KNN}, which was considered as one of the best methods. But it is weak here because, as indicated above, in  CIL, \texttt{KNN} can only use the replay data (which is very small) for each task to compute the OOD score. The CIL performances of the top OOD methods are competitive compared to CIL baselines in Table~\ref{tab.results}.


\begin{table}[t]
\centering
\caption{Average IND classification ACC (TIL accuracy) results of pre-trained DeiT on five datasets with five random seeds. The baselines are divided into three categories as Table 1 in the main text. The first category includes post-hoc detectors, the second category includes methods that exploit surrogate OOD data, and the last category includes methods that employ special training strategies. $\heartsuit$: logit-based OOD scores, $\spadesuit$: feature-based OOD scores, $\diamondsuit$: training-time techniques. Note that the post-hoc methods only differ in the OOD score computation, which means they share the same trained model and thus have the same TIL ACC.}
\vspace{0.5em}
\resizebox{0.7\linewidth}{!}{
\begin{tabular}{c|ccccc|c} 
\toprule
\multicolumn{1}{l|}{} & \textbf{C10-5T}         & \textbf{C100-10T}       & \textbf{C100-20T}       & \textbf{T-5T}           & \textbf{T-10T}          & \textbf{Average}         \\ 
\hline
\Tstrut
GradNorm $\heartsuit$             & \multirow{12}{*}{99.20} & \multirow{12}{*}{95.71} & \multirow{12}{*}{97.50} & \multirow{12}{*}{84.40} & \multirow{12}{*}{88.10} & \multirow{12}{*}{92.98}  \\
KL-Matching $\heartsuit$          &                         &                         &                         &                         &                         &                          \\
OpenMax $\heartsuit$             &                         &                         &                         &                         &                         &                          \\
ODIN $\heartsuit$                 &                         &                         &                         &                         &                         &                          \\
MSP $\heartsuit$                  &                         &                         &                         &                         &                         &                          \\
MLS $\heartsuit$                  &                         &                         &                         &                         &                         &                          \\
EBO $\heartsuit$                  &                         &                         &                         &                         &                         &                          \\
ReAct $\heartsuit$                &                         &                         &                         &                         &                         &                          \\
KNN $\spadesuit$                  &                         &                         &                         &                         &                         &                          \\
Residual $\spadesuit$             &                         &                         &                         &                         &                         &                          \\
MDS $\spadesuit$                  &                         &                         &                         &                         &                         &                          \\
VIM $\heartsuit\spadesuit$                  &                         &                         &                         &                         &                         &                          \\
\hdashline
\Tstrut
CMG $\diamondsuit$                  & 99.20                   & 95.70                   & 97.61                   & 84.52                   & 88.30                   & 93.07                    \\
VOS $\diamondsuit$                  & 99.20                   & 95.66                   & 97.59                   & 84.51                   & 88.30                   & 93.05                    \\
OE $\diamondsuit$                   & 99.17                   & 95.73                   & 97.72                   & 84.46                   & 88.64                   & 93.14                    \\
MixOE $\diamondsuit$                & 99.21                   & 95.82                   & 97.80                   & 82.06                   & 86.53                   & 92.28                    \\
\hdashline
\Tstrut
LogitNorm $\diamondsuit$            & 99.20                   & 95.71                   & 97.56                   & 84.55                   & 87.87                   & 92.98                    \\
MCDropout $\diamondsuit$            & 99.20                   & 95.60                   & 97.62                   & 84.51                   & 88.23                   & 93.03                    \\
Mixup $\diamondsuit$                & 99.24                   & 95.57                   & 97.71                   & 85.04                   & 88.47                   & 93.21                    \\
CutMix $\diamondsuit$               & 99.20                   & 95.40                   & 97.54                   & 84.54                   & 88.40                   & 93.02                    \\
\bottomrule
\end{tabular}}
\label{tab:til}
\end{table}

\textbf{(2) Similar TIL Performance.} We present the detailed results of TIL accuracy, for the OOD detection baselines as~\cref{tab:til}. It is evident from the results that there is minimal variation in the performance of OOD detection methods across different datasets, with average values of $\text{99.2}{\pm \text{0.02}}$, $ \text{95.7}{\pm \text{0.12}}$, $\text{97.6}{\pm \text{0.10}}$, $\text{84.3}{\pm \text{0.86}}$, $\text{88.1}{\pm \text{0.86}}$ for C10-5T, C100-10T, C100-20T, T-5T, and T-10T, respectively. The observation that the majority of OOD methods exhibit negligible impact on the IND classification performance aligns with a previous benchmark study on OOD detection conducted by~\citet{yang2022openood}. Notably, the aforementioned study investigated the finding on ResNet architecture \citep{he2016deep} without pre-training, whereas our experiments involved a pre-trained DeiT model. This suggests that the finding extends to various model backbones. 

\begin{figure}[t]
\centering
\includegraphics[width=0.4\textwidth]{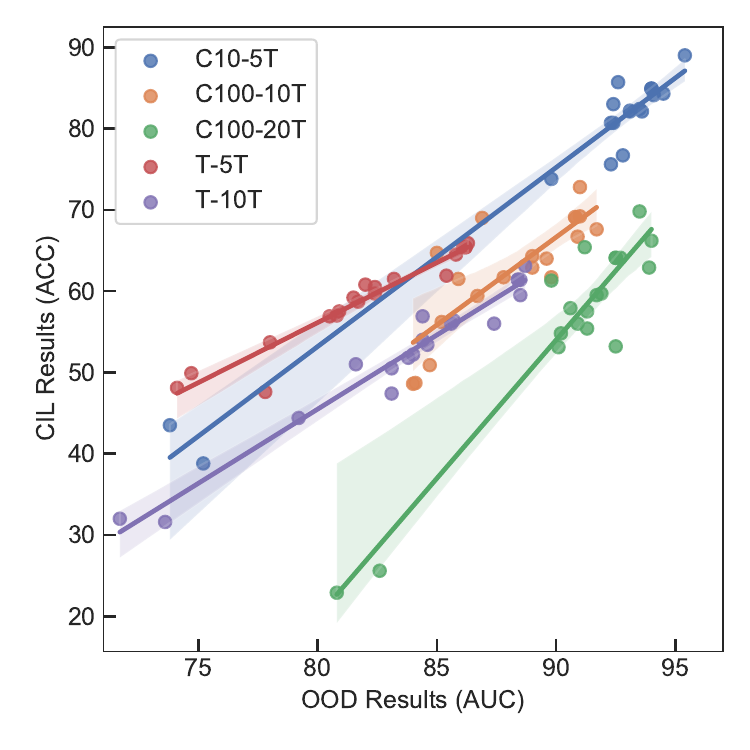}
\caption{The correlation between OOD (AUC) and CIL (ACC) results. Each point denotes the AUC and ACC of one method in~\cref{tab:ood} on the same dataset.}
\label{fig:correlation}
\end{figure}

\textbf{(3) Linear relationship between OOD AUC and CIL ACC}. We plot the relationship between OOD AUC and CIL performances in~\cref{fig:correlation}. Interestingly, we see a linear relationship 
with Pearson correlation coefficients of 0.976, 0.811, 0.941, 0.963, and 0.980 for the 5 datasets, respectively. This finding suggests that improving OOD AUC can bring about a linear improvement of $\times1.5\sim 3.4$ (which is the slopes of the fitted linear function) on CIL ACC. Note that we are not conditioning on TIL accuracy as (2) above showed the TIL results are similar for different training-time OOD techniques. 

\newpage

\section{Output Calibration}
\label{calibration}

We used the output calibration technique to balance the scales of task-id prediction scores for different tasks, which is motivated by~\citet{kim2022multi,kimtheoretical}. 
Even if the task-id prediction of each task-model is perfect, the system can make an incorrect task-id prediction if the magnitudes of the outputs across different tasks are different. 
As the task-specific modules are trained separately in \texttt{HAT}, it is useful to calibrate the outputs of different task modules. 

To ensure that the output values are comparable, we calibrate the outputs by scaling $\sigma_1^{(t)}$ and shifting $\sigma_2^{(t)}$ parameters for each task. The optimal parameters $\{(\sigma_1^{(t)},\sigma_2^{(t)})\}_{t=1}^T\in \mathbb R^{2T}$ ($T$ is the number of tasks) can be found by solving optimization problem using samples in the replay buffer \emph{Buf}. 

Specifically, we minimize the cross-entropy loss using SGD optimizer with batch size 64 for 100 epochs to find optimal calibration parameters $\{(\sigma_1^{(t)},\sigma_2^{(t)})\}_{t=1}^T$:
\begin{equation*}
    \mathcal L_{\textit{calibration}} = - \mathbb E_{(\boldsymbol x, y)\in \textit{Buf}} \log p(y|\boldsymbol x),
    \label{eq:calibration}
\end{equation*}
where $p(y|\boldsymbol x)$ is computed using~\cref{eq:til+ood_practice} and calibration parameters:
\begin{equation*}
    p(y_j^{(t)} | \boldsymbol{x}) = \sigma_1^{(t)}\cdot \left[\textit{softmax}\left(f(h(\boldsymbol{x}; \phi^{(t)}); \theta^{(t)})\right)\right]_j \cdot S(\boldsymbol{x};t) + \sigma_2^{(t)}
    \label{eq:final}
\end{equation*}
Given the optimal parameters $\{(\tilde{\sigma}_1^{(t)},\tilde{\sigma}_2^{(t)})\}_{t=1}^T$, we make the final prediction as:
\begin{equation*}
    \hat y = \argmax_{1\le t\le T, 1\le j \le |\mathcal Y_t|} p(y_j^{(t)} | \boldsymbol{x})
\end{equation*}

\newpage

\section{Evaluation Metrics}
\label{sec:metrics}

\subsection{Definitions of AIA. and Last accuracy}
\label{app.aia}

Fllowing~\citep{kim2022multi}, we give the formal definitions of \emph{average incremental accuracy} (\textbf{AIA} in~\Cref{tab.results} in the main text, denoted as $A_{\textit{AIA}}$) and \emph{accuracy after learning the final task} (\textbf{Last} in~\Cref{tab.results} in the main text, denoted as $A_{\textit{last}}$). Let the accuracy after learning the task $t$ be:
\[A^{(\le t)} = \frac{\# \textit{correctly}\ \textit{classified}\ \textit{samples}\ \textit{in}\ \bigcup_{k=1}^t \mathcal D_{\textit{test}}^{(k)}}{\# \textit{samples}\ \textit{in}\ \bigcup_{k=1}^t \mathcal D_{\textit{test}}^{(k)}}\]

Let $T$ be the last task. Then, $A_{\textit{last}}=A^{(\le T)}$ and 
$A_{\textit{AIA}} = \frac1T\sum_{k=1}^{T} A^{(\le k)}$.

Here $\mathcal D_{\textit{test}}^{(k)}$ denotes the test-set for task $k$, and $\#$ denotes ``the number of". To put it simply, $A^{(\le t)}$ means the accuracy of all the test data from task 1 to task $t$.  

\subsection{Rectified Forgetting Rate for CIL}

{Apart from the \emph{classification accuracy} (ACC), we report another popular CIL evaluation metric \emph{average forgetting rate}. The popular definition of average forgetting rate is the following 

\[\mathcal F^{(t)} = \frac1{t-1}\sum_{i=1}^{t-1} (A_i^{(i)} - A_i^{(t)}),\]

where $A_{i}^{(t)}$ is the accuracy of task $i$'s test set on the CL model after task $t$ is learned \citep{2020Mnemonics}, which is also referred to as \emph{backward transfer} in other literature \citep{lopez2017gradient}. 

However, this formula is only suitable for TIL but not appropriate for CIL. As the task-id for each test sample is given in testing for TIL, all the test samples from task $i$ will be classified into one of the classes of task $i$. If there is no forgetting for a TIL model, then $A_i^{(t)}$ will be equal to $A_i^{(i)} (i < t)$ in such a within-task classification, where the number of classes is fixed. But in CIL, the task-id is not provided in testing and we are not doing within-task classification. \textbf{Even if in the Non-CL setting as more tasks or classes are learned, the classification accuracy will usually decrease for the same test set due to the nature of multi-class classification.} That is, $A_i^{(t)} < A_i^{(i)} (i<t)$ is usually true as at task $i$ there are fewer learned classes than at task $t$. Their difference is not due to forgetting.  

Furthermore, as we discussed in~\Cref{sec:introduction},  forgetting (CF) is not the only issue of CIL. Inter-class separation (ICS) is another important one. When considering the performance degradation of each task during continual learning, it is hard to disentangle the effects of these two factors. Our rectified average forgetting rate metric for CIL considers both forgetting and ICS. Two new average forgetting rates for CIL are defined, one for the case where we use the Last accuracy as the evaluation metric and one for the case where we use AIA as the evaluation metric:


\[\mathcal F_{\textit{CIL, Last}}^{(t)} = \frac1{t}\sum_{i=1}^{t} (A_{i}^{(t,\textit{NCL})} - A_i^{(t)}) 
\] 

\[\mathcal F_{\textit{CIL, AIA}}^{(t)} = \frac1{t}\sum_{i=1}^{t}\mathcal F_{\textit{CIL, Last}}^{(i)},\] 

where NCL means Non-CL, $A_{i}^{(t)}$ is the accuracy of task $i$'s test set on the CL model after task $t$ is learned, and $A_{i}^{(t,\textit{NCL})}$ is the accuracy of task $i$'s test set on the Non-CL model that learns all tasks from 1 to $t$. If the test dataset sizes are the same across different tasks, then $\mathcal F_{\textit{CIL, Last}}^{(t)}$ is equal to $A_{\textit{last, NCL}} - A_{\textit{last}}$, where $A_{\textit{last}}$ is defined in~\cref{app.aia} and $A_{\textit{last, NCL}}$ is the $A_{\textit{last}}$ of Non-CL. The intuition of introducing NCL performance is to address the loss of accuracy for each task when more tasks are learned.

\label{sec:forgetting_rate}

\begin{table}[h]
\centering
\caption{Forgetting rate (\%) for CIL on the five datasets of the baselines in~\Cref{tab.results}. The lower the rate, the better the method is. }
\resizebox{\linewidth}{!}{
\Large
\begin{tabular}{c|cccccccccc|cc}
\toprule
\label{fig:fr}
\multirow{2}{*}{} &
  \multicolumn{2}{c}{\textbf{C10-5T}} &
  \multicolumn{2}{c}{\textbf{C100-10T}} &
  \multicolumn{2}{c}{\textbf{C100-20T}} &
  \multicolumn{2}{c}{\textbf{T-5T}} &
  \multicolumn{2}{c|}{\textbf{T-10T}} &
  \multicolumn{2}{c}{\textbf{Average}} \\ 
      Method   &  $\mathcal F_{\textit{CIL, Last}}^{(T)}$   &  $\mathcal F_{\textit{CIL, AIA}}^{(T)}$   &   $\mathcal F_{\textit{CIL, Last}}^{(T)}$   &   $\mathcal F_{\textit{CIL, AIA}}^{(T)}$   &   $\mathcal F_{\textit{CIL, Last}}^{(T)}$   &  $\mathcal F_{\textit{CIL, AIA}}^{(T)}$    &  $\mathcal F_{\textit{CIL, Last}}^{(T)}$    &  $\mathcal F_{\textit{CIL, AIA}}^{(T)}$    &  $\mathcal F_{\textit{CIL, Last}}^{(T)}$    &   $\mathcal F_{\textit{CIL, AIA}}^{(T)}$   &  $\mathcal F_{\textit{CIL, Last}}^{(T)}$    &    $\mathcal F_{\textit{CIL, AIA}}^{(T)}$  \Bstrut \\ \hline
\Tstrut 
OWM    & 54.10 & 41.01 & 61.37 & 47.10 & 65.78 & 54.95 & 47.97 & 31.85 & 55.00 & 41.28 & 56.84 & 43.24 \\
ADAM   & 11.87 & 6.68  & 21.55 & 14.65 & 23.77 & 16.64 & 22.41 & 15.18 & 22.84 & 15.59 & 20.49 & 13.75 \\
PASS   & 9.58  & 7.98  & 13.86 & 10.19 & 15.99 & 11.11 & 11.49 & 9.91  & 14.18 & 9.70  & 13.02 & 9.78  \\
HAT$_{\textit{CIL}}$    & 13.39 & 5.95  & 19.85 & 13.21 & 23.22 & 18.41 & 13.30 & 7.65  & 18.49 & 11.40 & 17.65 & 11.32 \\
iCaRL  & 39.46 & 28.82 & 57.55 & 43.37 & 60.77 & 51.56 & 41.99 & 27.77 & 50.62 & 37.45 & 50.08 & 37.79 \\
A-GEM  & 8.24 & 7.27 & 13.86 & 10.70 & 13.61 & 10.47 & 19.39 & 15.67 & 20.64 & 13.47 & 15.15 & 11.52 \\
EEIL   & 13.45 & 6.51  & 14.68 & 6.10  & 18.97 & 7.99  & 19.18 & 10.40 & 22.14 & 10.49 & 17.68 & 8.30  \\
GD     & 6.63  & 2.79  & 18.40 & 6.69  & 22.66 & 9.10  & 19.51 & 9.52  & 30.04 & 13.12 & 19.45 & 8.24  \\
DER++  & 9.00  & 4.18  & 9.46  & 4.31  & 10.76 & 4.74  & 12.95 & 6.71  & 15.34 & 6.82  & 11.50 & 5.35  \\
HAL    & 11.41 & 10.01 & 15.59 & 9.78  & 15.39 & 9.68  & 19.72 & 11.72 & 17.27 & 12.55 & 15.88 & 10.75 \\
DER++  & 11.16 & 8.00  & 13.03 & 6.56  & 12.73 & 5.81  & 16.68 & 10.48 & 18.32 & 9.89  & 14.38 & 8.15  \\
FOSTER & 9.70  & 5.47  & 11.07 & 6.04  & 9.85  & 4.51  & 18.08 & 7.08  & 16.82 & 7.03  & 13.10 & 6.03  \\
BEEF   & 8.69  & 3.91  & 10.67 & 5.29  & 10.88 & 6.08  & 11.11 & 5.82  & 14.36 & 5.87  & 11.14 & 5.39  \\
MORE     & 6.63 & 2.78 & 12.5 & 5.96 & 12.2 & 5.94 & 7.55  & 3.00& 9.46  & 4.29& 9.68 & 4.39 \\
ROW      & 4.82 & 2.56  & 8.04 & 4.33 & 8.16 & 4.41 & 7.41 & 2.87 & 9.31  & 4.12 & 7.55 & 3.66  \\
\hline
\Tstrut
TPL      & 3.46 & 1.90 & 6.23 & 3.10 & 6.42 & 3.07 & 3.88 & 0.26 & 5.32 & 1.31 & 5.06 & 1.93 \\
TPL$_{\text{PFI}}$ & 2.04 & 1.07  & 1.18 & 1.44 & 2.75 & 1.57 & 1.49  & 1.07  & 1.84  & 1.26 & 1.86 & 1.28 \\ \hline
\end{tabular}}
\end{table}

\Cref{fig:fr} shows the \textit{average forgetting rates} of each system based on the new definitions. We clearly observe that TPL and TPL$_{\text{PFI}}$ have the lowest average forgetting rates on the five datasets among all systems. {SLDA and L2P are not included as they use different architectures and cannot take the Non-CL results in \Cref{tab.results} in the main text as the upper bounds or NCL results needed in the proposed formulas above.

{It is important to note that for both TPL and TPL$_{\text{PFI}}$, the forgetting rate mainly reflects the performance loss due to the ICS problem rather than the traditional \textit{catastrophic forgetting} (CF) caused by network parameter interference in the incremental learning of different tasks because the TIL method \texttt{HAT} has effectively eliminated CF in TPL and TPL$_{\text{PFI}}$.} 
}

\newpage
\section{Additional Experimental Results}
\label{sec:more_experiments}

\subsection{CIL Experiments on a Larger Dataset}
\label{app:imagenet}

\begin{table}[htbp]
\centering
\large
\caption{The CIL ACC after the final task on ImageNet380-10T. We highlight the best results in bold.}
\vspace{0.5em}
\resizebox{0.75\linewidth}{!}{
\begin{tabular}{cccccccc} 
\toprule
HAT$_{CIL}$ & ADAM & SLDA & PASS & L2P & iCaRL & A-GEM & EEIL \\
\hline
\Tstrut
$\text{71.20}^{\pm0.99}$ & $\text{62.10}^{\pm0.91}$ & $\text{65.78}^{\pm0.05}$ & $\text{65.27}^{\pm1.24}$ & $\text{47.89}^{\pm3.24}$ & $\text{62.23}^{\pm0.66}$ & $\text{30.38}^{\pm10.02}$ & 
$\text{63.37}^{\pm0.49}$ \\
\toprule
DER++ & HAL &DER &FORSTER &BEEF & MORE &ROW & TPL \\
\hline\Tstrut
$\text{66.53}^{\pm2.36}$ & 
$\text{64.83}^{\pm2.60}$ &
$\text{69.19}^{\pm1.36}$ &
$\text{68.07}^{\pm1.88}$ &
$\text{70.07}^{\pm1.41}$ &
$\text{72.10}^{\pm1.44}$ & 
$\text{74.52}^{\pm1.38}$ &
$\textbf{78.49}^{\pm0.89}$
\\
\bottomrule
\end{tabular}}
\label{tab:imagenet}
\end{table}

To assess the performance of our proposed \texttt{TPL} on large-scale datasets, we use ImageNet-1k~\citep{imagenet15russakovsky}, a widely recognized benchmark dataset frequently examined in the CIL literature. However, due to the nature of our experiments, which involve a DeiT backbone pretrained on 611 ImageNet classes after excluding 389 classes similar to those in CIFAR and TinyImageNet, we cannot directly evaluate our model on the original ImageNet dataset to avoid potential information leak.

To overcome this limitation, we created a new benchmark dataset called \textbf{ImageNet-380}. We randomly selected 380 classes from the remaining 389 classes, excluding those similar to CIFAR and TinyImageNet, from the original set of 1k classes in the full ImageNet dataset. This new dataset consists of approximately 1,300 color images per class. For ImageNet-380, we divided the classes into 10 tasks, with each task comprising 38 classes. We set the replay buffer size to 7600, with 20 samples per class, which is a commonly used number in replay-based methods. We refer to these experiments as ImageNet380-10T. For other training configurations, we kept them consistent with the experiments conducted on T-10T.

The CIL Last ACC achieved after the final task on ImageNet380-10T can be found in~\Cref{tab:imagenet}. Notably, the results obtained by \texttt{TPL} still exhibit a significant improvement over the baselines, with a 3.97\% higher ACC compared to the best baseline method \texttt{ROW}. The results further provide strong evidence supporting the effectiveness of our proposed $\texttt{TPL}$ system.

\newpage
\subsection{CIL Experiments without Pre-training}
\label{app:cil_resnet}

\begin{table}[H]
\centering
\caption{CIL accuracy (\%) after the final task (last) based on ResNet-18 without pre-training over 5 runs with random seeds. ``-XT'': X number of tasks. The best result in each column is highlighted in bold. }
\label{tab.CIL_non_pretrain}
\resizebox{0.82\linewidth}{!}{
\begin{tabular}{c|ccccc|c} 
\toprule
\multicolumn{1}{l|}{} & \textbf{~~C10-5T~~}         & \textbf{C100-10T}       & \textbf{C100-20T}     & \textbf{~ ~T-5T~ ~}           & \textbf{~~T-10T~~}         &\textbf{Average}         \\ 
\hline
\Tstrut
OWM&$\text{51.8}^{\pm\text{0.05}}$ &$\text{28.9}^{\pm\text{0.60}}$ &$\text{24.1}^{\pm\text{0.26}}$ &$\text{10.0}^{\pm\text{0.55}}$ &$\text{8.6}^{\pm\text{0.42}}$ &$\text{24.7}$\\
PASS&$\text{47.3}^{\pm\text{0.98}}$ &$\text{33.0}^{\pm\text{0.58}}$ &$\text{25.0}^{\pm\text{0.69}}$ &$\text{28.4}^{\pm\text{0.51}}$ &$\text{19.1}^{\pm\text{0.46}}$ &$\text{30.6}$ \\
EEIL&$\text{64.5}^{\pm \text{0.93}}$ & $\text{52.3}^{\pm \text{0.83}}$ & $\text{48.0}^{\pm \text{0.44}}$& $\text{38.2}^{\pm \text{0.54}}$ & $\text{28.7}^{\pm \text{0.87}}$ &$\text{46.3}$ \\
GD  &$\text{65.5}^{\pm \text{0.94}}$ &$\text{51.4}^{\pm \text{0.83}}$ &$\text{50.3}^{\pm \text{0.88}}$ &$\text{38.9}^{\pm \text{1.01}}$ &$\text{29.5}^{\pm \text{0.68}}$ &$\text{47.1}$ \\
HAL &$\text{63.7}^{\pm\text{0.91}}$ &$\text{51.3}^{\pm\text{1.22}}$ &$\text{48.5}^{\pm\text{0.71}}$ &$\text{38.1}^{\pm \text{0.97}}$ &$\text{30.3}^{\pm\text{1.05}}$ &$\text{46.4}$ \\
A-GEM &$\text{64.6}^{\pm\text{0.72}}$&$\text{50.5}^{\pm\text{0.73}}$ &$\text{47.3}^{\pm\text{0.87}}$ &$\text{37.3}^{\pm\text{0.89}}$ & $\text{29.4}^{\pm\text{0.95}}$&$\text{45.8}$ \\
HAT$_{CIL}$&$\text{62.7}^{\pm\text{1.45}}$&$\text{41.1}^{\pm\text{0.93}}$&$\text{25.6}^{\pm\text{0.51}}$&$\text{38.5}^{\pm\text{1.85}}$&$\text{29.8}^{\pm\text{0.65}}$&$\text{39.5}$ \\
iCaRL&$\text{63.4}^{\pm\text{1.11}}$&$\text{51.4}^{\pm\text{0.99}}$&$\text{47.8}^{\pm\text{0.48}}$&$\text{37.0}^{\pm\text{0.41}}$&$\text{28.3}^{\pm\text{0.18}}$&$\text{45.6}$\\
DER++&$\text{66.0}^{\pm\text{1.20}}$&$\text{53.7}^{\pm\text{1.20}}$&$\text{46.6}^{\pm\text{1.44}}$&$\text{35.8}^{\pm\text{0.77}}$&$\text{30.5}^{\pm\text{0.47}}$&$\text{46.5}$\\
DER& $\text{62.1}^{\pm\text{0.97}}$& $\textbf{64.5}^{\pm\text{0.85}}$& $\textbf{62.5}^{\pm\text{0.76}}$& $\text{43.6}^{\pm\text{0.77}}$& $\text{38.3}^{\pm\text{0.82}}$&$\text{54.2}$ \\
FOSTER&$\text{65.4}^{\pm\text{1.05}}$ &$\text{62.5}^{\pm\text{0.84}}$ &$\text{56.3}^{\pm\text{0.71}}$ &$\text{40.5}^{\pm\text{0.92}}$ &$\text{36.4}^{\pm\text{0.85}}$ &$\text{52.2}$ \\
BEEF &$\text{67.3}^{\pm\text{1.07}}$ &$\text{60.9}^{\pm\text{0.87}}$ &$\text{56.7}^{\pm\text{0.72}}$ &$\text{44.1}^{\pm\text{0.85}}$ &$\text{37.9}^{\pm\text{0.95}}$ &$\text{53.4}$ \\
MORE&$\text{70.6}^{\pm\text{0.74}}$&$\text{57.5}^{\pm\text{0.68}}$&$\text{51.3}^{\pm\text{0.89}}$&$\text{41.2}^{\pm\text{0.81}}$&$\text{35.4}^{\pm\text{0.72}}$&$\text{51.2}$\\
ROW&$\text{74.6}^{\pm\text{0.89}}$&$\text{58.2}^{\pm\text{0.67}}$&$\text{52.1}^{\pm\text{0.91}}$&$\text{42.3}^{\pm\text{0.69}}$&$\text{38.2}^{\pm\text{1.34}}$&$\text{53.1}$\\
TPL&$\textbf{78.4}^{\pm\text{0.78}}$&$\text{62.2}^{\pm\text{0.52}}$&$\text{55.8}^{\pm\text{0.57}}$&$\textbf{48.2}^{\pm\text{0.64}}$&$\textbf{42.9}^{\pm\text{0.45}}$&$\textbf{57.5}$ \\
\bottomrule
\end{tabular}}
\end{table}

The experimental setup and results for CIL baselines without pre-training are presented in this section.

\textbf{Training details.} We follow~\citet{kimtheoretical} to use ResNet-18~\citep{he2016deep} for all the datasets (CIFAR-10, CIFAR-100, TinyImageNet) and all the baselines excluding \texttt{OWM}. {\texttt{MORE}, \texttt{ROW}, and our \texttt{TPL} are designed for pre-trained models, and we adapt them by applying \texttt{HAT} on ResNet-18. All other baselines adopted ResNet-18 in their original paper.} 
\texttt{OWM} adopts AlexNet as it is hard to apply the method to the ResNet. For the replay-based methods, we also use the same buffer size as specified in~\Cref{sec:exp setup}. We use the hyper-parameters suggested by their original papers. For \texttt{MORE}, \texttt{ROW}, and our proposed \texttt{TPL}, we follow the hyper-parameters used in \texttt{HAT} for training.

\textbf{Experimental Results.} The results for CIL Last accuracy (ACC) after the final task are shown in~\Cref{tab.CIL_non_pretrain}. We can observe that the network-expansion-based approaches (\texttt{DER}, \texttt{FOSTER}, \texttt{BEEF}) and approaches that predict task-id based on TIL+OOD (\texttt{MORE}, \texttt{ROW}) are two competitive groups of CIL baselines. Our proposed \texttt{TPL} achieves the best performance on C10-5T, T-5T, and T-10T, while \texttt{DER} achieves the best on C100-10T and C100-20T. Overall, our proposed \texttt{TPL} achieves the best average accuracy over the 5 datasets (57.5\%) while \texttt{DER} ranks the second (with an average ACC of 54.2\%). 

\newpage
\subsection{CIL Experiments on Smaller Replay Buffer Size}
\label{sec.small_replay}

\begin{table}[H]
\centering
\caption{CIL accuracy (\%) after the final task (Last) with smaller replay buffer size over 5 runs with random seeds. ``-XT'': X number of tasks. The best result in each column is highlighted in bold. The replay buffer size is set to 100 for CIFAR-10, and 1000 for CIFAR-100 and TinyImageNet. The pre-trained model is used.}
\label{tab.small_replay_full}
\resizebox{0.82\linewidth}{!}{
\begin{tabular}{c|ccccc|c} 
\toprule
\multicolumn{1}{l|}{} & \textbf{~~C10-5T~~}         & \textbf{C100-10T}       & \textbf{C100-20T}     & \textbf{~ ~T-5T~ ~}           & \textbf{~~T-10T~~}         &\textbf{Average}         \\ 
\hline
\Tstrut
iCaRL&$\text{86.08}^{\pm\text{1.19}}$ &$\text{66.96}^{\pm\text{2.08}}$ &$\text{68.16}^{\pm\text{0.71}}$ &$\text{47.27}^{\pm\text{3.22}}$ &$\text{49.51}^{\pm\text{1.87}}$ 
&$\text{63.60}$\\
A-GEM&$\text{56.64}^{\pm\text{4.29}}$ &$\text{23.18}^{\pm\text{2.54}}$ &$\text{20.76}^{\pm\text{2.88}}$ &$\text{31.44}^{\pm\text{3.84}}$ &$\text{23.73}^{\pm\text{6.27}}$ &$\text{31.15}$ \\
EEIL&$\text{77.44}^{\pm \text{3.04}}$ & $\text{62.95}^{\pm \text{0.68}}$ & $\text{57.86}^{\pm \text{0.74}}$& $\text{48.36}^{\pm \text{1.38}}$ & $\text{44.59}^{\pm \text{1.72}}$ &$\text{58.24}$ \\
GD  &$\text{85.96}^{\pm \text{1.64}}$ &$\text{57.17}^{\pm \text{1.06}}$ &$\text{50.30}^{\pm \text{0.58}}$ &$\text{46.09}^{\pm \text{1.77}}$ &$\text{32.41}^{\pm \text{2.75}}$ &$\text{54.39}$ \\
DER++ &$\text{80.09}^{\pm\text{3.00}}$ &$\text{64.89}^{\pm\text{2.48}}$ &$\text{65.84}^{\pm\text{1.46}}$ &$\text{50.74}^{\pm \text{2.41}}$ &$\text{49.24}^{\pm\text{5.01}}$ &$\text{62.16}$ \\
HAL &$\text{79.16}^{\pm\text{4.56}}$&$\text{62.65}^{\pm\text{0.83}}$ &$\text{63.96}^{\pm\text{1.49}}$ &$\text{48.17}^{\pm\text{2.94}}$ & $\text{47.11}^{\pm\text{6.00}}$&$\text{60.21}$ \\
DER 
&$\text{85.11}^{\pm\text{1.44}}$
&$\text{72.31}^{\pm\text{0.78}}$ 
&$\text{70.25}^{\pm\text{0.98}}$ 
&$\text{58.07}^{\pm\text{1.40}}$ 
&$\text{55.85}^{\pm\text{1.23}}$
&$\text{68.32}$ \\
FOSTER &$\text{84.99}^{\pm\text{0.89}}$
&$\text{70.25}^{\pm\text{0.58}}$ 
&$\text{71.14}^{\pm\text{0.76}}$ 
&$\text{53.35}^{\pm\text{0.54}}$ 
&$\text{54.58}^{\pm\text{0.85}}$
&$\text{66.86}$ \\
BEEF &$\text{86.20}^{\pm\text{1.59}}$
&$\text{70.87}^{\pm\text{2.77}}$ 
&$\text{70.44}^{\pm\text{1.24}}$ 
&$\text{60.15}^{\pm\text{0.98}}$ 
&$\text{57.02}^{\pm\text{0.87}}$
&$\text{68.94}$ \\
MORE&$\text{88.13}^{\pm\text{1.16}}$&$\text{71.69}^{\pm\text{0.11}}$&$\text{71.29}^{\pm\text{0.55}}$&$\text{64.17}^{\pm\text{0.77}}$&$\text{61.90}^{\pm\text{0.90}}$&$\text{71.44}$ \\
ROW&$\text{89.70}^{\pm\text{1.54}}$&$\text{73.63}^{\pm\text{0.12}}$&$\text{71.86}^{\pm\text{0.07}}$&$\text{65.42}^{\pm\text{0.55}}$&$\text{62.87}^{\pm\text{0.53}}$ &$\text{72.70}$\\
TPL&$\textbf{91.76}^{\pm\text{0.44}}$&$\textbf{75.83}^{\pm\text{0.28}}$&$\textbf{75.65}^{\pm\text{0.54}}$&$\textbf{68.08}^{\pm\text{0.61}}$&$\textbf{66.48}^{\pm\text{0.47}}$&$\textbf{75.56}$ \\
\bottomrule
\end{tabular}}
\end{table}

\newpage
\subsection{Ablation on $\beta_1$ and $\beta_2$}
\label{sec.beta-ablation}
\begin{table}[H]
\centering
\caption{The CIL accuracy (\%) after leraning the final task (Last) of our TPL on C10-5T for different $\beta_1$ and $\beta_2$.}
\label{tab:beta1_beta2}
\resizebox{0.5\linewidth}{!}{
\begin{tabular}{c|ccccc} 
\toprule
\diagbox{$\beta_1$}{$\beta_2$} & 15.0   & 17.5 & 20.0   & 22.5 & 25.0    \\ 
\hline
\Tstrut

0.5                    & 91.9 & 91.9 & 91.8 & 92.5 & 92.2  \\ 
0.6                    & 92.1 & 91.0 & 92.2 & 91.6 & 92.0  \\ 
0.7                    & 91.8 & 92.1 & 92.3 & 92.4 & 91.9  \\ 
0.8                    & 92.0 & 92.4 & 91.7 & 91.9 & 91.7  \\ 
0.9                    & 92.3 & 92.5 & 92.0 & 91.9 & 91.5  \\
\bottomrule
\end{tabular}}
\end{table}

$\beta_1$ and $\beta_2$ are two scaling hyper-parameters used in the definition of task-id prediction score $S_{\textit{TPL}}^{(t)}(\boldsymbol{x})$.~\Cref{tab:beta1_beta2} shows an ablation for different $\beta_1$ and $\beta_2$ on C10-5T, which indicates that $\beta_1$ and $\beta_2$ do not affect results much. For other hyper-parameters used in \texttt{TPL}, see~\Cref{sec:implementation} for more details.
\newpage

\subsection{Experiments based on More Pre-trained Models}
\label{app.visual_encoder}
\begin{table}[h]

\caption{\textbf{Last} TIL and CIL accuracy results (after the last task is learned) for TPL based on different pre-trained visual encoders or models. }

\resizebox{\linewidth}{!}{
\centering
\begin{tabular}{c|c|cccccccccc|cc} 
\toprule 
        \textbf{Visual}  &  & \multicolumn{2}{c}{\textbf{C10-5T}} & \multicolumn{2}{c}{\textbf{C100-10T}} & \multicolumn{2}{c}{\textbf{C100-20T}} & \multicolumn{2}{c}{\textbf{T-5T}} & \multicolumn{2}{c|}{\textbf{T-10T}} & \multicolumn{2}{c}{\textbf{Average}}   \\
\Bstrut
\textbf{Encoder}  &\textbf{Pre-training}  & TIL  & CIL                          & TIL  & CIL                            & TIL  & CIL                            & TIL  & CIL                        & TIL  & CIL       & TIL  & CIL                            \\ 
\hline 
\Tstrut
DeiT-small-IN661 {\color{black}(TPL)}  & supervised
& 99.20 & 92.33              
& 95.71 & 76.53                      
& 97.50 & 76.34                      
& 84.40 & 68.64                     
& 88.10 & 67.20 
& 92.98 & 76.21 \\
\hdashline
\Tstrut
ViT-tiny   & \multirow{4}{*}{supervised}
& 98.80 & 91.38
& 95.36 & 76.79
& 97.52 & 75.83
& 82.37 & 71.85                      
& 85.10 & 70.18
& 91.83 & 77.21  \\
DeiT-tiny   &
& 98.85 & 90.79
& 94.80 & 74.01
& 97.67 & 73.21
& 83.58 & 72.46
& 86.54 & 71.71
& 92.29 & 76.44  \\
ViT-small &
& 99.43 & 95.57
& 97.51 & 84.52
& 98.76 & 83.94
& 89.49 & 81.80
& 91.30 & 80.94
& 95.30 & 85.35  \\
DeiT-small {\color{black}(TPL$_{\text{PFI}}$)}&
& 99.24 & 94.86
& 96.79 & 82.43                      
& 97.78 & 80.86                       
& 89.78 & 84.06                     
& 92.51 & 83.87 
& 95.22 & 85.22 \\
ViT-small-Dino & self-supervised
& 98.75 & 87.82
& 94.51 & 73.83
& 96.95 & 72.62
& 80.77 & 68.78
& 82.49 & 65.97
& 90.69 & 73.80 \\
ViT-base-MAE & self-supervised
& 99.09 & 88.82
& 93.42 & 67.47
& 96.77 & 69.52
& 79.58 & 65.94
& 81.76 & 63.10
& 90.12 & 70.97 \\
\bottomrule
\end{tabular}}
\label{tab:visual_encoder}

\end{table}

In this section, we conduct an ablation study on the visual encoder (pre-trained model/network). To prevent data contamination or information leak, TPL uses DeiT-S/16 pre-trained with 611 classes of ImageNet (DeiT-small-IN661) after removing 389 classes that overlap with classes in the continual learning datasets. Here we dismiss this limitation and experiment on more open-sourced pre-trained visual encoders trained using the full ImageNet. The results are reported in~\Cref{tab:visual_encoder}, which also includes the TIL accuracy results if the task-id is provided for each test instance during testing.  We show DeiT-small-IN661, which is our TPL, in the first row of the table. It is the backbone used in our main experiments (\Cref{tab.results}) (IN: ImageNet). The other visual encoders are open-sourced in the timm~\citep{rw2019timm} library. We note that vanilla ViT and DeiT are pre-trained on ImageNet using \textbf{supervised training}, thereby there exists an information leak for CIFAR and Tiny-ImageNet datasets used in continual learning. The details of the models are as follows:

\begin{itemize}
    \item ViT-tiny: The full model name in timm is ``vit\_tiny\_patch16\_224''. It is trained on ImageNet (with additional augmentation and regularization) using supervised learning.
    \item DeiT-tiny: The full model name in timm is ``deit\_tiny\_patch16\_224''. It is trained on ImageNet (with additional augmentation and regularization) using supervised learning and distillation.
    \item ViT-small: The small version of ViT (``vit\_small\_patch16\_224'').
    \item DeiT-small: The small version of DeiT (``deit\_small\_patch16\_224'').
    \item ViT-small-Dino: The small version of ViT trained with self-supervised DINO method~\citep{caron2021emerging}.
    \item ViT-base-MAE: The base version of ViT trained with self-supervised MAE method~\citep{mae}.
\end{itemize}

\textbf{Analysis.} We found that different pre-trained visual encoders have varied CIL results. Compared to DeiT-small-IN661, the pre-trained small ViT and DeiT that use the full ImageNet to conduct supervised learning have overall better performance, which is not surprising due to the class overlap as we discussed above. For the self-supervised visual encoders (Dino, MAE), the performances are poorer than the supervised pre-trained visual encoders. We hypothesize that the variance between different visual encoders is mainly rooted in the learned feature representations during pre-training. It is an interesting future work to design an optimal pre-training strategy for continual learning
that does not need supervised data.

\newpage

\section{Theoretical Justifications}
\label{proof1}

\subsection{Preliminary}

In this section, we first give some primary definitions and notations of \emph{statistical hypothesis testing}, \emph{rejection region} and \textit{uniformly most powerful} (\emph{UMP}) \emph{test}.

\begin{definition}[statistical hypothesis testing and rejection region]
Consider testing a null hypothesis $H_0: \theta \in \Theta_0$ against an alternative hypothesis $H_1: \theta \in \Theta_1$, where $\Theta_0$ and $\Theta _1$ are subsets of the parameter space $\Theta$ and $\Theta_0 \cap \Theta_1 = \emptyset$. A test consists of a test statistic $T(X)$, which is a function of the data $\boldsymbol x$, and a rejection region $\mathcal R$, which is a subset of the range of $T$. If the observed value $t$ of $T$ falls in $\mathcal R$, we reject $H_0$.
\end{definition}

\textbf{Type I error} occurs when we reject a true null hypothesis $\mathcal H_0$. The probability of making Type I error is usually denoted by $\alpha$. \textbf{Type II error} occurs when we \emph{fail to reject} a false null
hypothesis $\mathcal H_0$. The \textbf{level of significance} $\alpha$ is the probability we are
willing to risk rejecting $\mathcal H_0$ when it is true. Typically $\alpha=0.1,0.05,0.01$ are used.

\begin{definition}[UMP test]
Denote the power function $\beta _{\mathcal R} (\theta) = P_\theta (T(x) \in \mathcal R)$, where $P_\theta$ denotes the probability measure when $\theta$ is the true parameter. A test with a test statistic $T$ and rejection region $\mathcal R$ is called a \textbf{uniformly most powerful} (UMP) test at significance level $\alpha$ if it satisfies two conditions:
\begin{enumerate}
    \item $\sup _{\theta \in \Theta_0}\beta _{\mathcal R} (\theta) \leq \alpha$.
    \item $\forall \theta \in \Theta_1, \beta _{\mathcal R}(\theta) \geq \beta _{R'} (\theta)$ for every other test $t'$ with rejection region $R'$ satisfying the first condition.
\end{enumerate}

\end{definition}

From the definition we see, the UMP test ensures that the probability of Type I error is less than $\alpha$ (with the first condition), while achieves the lowest Type II error (with the second condition). Therefore, UMP is considered an optimal solution in statistical hypothesis testing. 

\subsection{Proof of Theorem 4.1}

\begin{lemma}\citep{neyman1933ix}
\label{neyman}
Let $\{X_1,X_2,...,X_n\}$ be a random sample with likelihood function $L(\theta)$. The UMP test of the simple hypothesis $H_0:\theta = \theta _0$ against the simple hypothetis $H_a:\theta = \theta _a$ at level $\alpha$ has a rejection region of the form:
$$\dfrac{L(\theta _0)}{L(\theta _a)} < k$$
where $k$ is chosen so that the probability of a type I error is $\alpha$.
\end{lemma}

Now the proof of ~\Cref{theorem:lr} is straightforward. From Lemma~\ref{neyman}, the UMP test for~\Cref{hypothesis test} in the main text has a rejection region of the form:
\[\dfrac{p_t(\boldsymbol x)}{p_{t^c}(\boldsymbol x)} < \lambda_0\]
where $\lambda _0$ is chosen so that the probability of a type I error is $\alpha$.

\subsection{Proof of Theorem 4.2}

Note that the AUC is computed as the \emph{area under the ROC curve}. A ROC curve shows the trade-off between true positive rate (TPR) and false positive rate (FPR) across different decision thresholds. Therefore, 

\begin{align}
\textit{AUC} &=  \int_0^1 (\textit{TPR})\ \mathrm{d} (\textit{FPR}) \\
&=\int_0^1 (1-\textit{FPR})\ \mathrm{d}(\textit{TPR}) \\
&= \int_0^1 \beta_{\mathcal R}(\theta_{t^c})\ \mathrm{d} (1-\beta_{\mathcal R}(\theta_{t})) \\
&=\int_0^1 \beta_{\mathcal R}(\theta_{t^c})\ \mathrm{d} \beta_{\mathcal R}(\theta_{t})
\end{align}

where FPR and TPR are \emph{false positive rate} and \emph{true positive rate}. Therefore, an optimal AUC requires UMP test of any given level $\alpha = \beta_{\mathcal R}(\theta _{t})$ except on a null set. 

\subsection{Distance-Based OOD Detectors are IND Density Estimators}
\label{sec:distance_density}

In this section, we show that $S_{\textit{MD}}(\boldsymbol x)$ (MD: \textit{Mahalanobis distance}) and $S_{\textit{KNN}}(\boldsymbol x)$ (KNN: \textit{k}-\textit{nearest neighbor}) defined in~\Cref{eq:maha} and~\Cref{eq:knn} are IND (in-distribution) density estimators under different assumptions ({We omit the superscript ($t$) for simplicity)}:
\begin{align}
    S_{\textit{MD}}(\boldsymbol{x}) = 1/\min_{c\in\mathcal Y}(\boldsymbol{z} - \boldsymbol{\mu}_c)^T \boldsymbol{\Sigma}^{-1} (\boldsymbol{z} - \boldsymbol{\mu}_c),\label{eq:maha}\\
    S_{\textit{KNN}}(\boldsymbol{x};\mathcal D) = -||\boldsymbol{z}^* - \textit{kNN}(\boldsymbol{z}^*; \mathcal D^*)||_2.\label{eq:knn}
\end{align}
In~\Cref{eq:maha}, $\boldsymbol{\mu}_c$ is the class centroid for class $c$ and $\Sigma$ is the global covariance matrix, which are estimated on IND training corpus $\mathcal D$. In~\Cref{eq:knn}, $||\cdot||_2$ is Euclidean norm, $\boldsymbol{z}^*=\boldsymbol{z}/||\boldsymbol{z}||_2$ denotes the normalized feature $\boldsymbol{z}$, and $\mathcal D^*$ denotes the set of normalized features from training set $\mathcal D$. $\textit{kNN}(\boldsymbol{z}^*;{\mathcal D}^*)$ denotes the $k$-nearest neighbor of $\boldsymbol{z}^*$ in set ${\mathcal D}^*$.

Assume we have a feature encoder $\phi: \mathcal{X} \rightarrow \mathbb{R}^m$, and in training time we empirically observe $n$ IND samples $\{\phi (\boldsymbol x_1), \phi (\boldsymbol x_2)...\phi (\boldsymbol x_n)\}$.

\paragraph{Analysis of the MD score.} Denote $\boldsymbol \Sigma$ to be the covariance matrix of $\phi (\boldsymbol x)$. The final feature we extract from data $\boldsymbol x$ is:
$$\boldsymbol{z}(\boldsymbol x) = A^{-1} \phi (\boldsymbol x)$$
where $AA^T = \Sigma$. Note that the covariance of $\boldsymbol z$ is $\mathcal{I}$.

Given a class label $c$, we assume the distribution $z(x|c)$ follows a Gaussian $\mathcal{N}(A^{-1}\mu_c, \mathcal{I})$. Immediately we have $\boldsymbol \mu_c$ to be the class centroid for class $c$ under the maximum likelihood estimation. We can now clearly address the relation between MD score and IND density ($p(\boldsymbol{x})$): 
$$S_{\textit{MD}} (\boldsymbol x) = 1/(-2\max _{c \in \mathcal{Y}} (\ln{p(\boldsymbol x|c)}) - m \ln 2\pi)$$

\paragraph{Analysis of KNN score.} The normalized feature $\boldsymbol z(x) = \phi(\boldsymbol x) / ||\phi(\boldsymbol x)||_2$ is used for OOD detection. The probability density of $\boldsymbol{z}$ can be attained by:
$$p(\boldsymbol z) = \lim _{r\rightarrow 0} \dfrac{p(\boldsymbol z'\in B(\boldsymbol z, r))}{|B(z, r)|}$$
where $B(\boldsymbol z, r) = \{ \boldsymbol z': ||\boldsymbol z' - \boldsymbol z||_2 \leq r \wedge ||\boldsymbol z'||=1\}$

Assuming each sample $\boldsymbol z(\boldsymbol x_i)$ is $i.i.d$ with a probability mass $1/n$, the density can be estimated by KNN distance. Specifically, $r = ||\boldsymbol z - \textit{kNN}(\boldsymbol z)||_2$, $p(\boldsymbol z'\in B(\boldsymbol z, r)) = k/n$ and $|B(\boldsymbol z, r)| = \dfrac{\pi ^{(m-1)/2}}{\Gamma (\frac{m-1}{2} + 1)} r^{m-1} + o(r^{m-1})$, where $\Gamma$ is Euler's gamma function. When $n$ is large and $k / n$ is small, we have the following equations:
\begin{align*}
p(\boldsymbol x) &\approx \dfrac{k\Gamma (\frac{m-1}{2} + 1)}{\pi ^{(m-1)/2}nr^{m-1}}
\end{align*}
\begin{align*}
S_{\textit{KNN}}(\boldsymbol x) &\approx -(\dfrac{k\Gamma (\frac{m-1}{2} + 1)}{\pi ^{(m-1)/2}n})^{\frac{1}{m-1}} (p(\boldsymbol x))^{-\frac{1}{m-1}}
\end{align*}

{Recall that the CIL methods based on the TIL+OOD paradigm (i.e., \texttt{MORE} and \texttt{ROW}) use $S_{\textit{MD}}(\boldsymbol{x})$ to compute the task-prediction probability. As analyzed above, the MD score is in fact IND density estimator, which means $S_{\textit{MD}}(\boldsymbol{x})$ measures the likelihood of the task distribution $\mathcal P_t$. Therefore, the TIL+OOD methods ignores the likelihood of the distribution of other tasks ($\mathcal P_{t^c}$), which may fail to make the accurate task prediction. We put the detailed analysis in~\Cref{app:vis}.}

\newpage 

\section{Additional Details about \texttt{TPL}}

\subsection{Computation of the MD Score}
\label{sec:MD}

Mahalanobis distance score (MD) is an OOD score function initially proposed by~\citet{lee2018simple}, which is defined as:
\begin{equation*}
    S_{\textit{MD}}(\boldsymbol x) = 1/\min_{c\in\mathcal Y}\left((h(\boldsymbol x)-\boldsymbol \mu_c)^T\boldsymbol\Sigma^{-1}(h(\boldsymbol x) - \boldsymbol \mu_c)\right),
\end{equation*}
where $h(\boldsymbol x)$ is the feature extractor of a tested OOD detection model $\mathcal M$, $\boldsymbol \mu_c$ is the centroid for a class $c$ and $\boldsymbol \Sigma$ is the covariance matrix. The estimations of $\boldsymbol \mu_c$ and $\boldsymbol \Sigma$ are defined by

\[\boldsymbol \mu_c = \frac1{N_c}\sum_{\boldsymbol x\in\mathcal D^c_{\textit{train}}}h(\boldsymbol x),\]

\[\boldsymbol\Sigma = \frac1N \sum_{c\in |\mathcal Y|}\sum_{\boldsymbol x\in \mathcal D_{\textit{train}}^c}(h(\boldsymbol x)-\boldsymbol \mu_c)(h(\boldsymbol x)-\boldsymbol \mu_c)^T,\]

where $\mathcal D_{\textit{train}}^{c}:=\{\boldsymbol x:(\boldsymbol x,y)\in \mathcal D_{\textit{train}}, y=c\}$, $\mathcal D_{\textit{train}}$ is the training set, $N$ is the total number of training samples, and $N_c$ is the number of training samples belonging to class $c$. 

In the CIL setting, we have to compute $\boldsymbol \mu_c$ and $\boldsymbol \Sigma$ for each task (assuming all classes in the task have the same covariance matrix) with trained task-specific model $\mathcal M^{(t)}$. 
Specifically, after training on the $t$-th task dataset $\mathcal D^{(t)}$, we compute:
\begin{align}
    \boldsymbol{\mu}_c^{(t)} &= \frac1{N_c}\sum_{(\boldsymbol{x},c)\in\mathcal D^{(t)}} h(\boldsymbol x; \phi^{(t)}), \quad\forall c \in \mathcal Y^{(t)} \label{eq:mu}\\
    \boldsymbol\Sigma^{(t)} &= \frac1{|\mathcal D^{(t)}|} \sum_{c\in |\mathcal Y^{(t)}|}\sum_{(\boldsymbol{x}, c)\in \mathcal D^{(t)}}(h(\boldsymbol x; \phi^{(t)})-\boldsymbol{\mu}_c^{(t)})(h(\boldsymbol x; \phi^{(t)})-\boldsymbol{\mu}_c^{(t)})^T \label{eq:sigma}
\end{align}
We put the memory budget analysis of the saved class centroids $\boldsymbol \mu_c^{(t)}$ and co-variance matrices $\boldsymbol\Sigma^{(t)}$ in~\Cref{sec:comput_budge_analysis}.

\subsection{Computation of Logit-based Scores}
\label{app:logit_score_compute}

In~\Cref{sec:ablation}, we compare the performance of TPL with different logit-based scores \texttt{MSP}, \texttt{EBO}, and \texttt{MLS}. They are defined as follows:

\begin{equation}
    S^{(t)}_{\textit{MSP}}(\boldsymbol{x}) = \max_{j=1}^{|\mathcal Y_t|}\textit{softmax}\left(f(h(\boldsymbol{x};\phi^{(t)});\theta^{(t)})\right)
\end{equation}

\begin{equation}
    S^{(t)}_{\textit{EBO}}(\boldsymbol{x}) = \log\sum_{j=1}^{|\mathcal Y_t|}\left(\exp\left\{f(h(\boldsymbol{x};\phi^{(t)});\theta^{(t)})\right\}\right)
\end{equation}

\begin{equation}
    S^{(t)}_{\textit{MLS}}(\boldsymbol{x}) = \max_{j=1}^{|\mathcal Y_t|}\left(f(h(\boldsymbol{x};\phi^{(t)});\theta^{(t)})\right)
\end{equation}

In traditional OOD detection works, they are tested effective in estimating the probability of ``$\boldsymbol{x}$ belongs to IND classes". Thus we can adopt them to design our \texttt{TPL} method to estimate the probability of ``$\boldsymbol{x}$ belongs to task $t$".

\subsection{Pseudo-code}

\begin{algorithm}[htbp]
	\caption{Compute \texttt{TPL} Score with the $t$-th Task-specific Model $\mathcal M^{(t)}$}
    \textbf{Input:} $\textit{Buf}^\star$: replay buffer data without classes of task $t$; $\boldsymbol{x}$: test sample; $t$: task-id; $\mathcal M^{(t)}$: the trained $t$-th task model $\mathcal M^{(t)}$ with feature extractor $h(\boldsymbol{x};\phi^{(t)})$ and classifier $f(\boldsymbol{x};\theta^{(t)})$; $\{\boldsymbol{\mu}_c^{(t)}\}_{c\in\mathcal Y^{(t)}}$: pre-computed class centroids for task $t$; $\boldsymbol{\Sigma}^{(t)}$: pre-computed covariance matrix for task $t$; $k$: KNN hyper-parameter; $1/\beta_1^{(t)}$: pre-computed empirical mean of $S_{\textit{MLS}}(\boldsymbol{x})$; $1/\beta_2^{(t)}$: pre-computed empirical mean of $S_{\textit{MD}}(\boldsymbol{x})$.\\
    \textbf{Return:} \texttt{TPL} Score $S_{\textit{TPL}}(\boldsymbol x)$
	\begin{algorithmic}[1]
            \State $S_{\textit{MLS}}(\boldsymbol x)\leftarrow\max_{c\in\mathcal Y^{(t)}}f(h(\boldsymbol x; \phi^{(t)});\theta^{(t)})_c$
            \State $S_{\textit{MD}}(\boldsymbol x)\leftarrow 1/\left(\min_{c\in\mathcal Y^{(t)}} ((h(\boldsymbol x;\phi^{(t)})-\boldsymbol{\mu}_c)^T\boldsymbol{\Sigma}^{-1}(h(\boldsymbol x;\phi^{(t)})-\boldsymbol{\mu}_c))\right)$
            \State $S_{\textit{MLS}}(\boldsymbol{x}) \leftarrow S_{\textit{MLS}}(\boldsymbol{x}) * \beta_1^{(t)}$
            \State $S_{\textit{MD}}(\boldsymbol{x}) \leftarrow S_{\textit{MD}}(\boldsymbol{x}) * \beta_2^{(t)}$
            \State $\boldsymbol{z} \leftarrow h({\boldsymbol{x}};\phi^{(t)}) / ||h({\boldsymbol{x}};\phi^{(t)})||_2$ 
            \For{$\hat{\boldsymbol x}_i$ in $\textit{Buf}^\star$}
            \State $\boldsymbol{z}_i \leftarrow h(\hat{\boldsymbol{x}}_i;\phi^{(t)}) / ||h(\hat{\boldsymbol{x}}_i;\phi^{(t)})||_2$ 
            \State $d_i\leftarrow ||\boldsymbol{z}_i- \boldsymbol{z}||_2$
            \EndFor
            \State $\{d_{i_j}\}_{j=1}^{|\textit{Buf}^\star|}\leftarrow sorted(\{d_{i}\}_{i=1}^{|\textit{Buf}^\star|})$
            \State $S_{\textit{TPL}}(\boldsymbol x) \leftarrow -\log [\exp\{-S_{\textit{MLS}}(\boldsymbol x)\}+ \exp\{-S_{\textit{MD}}(\boldsymbol x)- d_{i_k}\}]$
	\end{algorithmic} 
 \label{TPL prediction code}
\end{algorithm}

\begin{algorithm}[htbp]
	\caption{CIL Training with \texttt{TPL}}
	\begin{algorithmic}[1]
            \State Initialize an empty replay buffer $\textit{Buf}$
            \For{training data $\mathcal D^{(t)}$ of each task}
            \For{each batch $(\boldsymbol{x}_j, y_j)\subset \mathcal D^{(t)}\cup \textit{Buf}$, until converge}
            \State Minimize~\Cref{eq.training_loss} (in the main text) and update the parameters with \texttt{HAT}
            \EndFor
            \State Compute $\{\boldsymbol{\mu}_c^{(t)}\}_{c\in\mathcal Y^{(t)}}$ using~\Cref{eq:mu}
            \State Compute $\boldsymbol{\Sigma}^{(t)}$ using~\Cref{eq:sigma}
            \State Compute $1/\beta_1^{(t)},1/\beta^{(t)}$ using~\Cref{eq:beta1_beta2}
            \State Update $\textit{Buf}$ with $\mathcal D^{(t)}$ 
            \EndFor
            \State Train the calibration parameters $\{(\tilde{\sigma}_1^{(t)},\tilde{\sigma}_2^{(t)})\}_{t=1}^T$ following~\Cref{eq:calibration}
	\end{algorithmic} 
 \label{training code}
\end{algorithm}

\begin{algorithm}[htbp]
	\caption{CIL Testing with \texttt{TPL}}
        \textbf{Input}: test sample $\boldsymbol{x}$ \\
        \textbf{Return:} predicted class $\hat c$
	\begin{algorithmic}[1]
            \State Compute $S_{\textit{TPL}}(\boldsymbol{x};t$) with each task model $\mathcal M^{(t)}$ using~\Cref{TPL prediction code}.
            \State Preicition with~\Cref{eq:til+ood_practice}: $\hat{c} = \argmax_{c,t} \sigma_1^{(t)}\cdot \left[\textit{softmax}\left(f(h(\boldsymbol{x}; \phi^{(t)}); \theta^{(t)})\right)\right]_j \cdot S(\boldsymbol{x};t) + \sigma_2^{(t)}$
	\end{algorithmic} 
 \label{testing code}
\end{algorithm}

To improve reproducibility, we provide the detailed pseudo-code for computing $S_{\textit{TPL}}(\boldsymbol{x})$ (using~\Cref{eq:final_score} in the main text) as~\Cref{TPL prediction code}. Then we give the pseudo-code for the CIL training as~\Cref{training code} and testing for \texttt{TPL} as~\Cref{testing code}.

\newpage
\section{Details of \texttt{HAT}}
\label{HAT}

\subsection{Training}

For completeness, we briefly describe the hard attention mechanism of \texttt{HAT}~\citep{serra2018overcoming} used in \texttt{TPL}. In learning the task-specific model $\mathcal M^{(t)}$ for each task, \texttt{TPL} at the same time trains a \emph{mask} for each adapter layer. To protect the shared feature extractor from previous tasks, their masks are used to block those important neurons so that the new task learning will not interfere with the parameters learned for previous tasks. The main idea is to use sigmoid to approximate a 0-1 gate function as \emph{hard attention} to mask or unmask the information flow to protect parameters learned for each previous task.

The hard attention at layer $l$ and task $t$ is defined as:
\begin{equation*}
    \boldsymbol a_l^{(t)} = \textit{sigmoid} (s\cdot \boldsymbol e^{(t)}_l),
\end{equation*}
where $s$ is a temperature scaling term, $\textit{sigmoid}(\cdot)$ denotes the sigmoid function, and $\boldsymbol e^{(t)}_l$ is a \emph{learnable} embedding for task $t$. The attention is element-wise multiplied to the ouptut $\boldsymbol h_{l}$ of layer $l$ as
\begin{equation*}
    \boldsymbol h_{l}' = \boldsymbol a_l^{(t)} \otimes \boldsymbol h_l
\end{equation*}
The sigmoid function converges to a 0-1 binary gate as $s$ goes to infinity. Since the binary gate is not differentiable, a fairly large $s$ is chosen to achieve a differential pseudo gate function. The pseudo binary value of the attention determines how much information can flow forward and backward between adjacent layers. Denote $\boldsymbol{h}_{l} = \textit{ReLU}(\boldsymbol {W}_l\boldsymbol{h}_{l-1} + \boldsymbol{b}_l)$, where $\textit{ReLU}(\cdot)$ is the rectifier function. For neurons of attention $\boldsymbol{a}_l^{(t)}$ with zero values, we can freely change the corresponding parameters in $\boldsymbol{W}_l$ and $\boldsymbol b_l$ without interfering the output $\boldsymbol h_l'$. The neurons with non-zero mask values are necessary to perform the task, and thus need a protection for \emph{catastrophic forgetting} (CF).

Specifically, during learning task $t$, we modify the gradients of parameters that are important in performing the previous tasks $1,2,...,t-1$ so they are not interfered. Denote the accumulated mask by
\begin{equation*}
    \boldsymbol a_l^{(<t)} = \max(\boldsymbol a_l^{(<t-1)}, \boldsymbol a_l^{(t-1)}) ,
\end{equation*}
where $\max(\cdot,\cdot)$ is an element-wise maximum and the initial mask $\boldsymbol a_l^{(0)}$ is defined as a zero vector. $\boldsymbol a_l^{(<t)}$ is a collection of mask values at layer $l$ where a neuron has value 1 if it has ever been activated previously. The gradient of parameter $w_{ij,l}$ is modified as
\begin{equation*}
    \nabla w_{ij,l}' = (1-\min(a_{i,l}^{(<t)}, a_{j,l-1}^{(<t)}))\nabla w_{ij,l}, 
\end{equation*}
where $a_{i,l}^{(<t)}$ is the $i$-th unit of $\boldsymbol a_l^{(<t)}$. The gradient flow is blocked if both neurons $i$ in the current layer and $j$ in the previous layer have been activated. We apply the mask for all layers of adapters except the last layer. The parameters in last layer do not require protection as they are task-specific parameters.

A regularization is introduced to encourage sparsity in $\boldsymbol a_l^{(t)}$ and parameter sharing with $\boldsymbol a_l^{(<t)}$. The capacity of a network depletes when $\boldsymbol a_l^{(<t)}$ becomes an all-one vector in all layers. Despite a set of new neurons can be added in network at any point in training for more capacity, we utilize resources more efficiently by minimizing the loss:
\begin{equation*}
    \mathcal L_{\textit{reg}} = \frac{\sum_l\sum_i a_{i,l}^{(t)} (1-a_{i,l}^{(<t)})}{\sum_l\sum_i (1-a_{i,l}^{(<t)})} ,
\end{equation*}
The intuition of this term is to regularize the number of masked neurons. Then the loss of \texttt{HAT} defined as the second term from the R.H.S. of Eq (7) in the main text is:
\begin{equation*}
    \mathcal L_{\textit{HAT}} = \mu\cdot \mathcal L_{\textit{reg}} ,
\end{equation*}
where $\mu$ is a hyper-parameter to balance the optimization of classification objective and \texttt{HAT} regularization.

\subsection{Inference}

{
In CIL without a task identifier during inference, we are required to forward input data across each task to derive task-specific features $h^{(t)}$ for each task $t = 1, 2, \cdots, T$. This approach can result in computation overhead, especially with extended task sequences. This section proposes to use parallel computing (PC) to mitigate this by achieving comparable time efficiency to one-pass CL methods, albeit with a trade-off of increased memory usage by a factor of $T$  compared to the one-pass CL methods for latent feature storage, a vector of 384  floating point numbers. 

Consider the model $\mathcal{M}$ comprising two components: a feature extractor $h(\cdot)$ and a classifier $f(\cdot)$. Unlike the standard model, where both extractor and classifier are universal across tasks, our model uses a shared feature extractor with task-specific classifiers for each task. We analyze the computational costs for each component separately.

For the feature extractor, break it down into $L$ layers. Each layer $l$ involves an affine transformation (with weight $\boldsymbol{W}_l$ and bias $\boldsymbol{b}_l$) followed by an activation function, specifically $\textit{ReLU}(\cdot)$.\footnote{~This aligns with our adapter implementation approach.} In the standard model, computation at each layer $l$ follows:

\begin{equation}
\boldsymbol{h}_l = \textit{ReLU}(\boldsymbol{W}_l\boldsymbol{h}_{l-1}+\boldsymbol{b}_l)
\end{equation}

For our model, the computation is extended to:

\begin{equation}
\boldsymbol{h}_l^{(t)} = \boldsymbol{a}_l^{(t)} \otimes \textit{ReLU}(\boldsymbol{W}_l\boldsymbol{h}_{l-1}^{(t)}+\boldsymbol{b}_l),\quad t=1,2,\cdots,T
\end{equation}

Here, $\boldsymbol{a}_l^{(t)}$ denotes the stored hard attention at layer $l$ for task $t$. Comparing the equations, the additional operation in our method is the element-wise product, which is efficiently parallelizable by vectorizing the sets $\{\boldsymbol{h}_l^{(t)}\}_{t=1}^T$, $\{\boldsymbol{h}_{l-1}^{(t)}\}_{t=1}^T$, and $\{\boldsymbol{a}_l^{(t)}\}_{t=1}^T$ into matrices $\boldsymbol{H}_l$, $\boldsymbol{H}_{l-1}$, and $\boldsymbol{A}_l$. This parallelization achieves near-equivalent time consumption to the standard model, with the trade-off of a $T$-fold increase in memory usage as our feature $\boldsymbol{H}_{l}$ and $\boldsymbol{H}_{l-1}$ are $T$ times larger compared to $\boldsymbol{h}_{l}$ and $\boldsymbol{h}_{l-1}$.

For the classifier, each task employs a simple affine transformation (weight $\boldsymbol{W}^{(t)}$ and bias $\boldsymbol{b}^{(t)}$). The standard model uses $\boldsymbol{W} \in \R^{H \times C}$ and $\boldsymbol{b} \in \R^{C}$, where $H$ and $C$ represent the hidden size and class count, respectively. The standard model's computation is:

\begin{equation}
\textit{logits} = \boldsymbol{W} \boldsymbol{h}_L + \boldsymbol{b}
\end{equation}

In contrast, our method involves $T$ task-specific classifiers, each with weight $\boldsymbol{W}^{(t)} \in \R^{H \times \frac{C}{T}}$ and bias $\boldsymbol{b}^{(t)} \in \R^{\frac{C}{T}}$. The task-specific logits are calculated as:

\begin{equation}
\textit{logits}^{(t)} = \boldsymbol{W}^{(t)} \boldsymbol{h}_L^{(t)} + \boldsymbol{b}^{(t)}
\end{equation}

By vectorizing task-specific weights, features, and biases, time consumption is maintained at levels comparable to the standard model, and memory consumption remains unaffected.

In conclusion, our approach achieves similar time efficiency as the standard model in CL inference, at the expense of increased runtime memory for storing task-specific features. This trade-off also allows flexibility between memory usage and time consumption by adjusting the \textbf{level of parallelism}, offering a balance between runtime duration and memory requirements. For example, one can reduce the running time memory by forwarding the input more times to the model with a lower parallelism level each time.}

\subsection{Remarks}

{It is important to note that our method can also leverage some other TIL methods to prevent CF other than \texttt{HAT}. The reason we chose \texttt{HAT} is to make it easy to compare with previous techniques (e.g., \texttt{MORE} and \texttt{ROW}) based on the same setting. For example, we can also exploit \texttt{SupSup}~\citep{wortsman2020supermasks}, which has  already been applied in~\citet{kimtheoretical} to construct an effective TIL+OOD method and the empirical performance are similar to that using \texttt{HAT}. Furthermore, our method is also compatible with other architecture-based TIL methods such as \texttt{PNN}~\citep{rusu2016progressive}, \texttt{PackNet}~\citep{mallya2018packnet} or {Ternary Masks}~\citep{masana2021ternary}.}

\newpage
{
\section{Visualization of Task Distribution}
\label{app:vis}
The key theoretical analysis behind \texttt{TPL},~\cref{theorem:lr,theorem:max_auc}, suggest that an accurate estimate of $\mathcal P_{t^c}$ (the distribution of the other tasks, i.e., the complement) is important. {If we only estimate $\mathcal P_{t}$ (the feature distribution of task $t$) when doing task prediction and assume $\mathcal P_{t^c}$ to be a uniform distribution (which was done by existing 
methods as analyzed in~\Cref{sec:distance_density}}), there will be potential risks as shown by the toy example on 1D Gaussian in~\Cref{sec:likelihood_ratio_principle}. 

Intuitively, the failure happens when $\mathcal P_{t}$ and $\mathcal P_{t^c}$ have some overlap. A test case $\boldsymbol{x}_1$ may have higher likelihood in $\mathcal P_{t}$ than another test case $\boldsymbol{x}_2$, but if it gets even higher likelihood in $\mathcal P_{t^c}$, then $\boldsymbol{x}_1$ will be less likely to be drawn from $\mathcal P_t$. In this section, we plot $\mathcal P_t$ and $\mathcal P_{t^c}$ to demonstrate this phenomenon.

Recall that in our CIL scenario, we compute the likelihood ratio for each task to estimate the task prediction probability. For each task $t$, we compute the likelihood of input data $\boldsymbol{x}$ under the task distribution $\mathcal P_t$, and use the saved replay data from the other tasks to estimate $\mathcal P_{t^c}$. Notice that as we do not conduct task prediction in training, all the discussion here is about inference or test time and we have the small among of saved data from all the learned tasks in the memory buffer. Our goal is to analyze whether $\mathcal P_t$ and $\mathcal P_{t^c}$ have some overlap.

Specifically, we draw the feature distribution of the five tasks on C10-5T after all tasks are learned under the pre-training setting. {To facilitate an accurate estimation of $\mathcal P_t$ and $\mathcal P_{t^c}$, we use the whole training set of CIFAR-10 to prepare the extracted feature as it contains more data.} The vanilla features are high-dimensional vectors in $\R^{384}$, and we use Principled Component Analysis (PCA)~\citep{wold1987principal} to project the vectors into $\R^2$. We then use Kernel Density Estimation (KDE)~\citep{terrell1992variable} to visualize the distribution (density) of the data.
The figures are shown as follows:



\begin{figure}[H]
    \centering
    \vspace{-1em}
    \includegraphics[width=\linewidth]{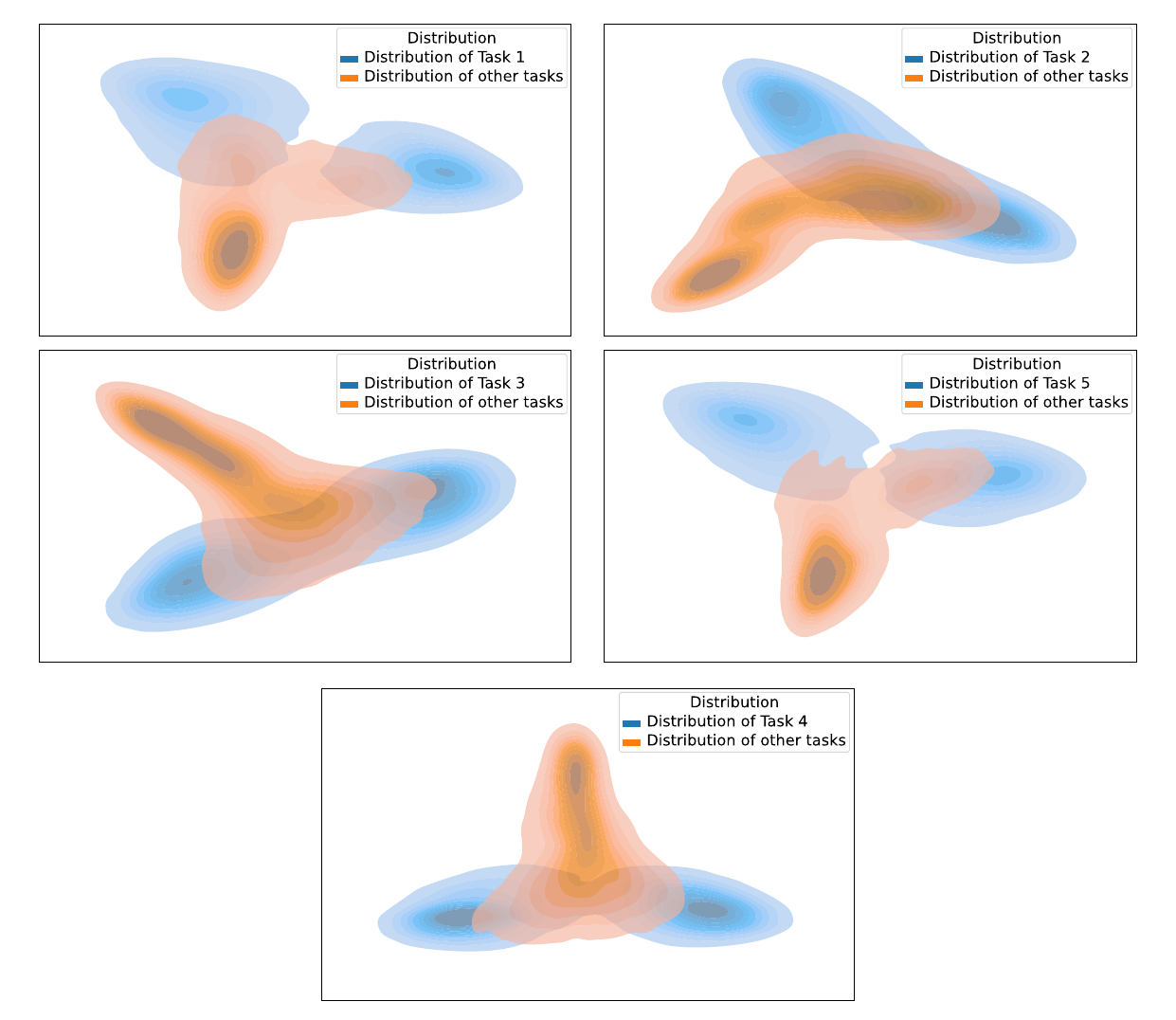}
    \caption{Visualization of feature distribution of Task $t$ ($t$=1,2,3,4,5) data and the other 4 tasks. We use the trained task-specific feature extractor $h(\boldsymbol{x};\phi^{(t)})$ to extract features from the {the training data} that belongs to task $t$ (which represent $\mathcal P_{t}$) and {the training data} that belongs to the other 4 tasks (which represent $\mathcal P_{t^c}$).}
    \label{fig:enter-label}
\end{figure}

Notice that the task distributions $\{\mathcal P_{t}\}$ are bimodal as  there are two classes in each task in C10-5T. The most interesting observation is that the distribution of the other tasks $\{\mathcal P_{t^c}\}$ have overlap with the task distribution $\{\mathcal P_{t}\}$. This indicates that the failure may happen as we analyzed above. For example, we draw a demonstrative failure case of the existing TIL+OOD methods {(i.e., \texttt{MORE} and \texttt{ROW})} in Task 1 prediction in~\Cref{fig:case}. In this Figure, TIL+OOD methods will compute the task prediction probability solely based on the high likelihood of $\mathcal P_{t}$ ($t=1$), while our TPL will consider both high likelihood of $\mathcal P_t$ and low likelihood of $\mathcal P_{t^c}$. In this case, the red star has higher likelihood in $\mathcal P_{t}$ than the green star (0.9 v.s. 0.4). However, the likelihood ratio between $\mathcal P_{t}$ and $\mathcal P_{t^c}$ of the red star is lower than the green star (3 v.s. 20). Therefore, using likelihood ratio between $\mathcal P_{t}$ and $\mathcal P_{t^c}$ is crucial in estimating the task prediction probability.

\begin{figure}[H]
    \centering
    \includegraphics[width=0.63\linewidth]{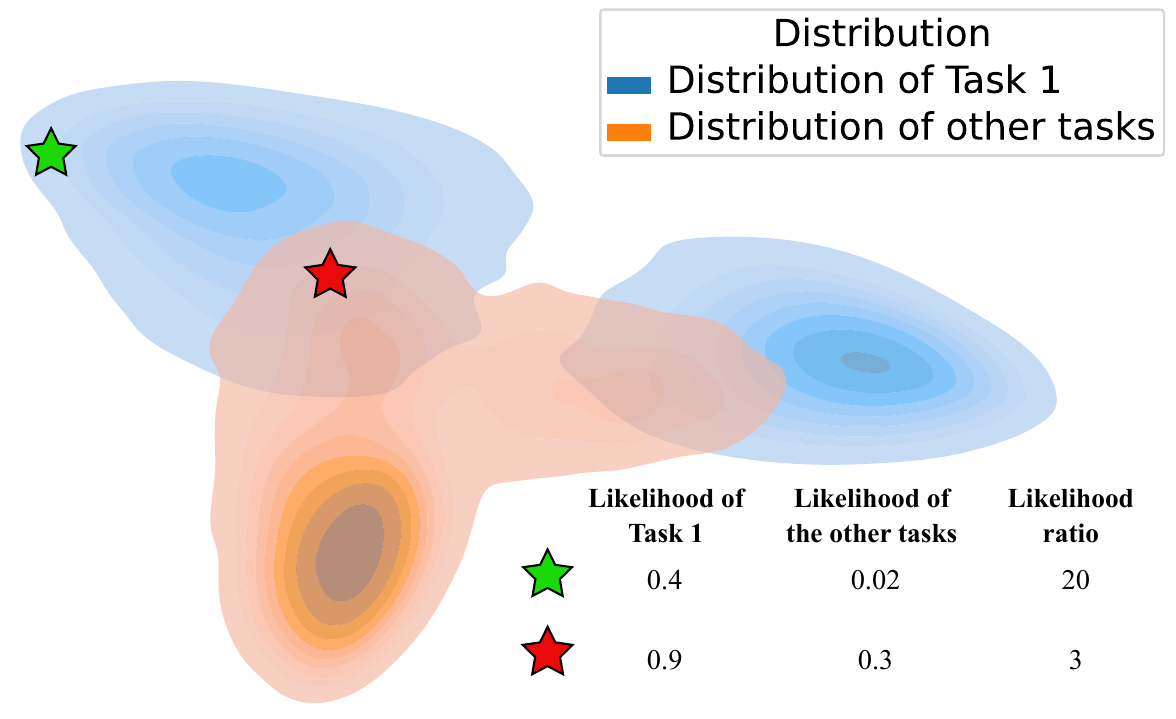}
    \caption{A failure case of TIL+OOD methods that predict the task based on the likelihood of $\mathcal P_{t}$ (e.g., \texttt{MORE} and \texttt{ROW}). In the figure, the red star has higher likelihood in $\mathcal P_{t}$ ($t=1$) than the green star. However, the likelihood ratio between $\mathcal P_{t}$ and $\mathcal P_{t^c}$ of the red star is lower than the green star. The correct choice is to accept the green star to be from Task 1 instead of the red star.}
    \label{fig:case}
\end{figure}
}
\newpage

\section{Implementation Details, Network Size and Running Time}
\label{app:training_details}

\subsection{Implementation Details of Baselines}
\label{sec:implementation}

\textbf{Datasets}. We use three popular image datasets. 
(1) \textbf{CIFAR-10}~\citep{krizhevsky2010convolutional} consists of 
images of 10 classes with 50,000 / 10,000 training / testing samples. (2) \textbf{CIFAR-100}~\citep{krizhevsky2009learning} consists of  
images with 50,000 / 10,000 training / testing samples. (3) \textbf{Tiny-ImageNet}~\citep{le2015tiny} has 120,000 
images of 200 classes with 500 / 50 images per class for training / testing.


\textbf{Implementation of CIL baselines (pre-trained setting).} For CIL baselines, we follow the experiment setups as reported in their official papers unless additionally explained in~\Cref{sec:exp setup}. For the regularization hyper-parameter $\mu$ and temperature annealing term $s$ (see~\Cref{HAT}) used in \texttt{HAT}, we follow the baseline \texttt{MORE} and use $\mu=0.75$ and $s=400$ for all experiments as recommended in~\citep{serra2018overcoming}. For the approaches based on network expansion (\texttt{DER}, \texttt{BEEF}, \texttt{FOSTER}), we expand the network of adapters when using a pre-trained backbone. For \texttt{ADAM}, we choose the \texttt{ADAM(adapter)} version in their original paper, which is the best variant of \texttt{ADAM}. As some of the baselines are proposed to continually learn from scratch, we carefully tune\textbf{ their hyper-parameters} to improve the performance to ensure a fair comparison. The implementation details of baselines under non-pre-training setting are shown in~\Cref{app:cil_resnet}.

\textbf{Hyper-parameter tuning.} Apart from $\beta_1$ and $\beta_2$ discussed in~\Cref{sec:ablation}, the only hyper-parameters used in our method \texttt{TPL} are $\gamma$, which is the temperature parameter for task-id prediction, and $k$, which is the hyper-parameter of $d_{\textit{KNN}}(\boldsymbol{x}, \textit{Buf}^\star)$ in~\Cref{eq:LR}. The value of $\gamma$ and $k$ are searched from $\{0.01, 0.05, 0.10, 0.50, 1.0, 2.0, 5.0, 10.0\}$ and $\{1,2,5,10,50,100\}$, respectively. We choose $\gamma=0.05$, and $k=5$ for all the experiments as they achieve the overall best results.

\textbf{Training Details.} To compare with the strongest baseline \texttt{MORE} and \texttt{ROW}, we follow their setup~\citep{kim2022multi, kim2023learnability} to set the training epochs as 20, 40, 15, 10 for CIFAR-10, CIFAR-100, T-5T, T-10T respectively. And we follow them to use SGD optimizer, the momentum of 0.9, the batch size of 64, the learning rate of 0.005 for C10-5T, T-5T, T-10T, C100-20T, and 0.001 for C100-10T.

\subsection{Hardware and Software}
We run all the experiments on NVIDIA GeForce RTX-2080Ti GPU. Our implementations are based on Ubuntu Linux 16.04 with Python 3.6.

\subsection{Computational Budget Analysis}
\label{sec:comput_budge_analysis}

\subsubsection{Memory Consumption} 

We present the network sizes of the CIL systems (with a pre-trained network) after learning the final task in~\Cref{tab.network_size}.

\begin{table}[h]
\centering
\caption{Network size measured in the number of parameters (\# parameters) for each method without the memory buffer.}

\resizebox{0.7\linewidth}{!}{
\begin{tabular}{c!{\vrule width \lightrulewidth}ccccc} 
\toprule
         & \textbf{~~ C10-5T~~} & \textbf{C100-10T} & \textbf{C100-20T} & \textbf{~ ~T-5T~ ~} & \textbf{~ ~T-10T~ ~} \\ 
\midrule
OWM      & 24.1M & 24.4M & 24.7M & 24.3M & 24.4M   \\
ADAM    & 22.9M & 24.1M & 24.1M & 24.1M & 24.1M \\
PASS    & 22.9M & 24.2M & 24.2M & 24.3M & 24.4M               \\
HAT$_{CIL}$     & 24.1M & 24.4M & 24.7M & 24.3M & 24.4M          \\
SLDA    & 21.6M & 21.6M & 21.6M & 21.7M & 21.7M               \\
\Bstrut L2P  & 21.7M & 21.7M & 21.7M & 21.8M & 21.8M               \\
\hdashline
\Tstrut iCaRL   & 22.9M & 24.1M & 24.1M & 24.1M & 24.1M               \\
A-GEM    & 26.5M & 31.4M & 31.4M & 31.5M & 31.5M              \\
EEIL    & 22.9M & 24.1M & 24.1M & 24.1M & 24.1M             \\
GD       & 22.9M & 24.1M & 24.1M & 24.1M & 24.1M             \\
DER++    & 22.9M & 24.1M & 24.1M & 24.1M & 24.1M           \\
HAL      & 22.9M & 24.1M & 24.1M & 24.1M & 24.1M                \\
DER      & 27.7M & 45.4M & 69.1M & 33.6M & 45.5M                \\
FOSTER      & 28.9M & 46.7M & 74.2M & 35.8M & 48.1M                \\
BEEF      & 30.4M & 48.4M & 82.3M & 37.7M & 50.6M                \\
MORE     & 23.7M & 25.9M & 27.7M & 25.1M & 25.9M   \\
ROW      & 23.7M & 26.0M & 27.8M & 25.2M & 26.0M                \\
\midrule 
\Bstrut
\textbf{TPL} & 23.7M & 25.9M & 27.7M & 25.1M & 25.9M   \\
\bottomrule
\end{tabular}}
\label{tab.network_size}
\end{table}

With the exception of \texttt{SLDA} and \texttt{L2P}, all the CIL methods we studied utilize trainable adapter modules. The transformer backbone consumes 21.6 million parameters, while the adapters require 1.2M for CIFAR-10 and 2.4M for other datasets. In the case of \texttt{SLDA}, only the classifier on top of the fixed pre-trained feature extractor is fine-tuned as it requires a fixed feature extractor for all tasks, while \texttt{L2P} maintains a prompt pool with 32k parameters. Additionally, each method requires some specific elements (e.g., task embedding for \texttt{HAT}), resulting in varying parameter requirements for each method.

As mentioned in~\Cref{sec:MD}, our method also necessitates the storage of class centroids and covariance matrices. For each class, we save a centroid of dimension 384, resulting in a total of 3.84k, 38.4k, and 76.8k parameters for CIFAR-10, CIFAR-100, and TinyImageNet, respectively. The covariance matrix is saved per task, with a size of $384\times 384$. {Then the parameter count of covariance matrix can be computed by $T\times 384\times 384$ for a dataset with $T$ tasks.} Consequently, the total parameter count is 
737.3k, 1.5M, 2.9M, 737.3k, and 1.5M for C10-5T, C100-10T, C100-20T, T-5T, and T-10T, respectively. It is worth noting that this consumption is relatively small when compared to certain replay-based methods like \texttt{iCaRL} and \texttt{HAL}, which require a teacher model of the same size as the training model for knowledge distillation.

\subsubsection{Running Time}

The computation for our method \texttt{TPL} is very efficient. It involves standard classifier training and likelihood raio score computation, which employed some OOD detection methods. The OOD score computation only involves mean, covariance computation, and KNN search, which are all very efficient with the Python packages \texttt{scikit-learn}\footnote{~https://scikit-learn.org/stable/} and \texttt{faiss}\footnote{~https://github.com/facebookresearch/faiss}. We give the comparison in running time in~\Cref{tab.running_time}. We use \texttt{HAT} as the base as \texttt{MORE}, \texttt{ROW} and \texttt{TPL} all make use of \texttt{HAT} and \texttt{MORE} and \texttt{ROW} are the strongest baselines.

\begin{table}[h]
\centering
\caption{Average running time measured in minutes per task (min/T) for three systems.}

\resizebox{0.7\linewidth}{!}{
\begin{tabular}{c!{\vrule width \lightrulewidth}ccccc} 
\toprule
         & \textbf{~~ C10-5T~~} & \textbf{C100-10T} & \textbf{C100-20T} & \textbf{~ ~T-5T~ ~} & \textbf{~ ~T-10T~ ~} \\ 
\midrule
HAT$_{CIL}$ & 17.8 min/T & 17.6 min/T & 9.4 min/T & 28.0 min/T & 9.48 min/T \\
MORE & 20.6 min/T & 23.3 min/T & 11.7 min/T & 32.8 min/T & 11.2 min/T \\
ROW& 21.8 min/T & 25.2 min/T & 12.6 min/T & 34.1 min/T & 11.9 min/T \\
TPL & 20.7 min/T & 23.3 min/T & 11.7 min/T & 32.5 min/T & 11.2 min/T \\
\bottomrule
\end{tabular}}
\label{tab.running_time}
\end{table}

\newpage

\section{Limitations}
\label{sec.limitations}

Here we discuss the limitations of our proposed method \texttt{TPL}. 

{\textbf{First,} our \texttt{TPL} method relies on the saved data in the memory buffer like traditional replay-based methods, which may have privacy concerns in some situations and also need extra storage. We will explore how to improve task-id prediction in the CIL setting without saving any previous data in our future work. \textbf{Second,} \texttt{TPL} uses a naive saving strategy that samples task data randomly to put in the memory buffer for simplicity. In our future work, we would also like to consider better buffer saving strategies~\citep{Jeeveswaran2023BiRTBR} and learning algorithm for buffer data~\citep{Bhat2022ConsistencyIT}, which may enable more accurate likelihood ratio computation. \textbf{Third,} this paper focuses on the traditional \emph{offline} CIL problem set-up. In this mode, all the training data for each task is available upfront when the task arrives and the training can take any number of epochs. Also, the label space $\mathcal Y^{(t)}$ of different tasks are disjoint. In online CIL~\citep{guo2022online}, the data comes in a stream and there may not be clear boundary between tasks (called \emph{Blurry Task Setting}~\citep{bang2022online}), where the incoming data labels may overlap across tasks. It is an interesting direction for our future work to explore how to adapt our method to exploit the specific information in this setting.}

\end{document}